\documentclass[10pt,twocolumn,letterpaper]{article}

\usepackage{cvpr}              %
\usepackage{graphicx}
\usepackage{amsmath}
\usepackage{amssymb}
\usepackage{booktabs}
\usepackage{colortbl}
\usepackage{subcaption}
\usepackage{multirow}
\usepackage{graphicx} %
\usepackage{soul}
\definecolor{mygray}{gray}{.9}

\usepackage[accsupp]{axessibility}  %

\newcommand{\CUT}[1]{}

\usepackage[pagebackref,breaklinks,colorlinks]{hyperref}

\usepackage[capitalize]{cleveref}
\crefname{section}{Sec.}{Secs.}
\Crefname{section}{Section}{Sections}
\Crefname{table}{Table}{Tables}
\crefname{table}{Tab.}{Tabs.}

\def\cvprPaperID{3654} %
\def\confName{CVPR}
\def\confYear{2023}

\begin{document}

\title{Fine-Grained Face Swapping via Regional GAN Inversion}

\author{Zhian Liu${^{1\dag}}$ Maomao Li${^{2\dag}}$ Yong Zhang${^{2\dag}}$ Cairong Wang${^{3}}$ Qi Zhang${^2}$ Jue Wang${^2}$ Yongwei Nie${^{1*}}$ \\
${^1}$ South China University of Technology \qquad ${^2}$Tencent AI Lab  \\
${^3}$Tsinghua Shenzhen International Graduate School \\
}

\twocolumn[{%
\renewcommand\twocolumn[1][]{#1}%
\vspace{-1.3cm}
\maketitle
\vspace{-1.3cm}
\begin{center}
    \centering
    \captionsetup{type=figure}
    \captionsetup[subfigure]{labelformat=empty}
    \begin{subfigure}[b]{0.107\textwidth}
      \begin{subfigure}[t]{\textwidth}
        \includegraphics[width=\textwidth]{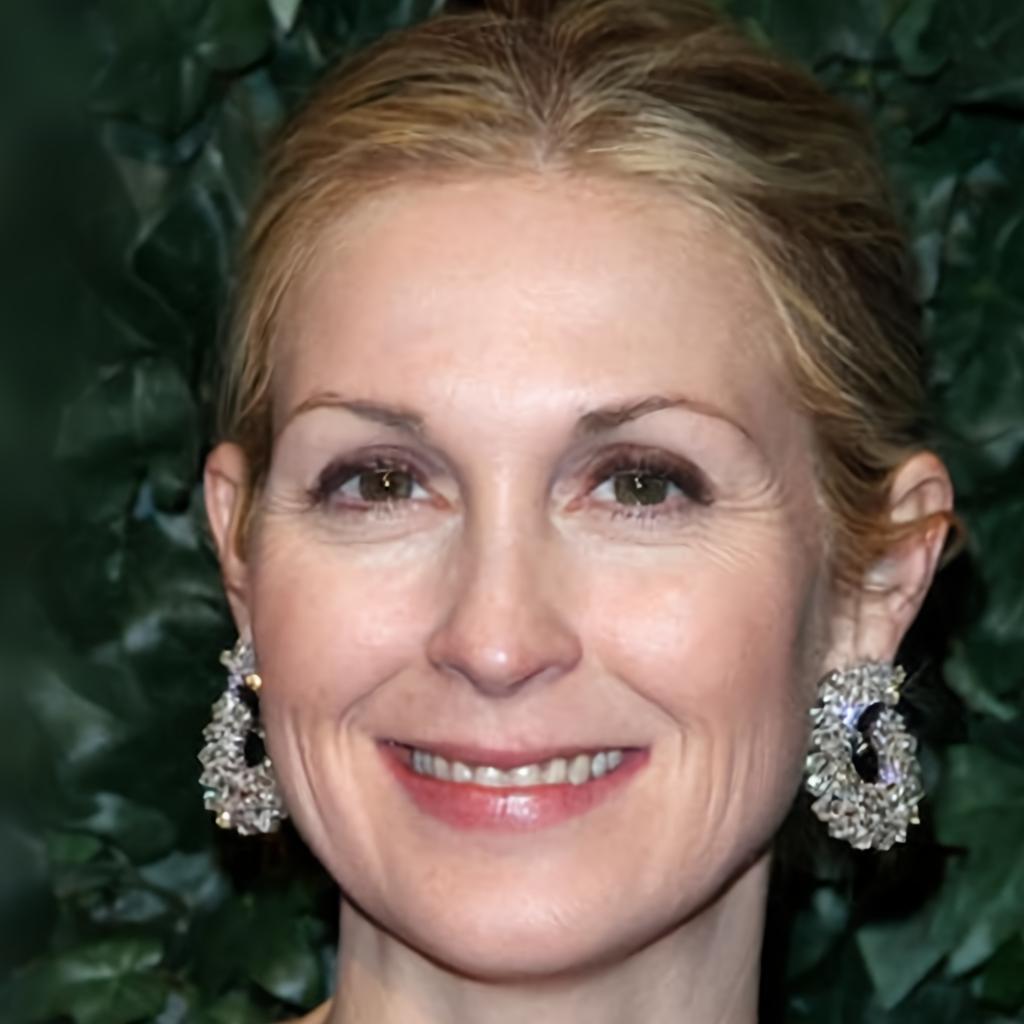}
      \end{subfigure}
      \begin{subfigure}[b]{\textwidth}
        \includegraphics[width=\textwidth]{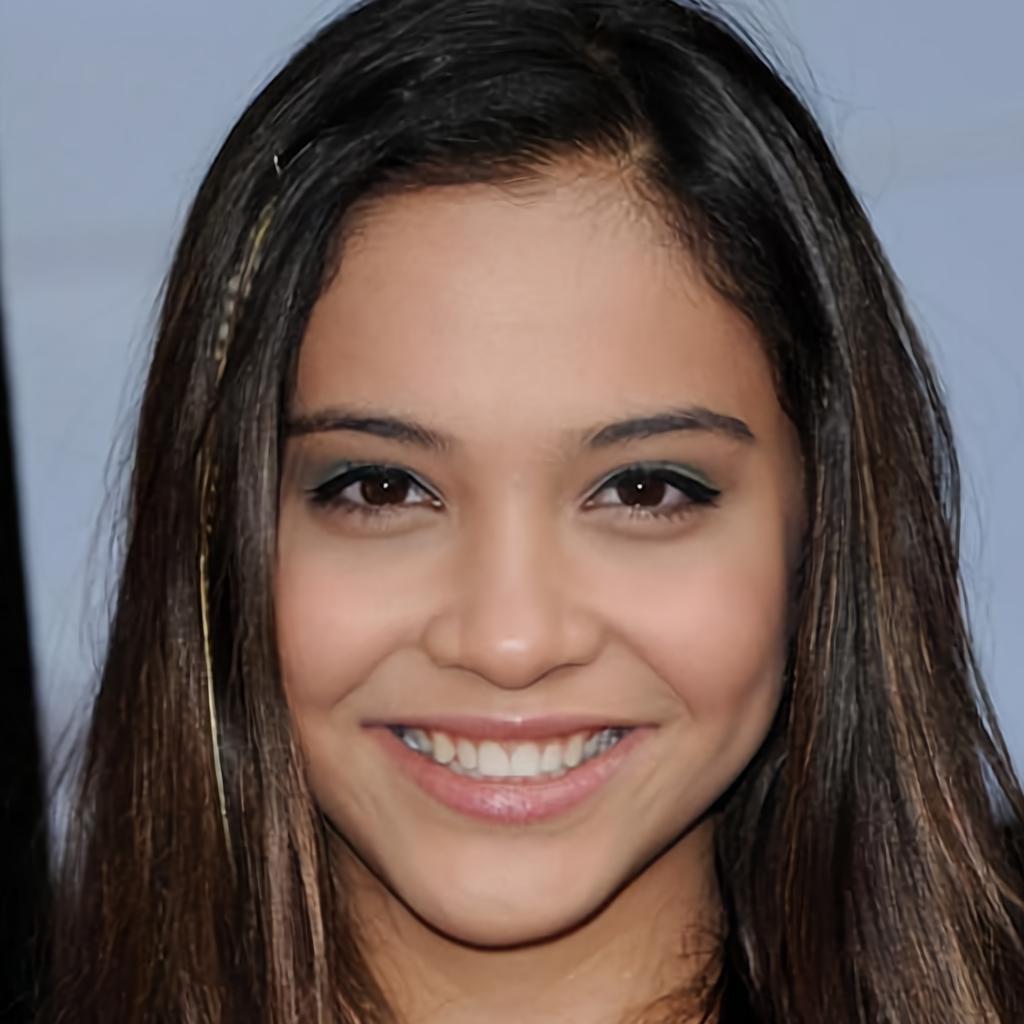}
      \end{subfigure}
      \begin{subfigure}[t]{\textwidth}
        \includegraphics[width=\textwidth]{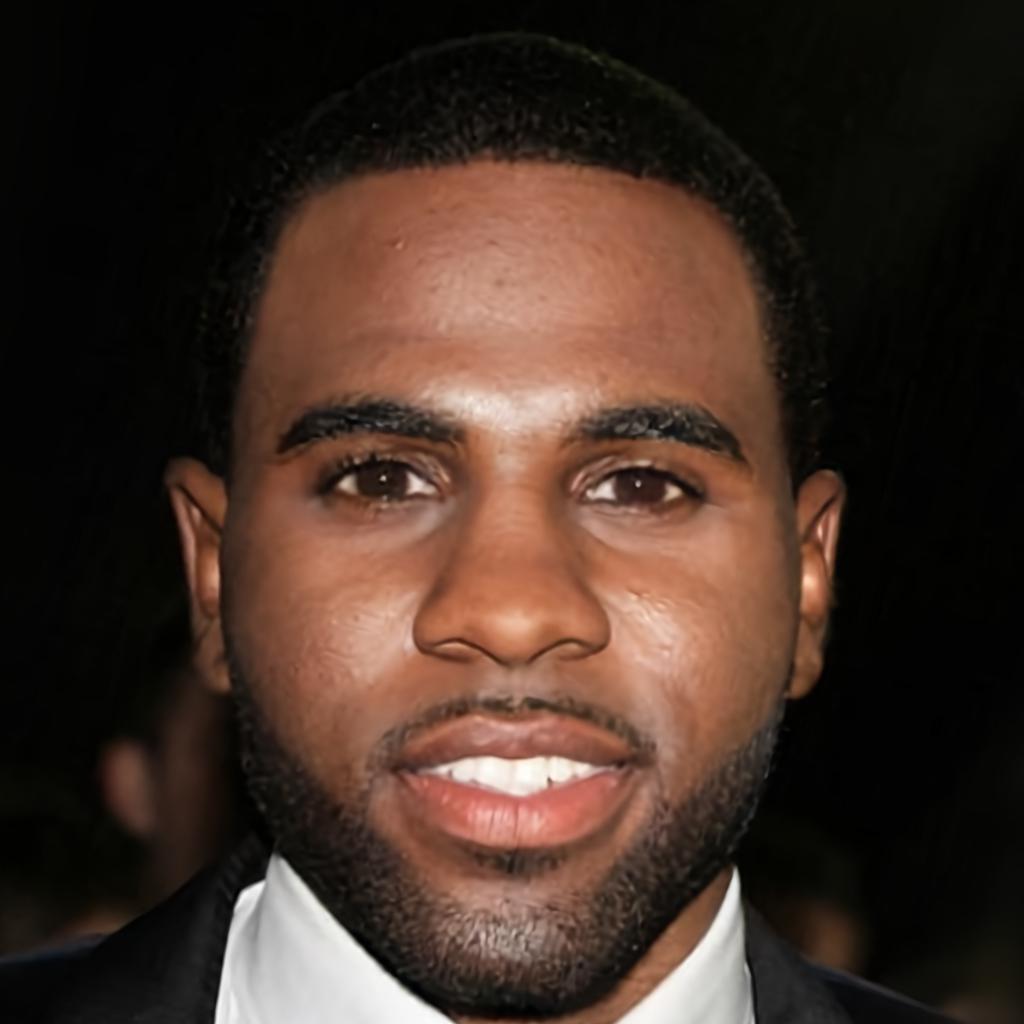}
      \end{subfigure}
      \begin{subfigure}[b]{\textwidth}
        \includegraphics[width=\textwidth]{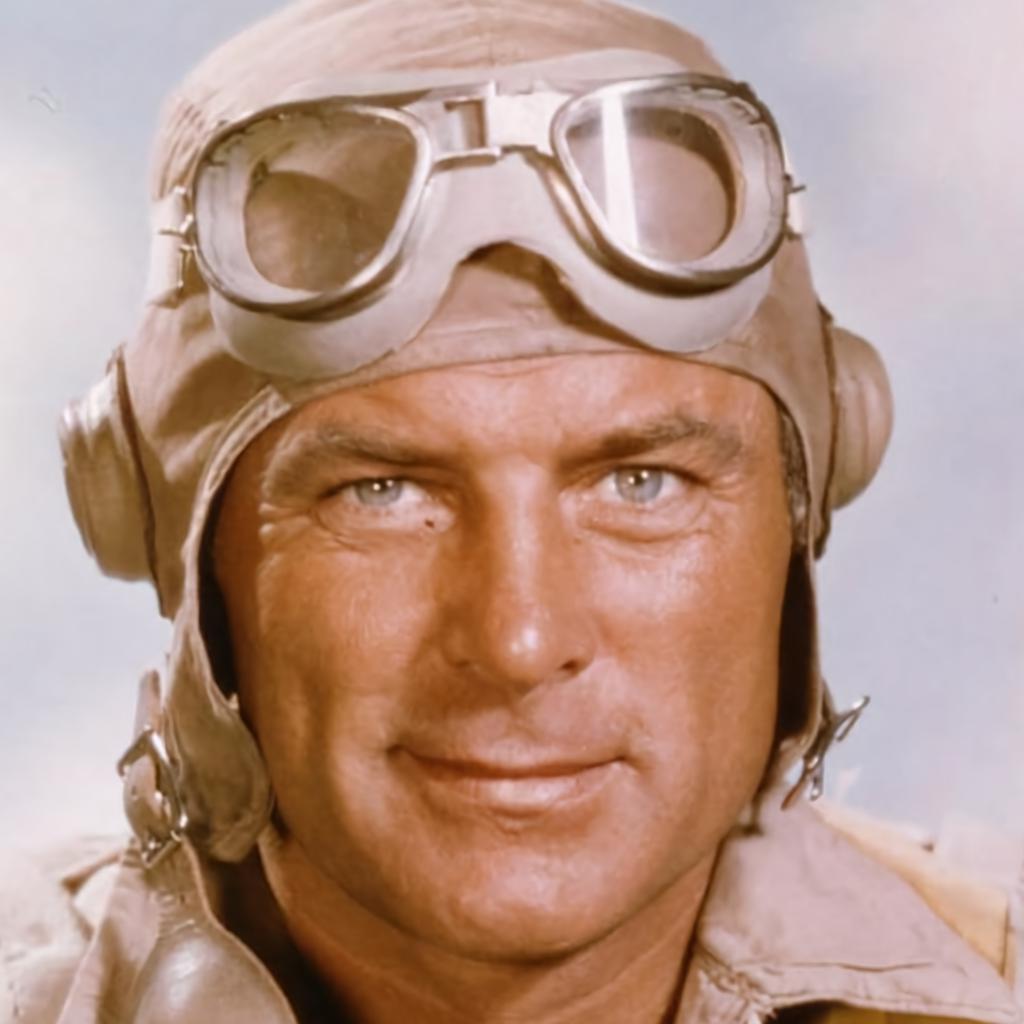}
        \caption{Source / Target}
      \end{subfigure}
    \end{subfigure}
    \begin{subfigure}[b]{0.215\textwidth}
      \begin{subfigure}[t]{\textwidth}
        \includegraphics[width=\textwidth]{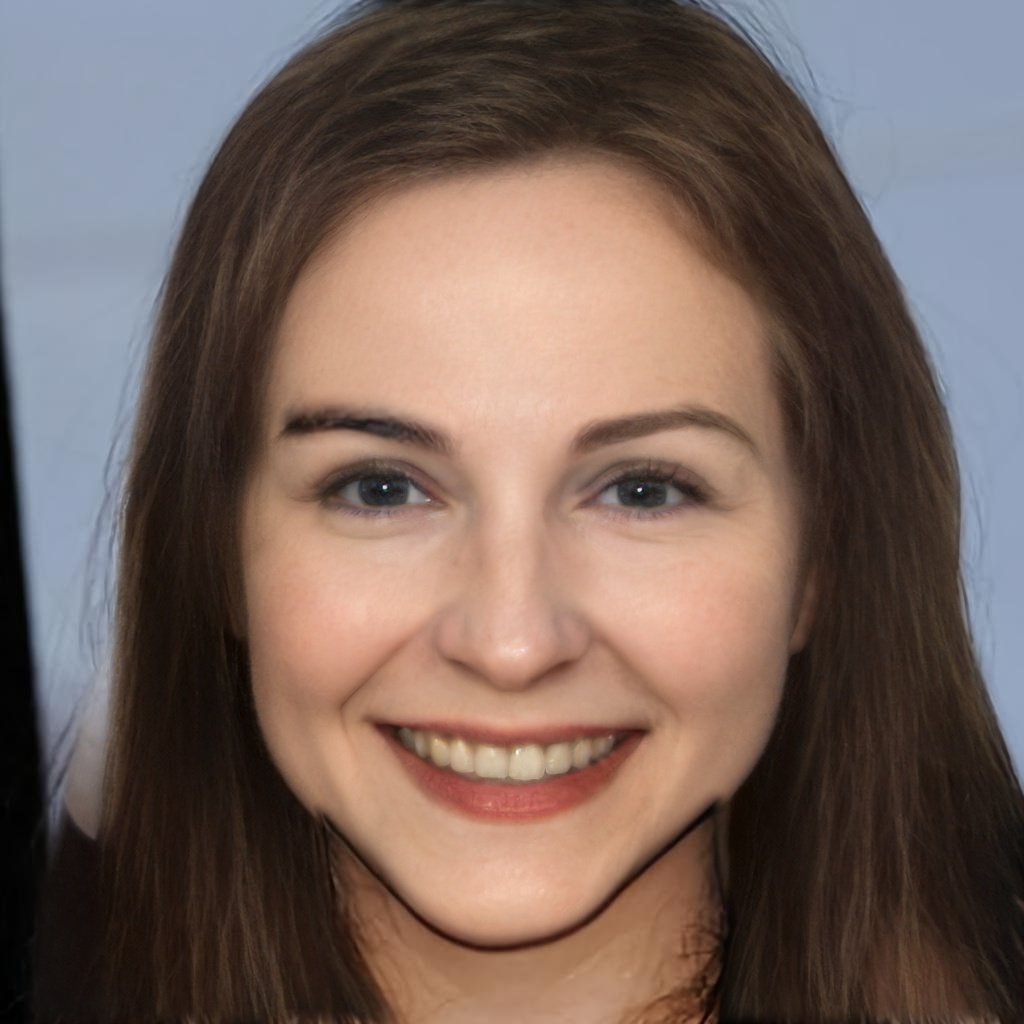}
      \end{subfigure}
      \begin{subfigure}[b]{\textwidth}
        \includegraphics[width=\textwidth]{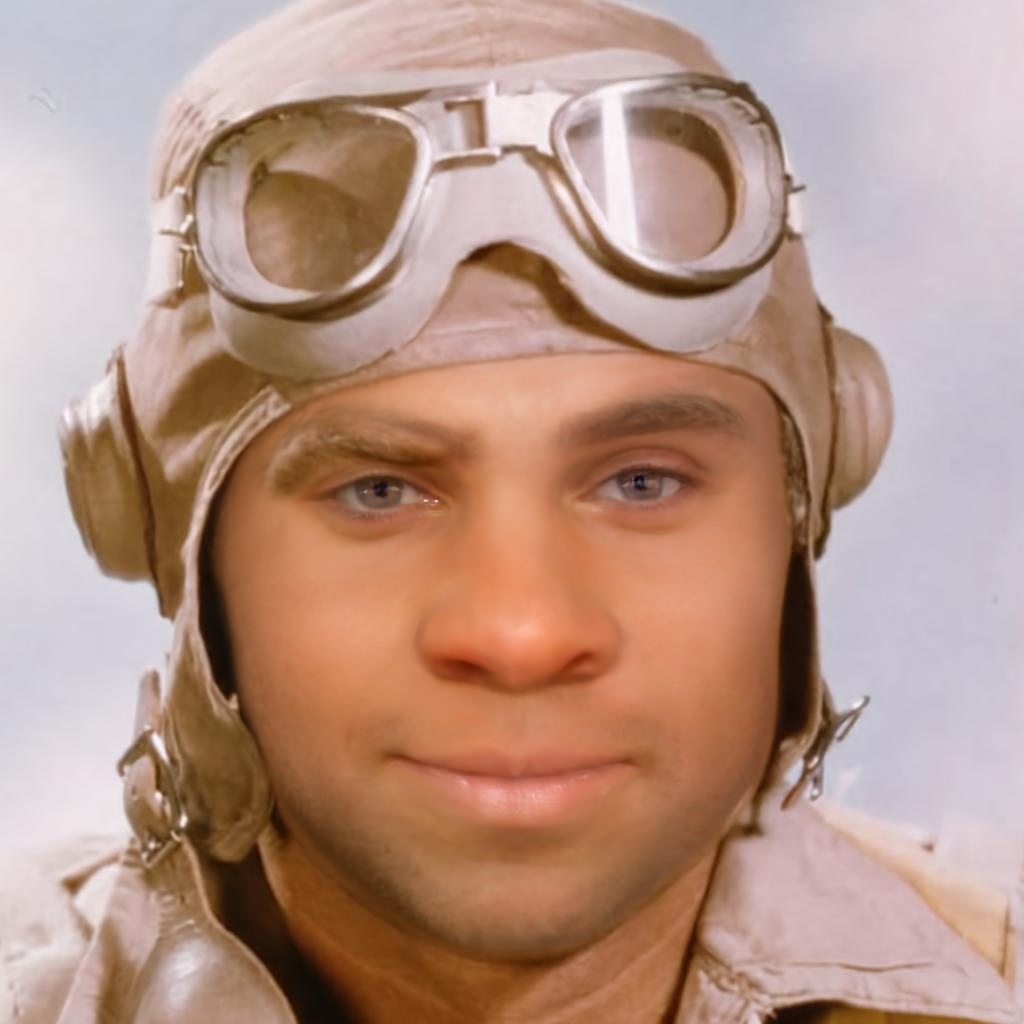}
        \caption{MegaFS~\cite{zhu2021MegaFS}}
      \end{subfigure}
    \end{subfigure}
    \begin{subfigure}[b]{0.215\textwidth}
      \begin{subfigure}[t]{\textwidth}
        \includegraphics[width=\textwidth]{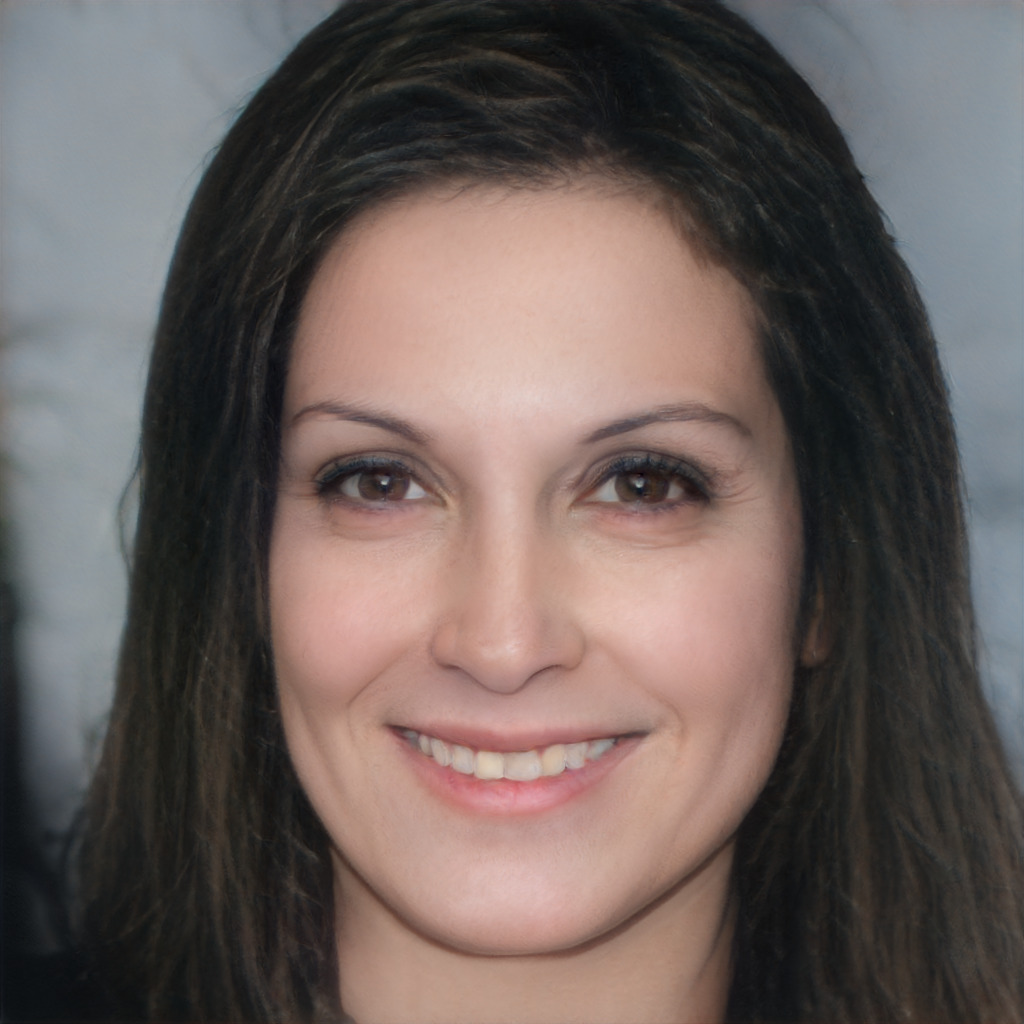}
      \end{subfigure}
      \begin{subfigure}[b]{\textwidth}
        \includegraphics[width=\textwidth]{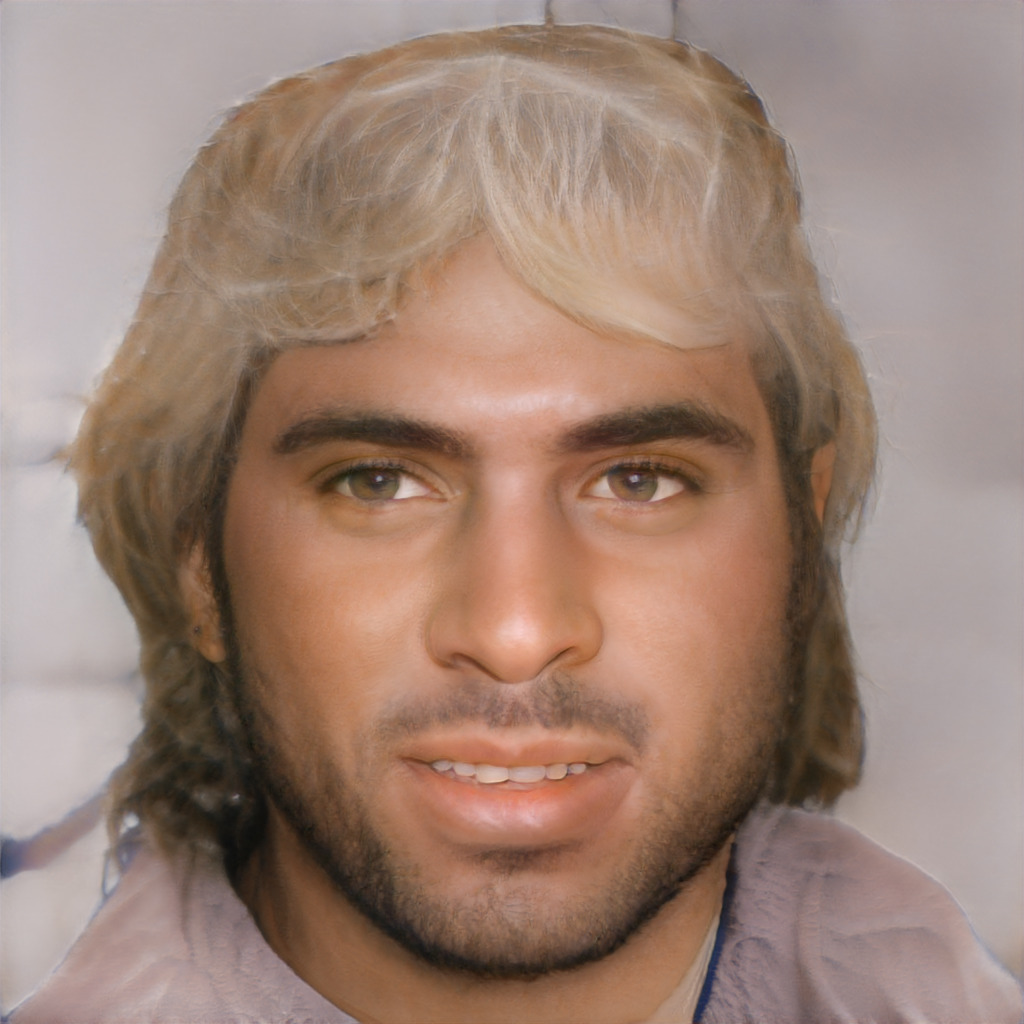}
        \caption{StyleFusion~\cite{kafri2021stylefusion}}
      \end{subfigure}
    \end{subfigure}
    \begin{subfigure}[b]{0.215\textwidth}
      \begin{subfigure}[t]{\textwidth}
        \includegraphics[width=\textwidth]{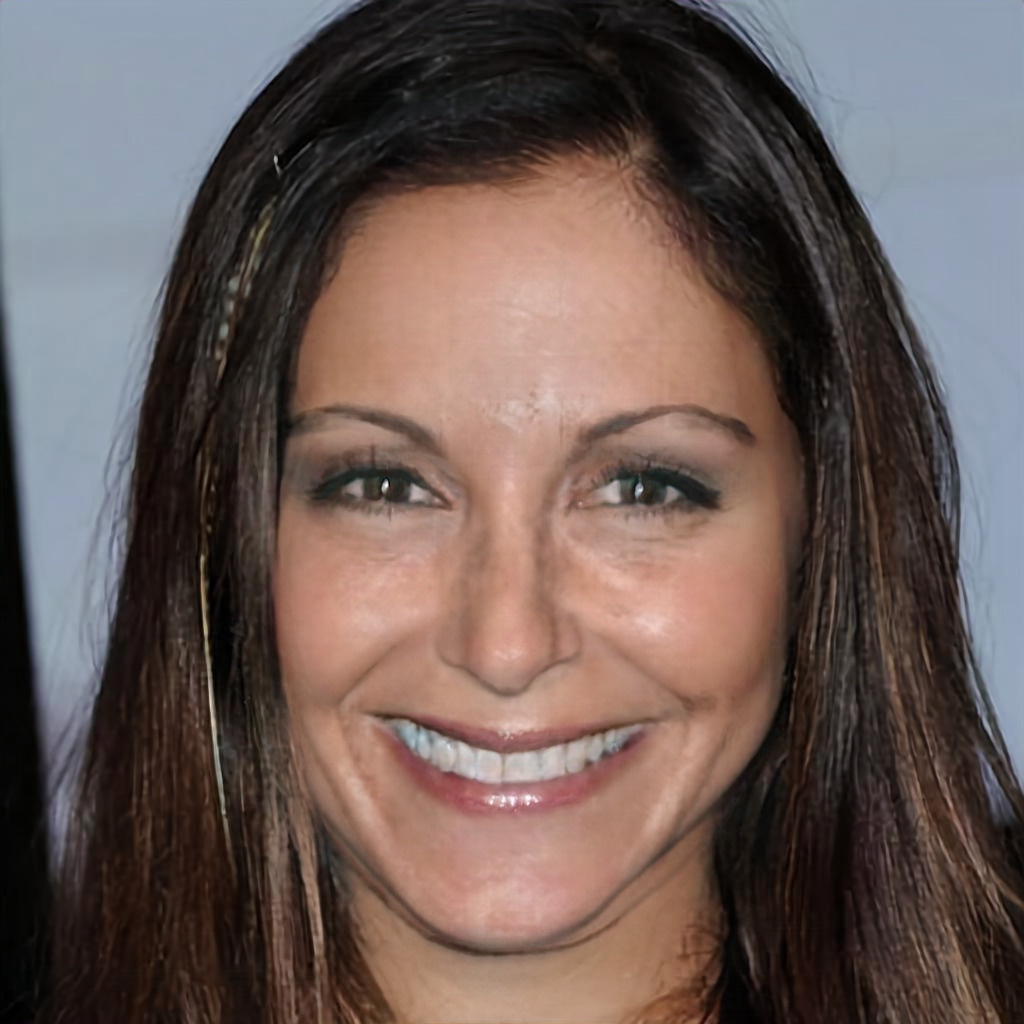}
      \end{subfigure}
      \begin{subfigure}[b]{\textwidth}
        \includegraphics[width=\textwidth]{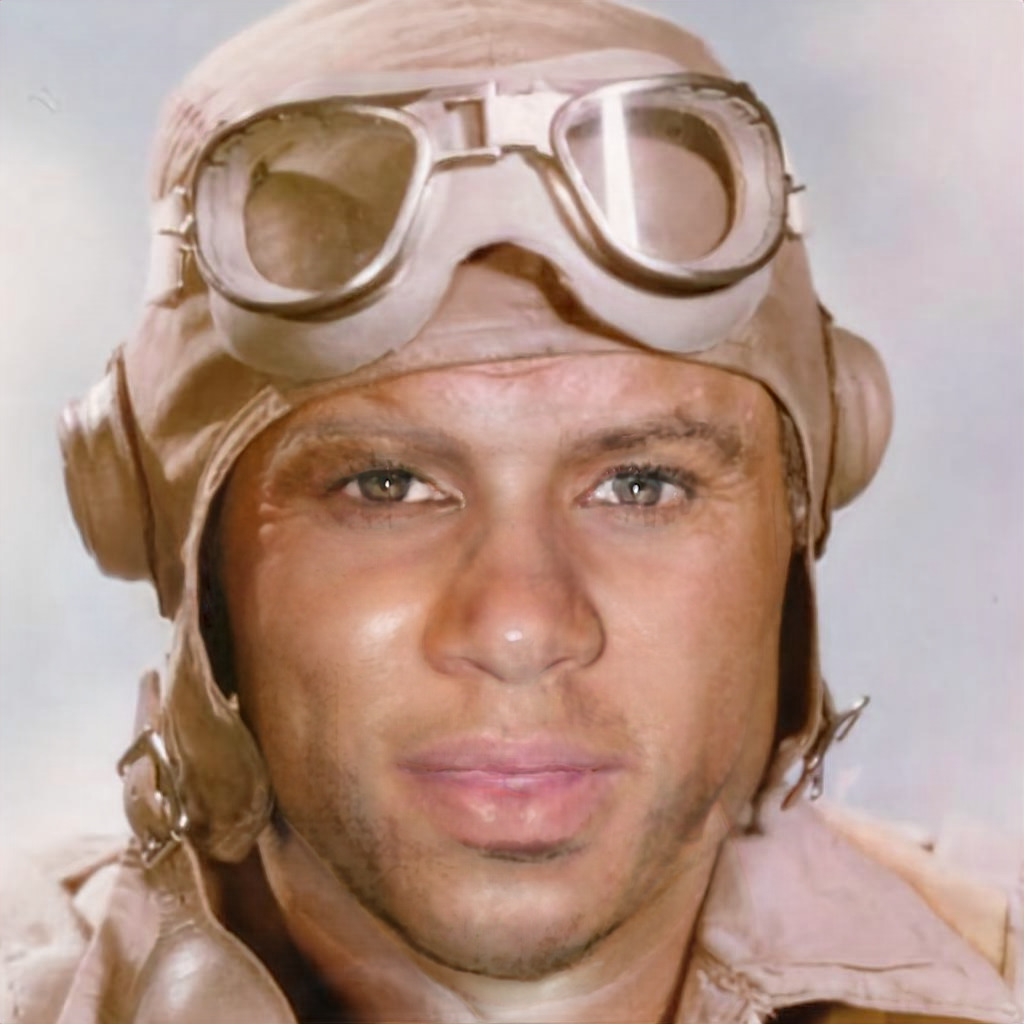}
        \caption{HiRes~\cite{xu2022high}}
      \end{subfigure}
    \end{subfigure}
    \begin{subfigure}[b]{0.215\textwidth}
      \begin{subfigure}[t]{\textwidth}
        \includegraphics[width=\textwidth]{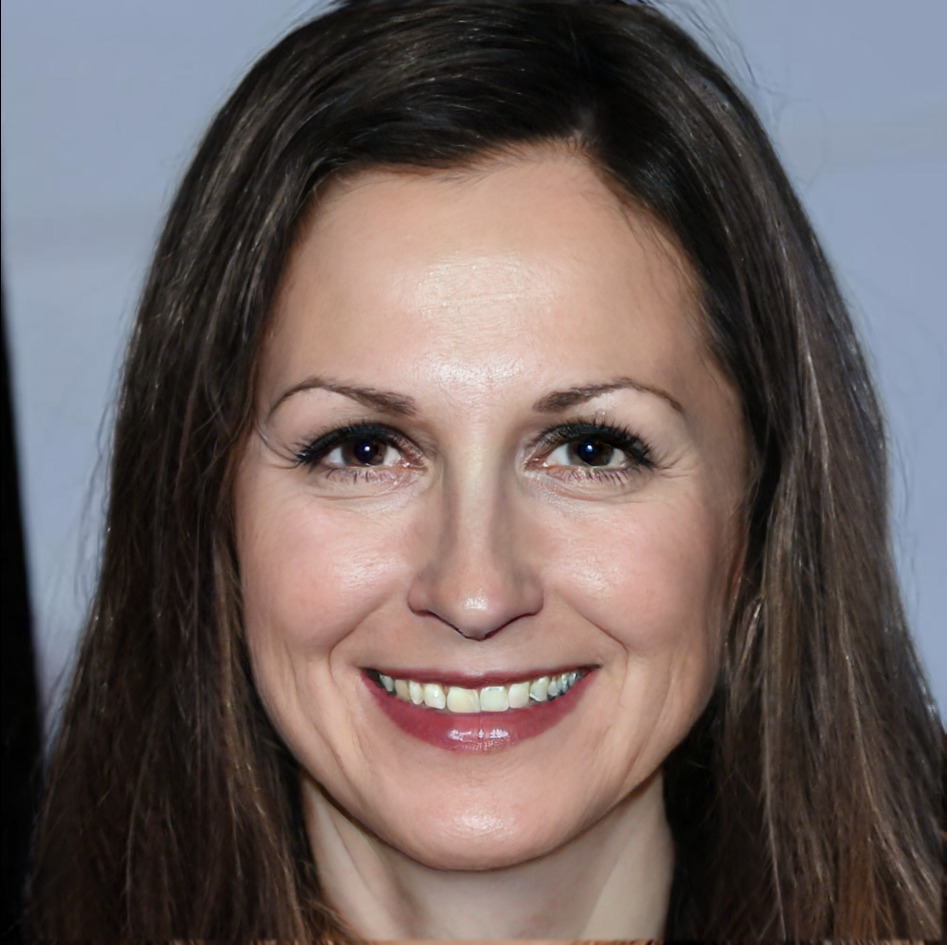}
      \end{subfigure}
      \begin{subfigure}[b]{\textwidth}
        \includegraphics[width=\textwidth]{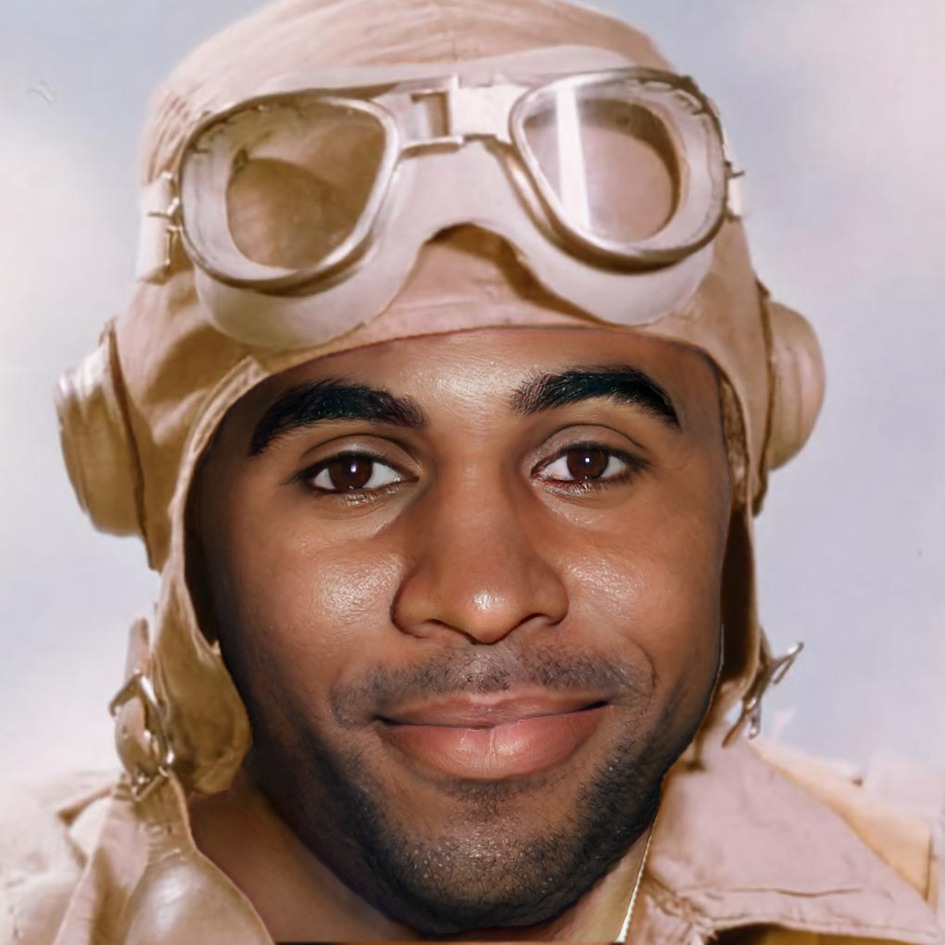}
        \caption{Ours}
      \end{subfigure}
    \end{subfigure}
    \vspace{-2mm}
    \caption{Compared with the existing StyleGAN-based face swapping approaches~\cite{zhu2021MegaFS, kafri2021stylefusion, xu2022high}, our proposed method can achieve high-fidelity results that show better identity keeping from the source while keeping the similar pose and expression as the target. Note that skin color preservation and proper occlusion handling are our advantages over others. All the facial images are at 1024$\times$1024.}

    \label{fig:faceSwappingDemos}
\end{center}%

}]

\let\thefootnote\relax\footnotetext{Work done when Zhian Liu was an intern at Tencent AI Lab}
\let\thefootnote\relax\footnotetext{$\dag$ Equal contribution.}
\let\thefootnote\relax\footnotetext{$*$ Corresponding author: nieyongwei@scut.edu.cn}

\begin{abstract}
\vspace{-0.2cm}
We present a novel paradigm for high-fidelity face swapping that faithfully preserves the desired subtle geometry and texture details. We rethink face swapping from the perspective of fine-grained face editing, \textit{i.e., ``editing for swapping'' (E4S)}, and propose a framework that is based on the explicit disentanglement of the shape and texture of facial components. Following the E4S principle, our framework enables both global and local swapping of facial features, as well as controlling the amount of partial swapping specified by the user. Furthermore, the E4S paradigm is inherently capable of handling facial occlusions by means of facial masks. 
At the core of our system lies a novel Regional GAN Inversion (RGI) method, which allows the explicit disentanglement of shape and texture. It also allows face swapping to be performed in the latent space of StyleGAN. Specifically, we design a multi-scale mask-guided encoder to project the texture of each facial component into regional style codes. We also design a mask-guided injection module to manipulate the feature maps with the style codes. Based on the disentanglement, face swapping is reformulated as a simplified problem of style and mask swapping. Extensive experiments and comparisons with current state-of-the-art methods demonstrate the superiority of our approach in preserving texture and shape details, as well as working with high resolution images. The project page is \url{https://e4s2022.github.io}
\end{abstract}

\vspace{-0.3cm}
\section{Introduction}
Face swapping aims at transferring the identity information (\eg, shape and texture of facial components) of a source face to a given target face, while retaining the identity-irrelevant attribute information of the target (\eg, expression, head pose, background, etc.).  
It has immense potential applications in the entertainment and film production industry, and thus has drawn considerable attention in the field of computer vision and graphics.

The first and foremost challenge in face swapping is \textbf{identity preservation}, \ie, how to faithfully preserve the unique facial characteristics of the source image.
Most existing methods~\cite{chen2020simswap, li2019faceshifter, wang2021hififace} rely on a pre-trained 2D face recognition network~\cite{deng2019arcface}  or a 3D morphable face model (3DMM)~\cite{blanz19993dmm, deng2019accurate} to extract the global identity-related features, which are then injected into the face generation process. 
However, these face models are mainly designed for classification rather than generation, thus some informative and important visual details related to facial identity may not be captured.   
Furthermore, the 3D face model built from a single input image can hardly meet the requirement of robust and accurate facial shape recovery. 
Consequently, results from previous methods often exhibit the ``in-between effect'': \ie, the swapped face resembles both the source and the target faces, which looks like a third person instead of faithfully preserving the source identity. A related problem is \textbf{skin color}, where we argue that skin color is sometimes an important aspect of the source identity and should be preserved, while previous methods 
will always maintain the skin color of the target face, resulting in unrealistic results when swapping faces with distinct skin tones. 

Another challenge is how to properly handle \textbf{facial occlusion.} 
In real applications, for example, it is a common situation that some face regions are occluded by hair in the input images. 
An ideal swapped result should maintain the hair from the target, meaning that the occluded part should be recovered in the source image. 
To handle occlusion, FSGAN~\cite{nirkin2019fsgan} designs an inpainting sub-network to estimate the missing pixels of the source, but their inpainted faces are blurry. A refinement network is carefully designed in FaceShifter~\cite{li2019faceshifter} to maintain the occluded region in the target; however, the refinement network may bring back some identity information of the target.

To address the above challenges more effectively, we rethink face swapping from a new perspective of fine-grained face editing, \ie, \textit{``editing for swapping'' (E4S)}. 
Given that both the shape and texture of individual facial components are correlated with facial identity, we consider to disentangle shape and texture explicitly for better identity preservation.  
Instead of using a face recognition model or 3DMMs to extract global identity features, inspired by fine-grained face editing~\cite{lee2020maskgan}, we exploit component masks for local feature extraction.    
With such disentanglement, face swapping can be transformed as the replacement of local shape and texture between two given faces.
The locally-recomposed shapes and textures are then fed into a mask-guided generator to synthesize the final result. 
One additional advantage of our \textit{E4S} framework is that the occlusion challenge can be naturally handled by the masks, as the current face parsing network~\cite{yu2021bisenetv2} can provide the {pixel-wise} label of each face region. The generator can fill out the missing pixels with the swapped texture features adaptively according to those labels. It requires no additional effort to design a dedicated module as in previous methods ~\cite{li2019faceshifter, nirkin2019fsgan}.  

The key to our \textit{E4S} is the disentanglement of shape and texture of facial components. 
Recently, StyleGAN~\cite{karras2020styleGAN2} has been applied to various image synthesis tasks due to its amazing performance on high-quality image generation, which inspires us to exploit a pre-trained StyleGAN for the disentanglement.
This is an ambitious goal because current GAN inversion methods~\cite{shen2020interfacegan, richardson2021psp, wang2021HFGI} only focus on global attribute editing (age, gender, expression, etc.) in the global style space of StyleGAN, and provide no mechanism for local shape and texture editing. 

To solve this, we propose a novel Regional GAN Inversion (RGI) method that resides in a new regional-wise $\mathcal{W}^{+}$ space, denoted as $\mathcal{W}^{r+}$. 
Specifically, we design a mask-guided multi-scale encoder to project an input face into the style space of StyleGAN. 
Each facial component has a set of style codes for different layers of the StyleGAN generator. 
We also design a mask-guided injection module that uses the style codes to manipulate the feature maps in the generator according to the given masks.  
In this way, the shape and texture of each facial component are fully disentangled, where
the texture is represented by the style codes while the shape is by the masks. 
Moreover, this new inversion latent space supports the editing of each individual face component in shape and texture, enabling various applications such as face beautification, hairstyle transfer, and controlling the swapping extent of face swapping.
To sum up, our contributions are:
\vspace{-0.2cm}
\begin{itemize}
    \item We tackle face swapping from a new perspective of fine-grained editing, \ie, \textit{editing for swapping},  and propose a novel framework for high-fidelity face swapping with identity preservation and occlusion handling. 
    \vspace{-0.5cm}
    \item We propose a StyleGAN-based Regional GAN Inversion (RGI) method that resides in a novel $\mathcal{W}^{r+}$ space, for the explicit disentanglement of shape and texture. It simplifies face swapping as the swapping of the corresponding style codes and masks. 
   \vspace{-0.2cm} 
    \item The extensive experiments on face swapping, face editing, and other extension tasks demonstrate the effectiveness of our E4S framework and RGI.
\end{itemize}

\section{Related Work}
\noindent{\textbf{GAN Inversion }}aims to map an image to its corresponding GAN latent code that can reconstruct the input as faithfully as possible. In this way, one can send the  inverted-then-edited code to the generator to complete the expected editing. A number of StyleGAN inversion approaches have been proposed for face manipulation. Generally, they can be classified into three categories: (1) learning-based~\cite{richardson2021psp, tov2021e4e, alaluf2021restyle, wang2021HFGI, yao2022FSspace, yao2021latent}, (2) optimization-based~\cite{abdal2019image2stylegan, abdal2020image2stylegan++, kang2021GANforOORimages, saha2021loho, zhu2021barbershop} and (3) hybrid methods~\cite{zhu2020indomainGAN}. 
The learning-based methods train an encoder to map the image to the latent space. In contrast, the optimization-based methods directly optimize the latent code to minimize the reconstruction error of the given image. The optimization-based approaches usually give better inversion performance but the learning-based methods cost less time. The hybrid methods make a trade-off between the above two, and use the inverted code as the starting point to conduct further optimization. 
Although specific face editing can be achieved by using existing GAN inversion methods, they work in a global fashion (e.g., growing old, pose changing, male to female) and cannot make precise control of the local facial component.
Our Regional GAN Inversion fills in the gap of high-fidelity local editing via a {novel $\mathcal{W}^{r+}$} latent space based on a pre-trained StyleGAN.

\noindent{\textbf{Face Swapping.}}
The existing face swapping approaches can be generally classified into two categories~\cite{chen2020simswap}, \ie, source-oriented and target-oriented. 
{The source-oriented approaches}~\cite{blanz2004exchangingFace,bitouk2008faceSwapping,nirkin2018OnFaceSeg,nirkin2019fsgan,nirkin2022fsganv2} start from the source and manage to transfer the attributes of the target to the source. The early methods in this camp can date back to~\cite{blanz2004exchangingFace}, where 3D shape and relevant scene parameters were estimated to align pose and lighting.
Then, \cite{nirkin2018OnFaceSeg} claimed that 3D shape estimation is unnecessary and proposed a face segmentation network to fulfill face swapping. 
Recently, a two-stage pipeline was introduced in FSGAN~\cite{nirkin2022fsganv2,nirkin2019fsgan}, where a reenactment and an inpainting network tackle pose aligning and occlusion problems respectively.
{The target-oriented approaches} 
~\cite{korshunova2017fastFaceSwap,bao2018IPGAN,chen2020simswap,li2019faceshifter,wang2021hififace,xu2022region,kim2022smoothswap} begin with the target and tend to transport the identity from the source. 
Generally, these technologies preserve the identity of the source by using a pre-trained face recognition model or 3DMMs. 
As the recognition model is trained for classification and 3DMMs are not accurate and robust, identity-related details cannot fully be captured for generation, bringing the ``in-between effect''. 

As for StyleGAN-based face swapping,
MegaFS~\cite{zhu2021MegaFS} applies the prior knowledge of pre-trained StyleGAN, raising the image resolution to $1024^2$. 
StyleFusion~\cite{kafri2021stylefusion} operates the latent fusion within the $\mathcal{S}$ space~\cite{collins2020editingInStyle,chong2021retrieveInStyle}, enabling controllable generation of local semantic region. However, the shape and texture of each facial region are still entangled in the $\mathcal{S}$ space. Beyond the global latent fusion, ~\cite{xu2022region} designs a region-aware projector to transfer source identity to the target face adaptively.
HiRes~\cite{xu2022high} employs an additional encoder-decoder for target features aggregation in a multi-scale manner. However, fine-grained and selective swapping is not supported in these two methods.

Our \textit{E4S} belongs to the source-oriented camp. 
Inspired by mask-guided face editing~\cite{park2019semantic,lee2020maskgan,zhu2020sean,chen2021sofgan}, we rethink face swapping from the perspective of face editing and treat it as editing of shape and texture for all facial components, \ie, fine-grained face swapping. 
We propose to explicitly disentangle the shape and texture of facial components for better identity preservation based on the proposed RGI method, rather than using a face recognition model or 3DMMs.

\begin{figure*}[t]
  \centering
\vspace{-1.0cm}
\includegraphics[width=0.93\linewidth]{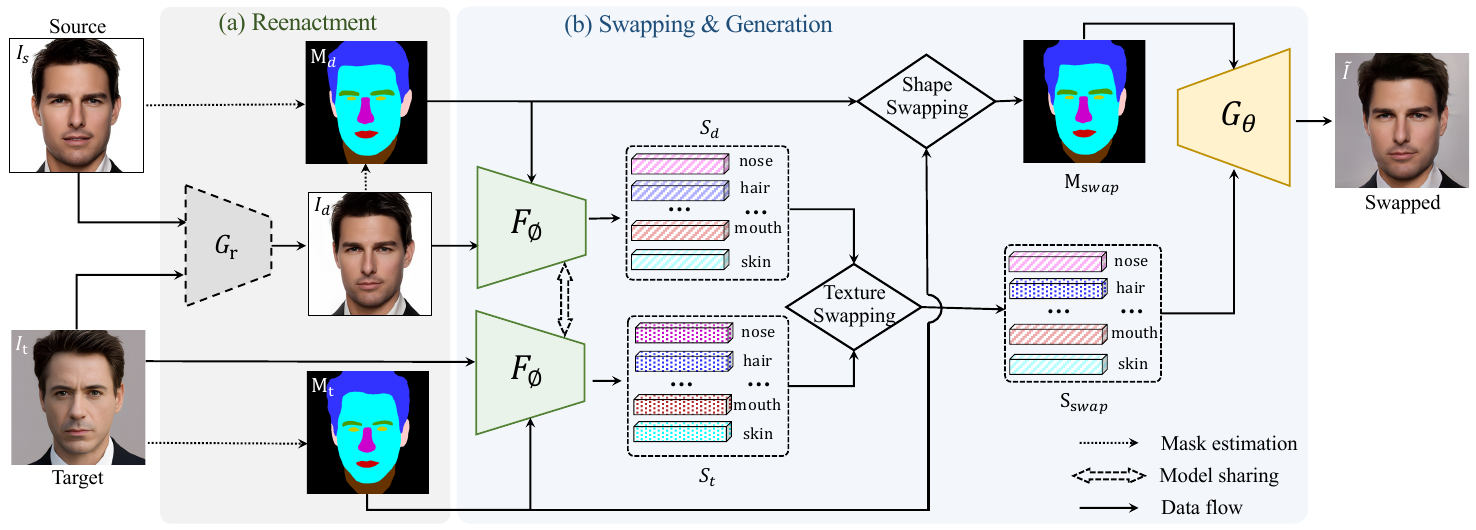}
\captionsetup{belowskip=-10pt}
\caption{Overview of our proposed E4S framework. (a) For the source image $I_s$ and the target $I_t$, a reenactment network $G_r$ is used to drive $I_s$ to show similar pose and expression towards $I_t$, obtaining $I_d$. The segmentation masks of $I_t$ and $I_d$ are also estimated. (b) The driven and target pairs $(I_d, M_d)$ and $(I_t, M_t)$ are fed into the mask-guided encoder $F_{\phi}$ to extract the per-region style codes to depict the texture respectively, producing texture codes $S_d$ and $S_t$. We then swap the masks and the corresponding texture codes, and send them to the pre-trained StyleGAN generator $G_{\theta}$ with a mask-guided injection module to synthesize the swapped face $\tilde{I}$. 
}
  \label{fig:pipelineSwappingFace}
\end{figure*}

\vspace{-0.4cm}
\section{Methodology}
\subsection{Editing-for-Swapping (E4S) Framework}
\label{sec:swappingFace}
Our \textit{E4S} framework mainly consists of two phases inside: (a) reenactment, and (b) swapping and generation, where the overall pipeline is illustrated in Fig.~\ref{fig:pipelineSwappingFace}.

\noindent{\textbf{Reenactment.}}
We first crop the face region of the source image and target image, obtaining the cropped faces $I_s$ and $I_t$.
 Then, we use the dlib~\cite{dlib09} toolbox to crop the face region and detect the facial landmarks. Next, we follow the original StyleGAN~\cite{karras2019styleGAN} to align the cropped face and resize it to the resolution of 1024$\times$1024.

In order to drive $I_s$ to reach a similar pose and expression as $I_t$, we employ a pre-trained face reenactment model FaceVid2Vid~\cite{wang2021faceVid2Vid}, resulting in a driven face $I_d$. 
Such a face reenactment processing can be described as: $I_d = G_r(I_s, I_t)$,
where $G_r$ denotes the FaceVid2Vid model. 
Further, we estimate the segmentation masks $M_t$ of the target face $I_t$ and $ M_d$ of the driven face $I_d$, thus obtaining the target and driven pairs ($I_t, M_t$) and ($I_d, M_d$). 
For face parsing, we utilize an off-the-shell face parser~\cite{zllrunning2013faceParser}, where each segmentation mask belongs to one of the 19 semantic categories. 
For simplicity, we aggregate the categories of symmetric facial components, resulting in 12 categories, \ie, \textit{background, eyebrows, eyes, nose, mouth, lips, face skin, neck, hair, ears, eyeglass, and ear rings}.
\noindent{\textbf{Swapping and Generation.}}
\label{subsec:faceSwapping}
In this phase, we would elaborate on the face swapping process in our E4S. We first feed the driven pair ($I_d, M_d$) and the target pair ($I_t, M_t$) into a mask-guided multi-scale encoder $F_{\phi}$ respectively, which extracts the style codes to represent the texture of each facial region. 
This step can be summarized as:
\vspace{-2mm}
\begin{equation}
    S_t = F_{\phi}(I_t, M_t),\quad S_d = F_{\phi}(I_d, M_d),
\vspace{-1mm}
\end{equation}
where $S_t$ and $S_d$ denote the extracted texture codes of the target and driven face, respectively. 
The detailed modules of the encoder $F_{\phi}$ are introduced in~\cref{sec:regionalGANInversion}. 
Then, we exchange the texture codes of several facial components of $S_t$ with those of $S_d$, obtaining the recomposed texture codes $S_{swap}$. 
Here, the swapped components are: \textit{eyebrows, eyes, nose, mouth, lips, face skin, neck, and ears}. 
Note that the skin is carefully considered here since it is identity-related, while it is neglected by most existing works.

In addition to texture swapping, shape swapping is also required to realize the aim of face swapping. 
Since the shape is represented by facial masks, we start with an empty mask $M_{\text{swap}}$ as a canvas and then complete the mask recomposition in the following steps. 
First, we keep the \textit{neck} and the \textit{background} layout of the target mask $M_t$ and stitch their masks onto $M_{\text{swap}}$. 
Then, we stitch the inner face regions of the driven mask $M_d$, including \textit{face skin, eyebrows, eyes, nose, lips, and mouth}. 
Finally, we stitch the \textit{hair, eye glasses, ear, and ear rings} of the target mask $M_t$ onto $M_{\text{swap}}$. 
Note that the driven mask $M_d$ and the target mask $M_t$ may not be aligned with each other perfectly, leading to some missing pixels in the swapped mask, which is always caused by the occlusion. 
We observe that these missing areas are usually between the facial skin and hair or between the facial skin and neck. 
As a solution, we fill up these areas with \textit{face skin}, which is the unique advantage of our method.
Compared with the existing methods FSGAN~\cite{nirkin2019fsgan} or FaceShifter~\cite{li2019faceshifter}, our method \textit{does not} need to train an extra sub-network to deal with the occlusion.
Please refer to our Supp. for more details.

After obtaining the recomposed mask $M_{\text{swap}}$ and texture codes ${S_{\text{swap}}}$, we feed them into the StyleGAN generator $G_{\theta}$ with a mask-guided style injection module to synthesize the swapped face, which can be expressed as $ \tilde{I} = G_{\theta}(M_{\text{swap}}, S_{\text{swap}})$. Here, $G_{\theta}$ will be detailed in~\cref{sec:encodingStyle}. Finally, the swapped face $\tilde{I}$ and target image $T$ are blended together to produce the final swapped image.

\begin{figure}[t]
  \centering
  \captionsetup{belowskip=-13pt}
  \includegraphics[width=1.05\linewidth]{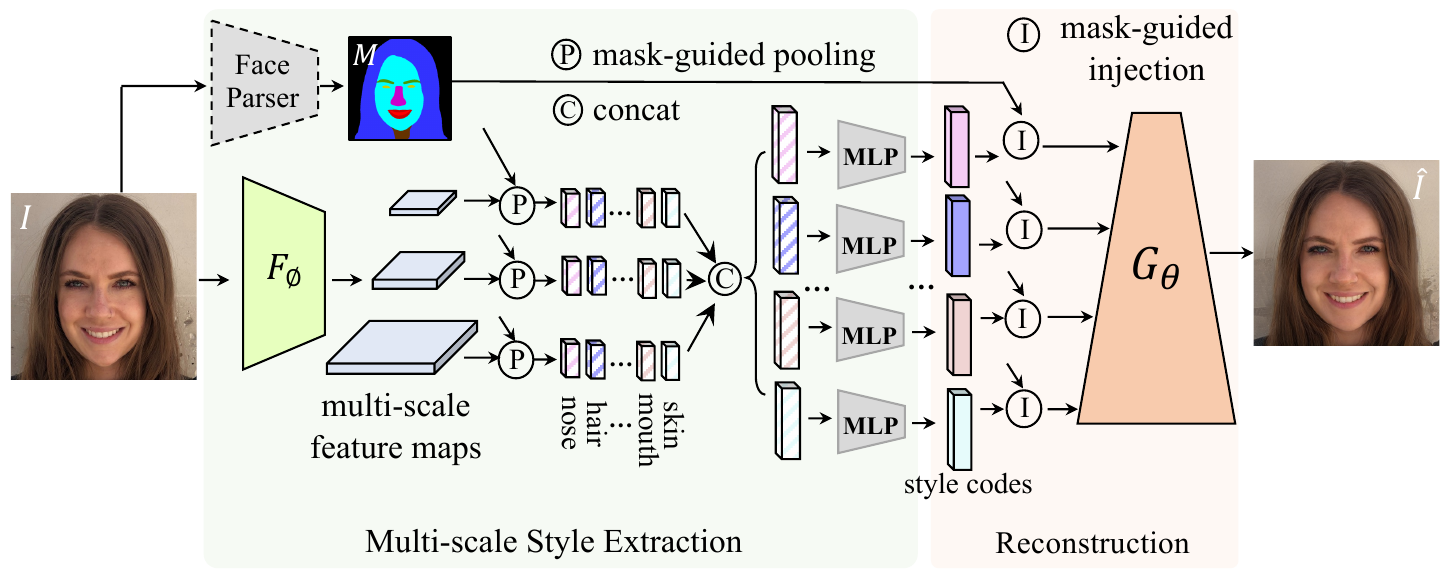}
  \caption{Overview of our proposed RGI. The input face $I$ and its segmentation map $M$ are fed into a multi-scale encoder $F_{\phi}$ to extract the per-region texture vectors. The multi-scale texture vectors are then concatenated and passed through some MLPs to obtain the style codes resident in {a novel $\mathcal{W}^{r+}$} latent space of StyleGAN. The regional style codes and the mask $M$ are used by our mask-guided StyleGAN generator to produce the reconstructed face $\hat{I}$.}
  \label{fig:GANInversionPipeline}
\end{figure}

\vspace{-0.15cm}
\subsection{Disentanglement of Shape and Texture}\label{sec:regionalGANInversion}

The core of our \textit{E4S} framework is how to precisely encode the per-region textures which is disentangled with their shapes. 
Previous mask-guided face editing methods \cite{park2019semantic,lee2020maskgan,zhu2020sean} attempt to use masks as input of a generator and inject the texture style to guide the generation, while they still struggle to
preserve the identity and facial details during editing (see ~\cref{fig:edit}). 
Besides, they have a limited resolution of the generated face, where ~\cite{lee2020maskgan} reaches the resolution of $512^2$ while the rest are with $256^2$. 

To pursue a better disentanglement of shape and texture as well as high-resolution and high-fidelity generation, we resort to the powerful generative model StyleGAN that can generate images with $1024^2$ resolution. 
Instead of training StyleGAN from scratch, we explore the possibility of developing a GAN inversion method. Specifically, we use a pre-trained StyleGAN for {the} disentanglement, avoiding the massive computing resources and training instability.  
Although there are a number of GAN inversion techniques~\cite{richardson2021psp, tov2021e4e, yao2022FSspace} have been proposed for face editing in the $\mathcal{W}$ or $\mathcal{W}^{+}$ space, 
they focus on global facial attribute editing only, \eg, age, pose, and expression.  
Hence, they cannot be applied to the disentanglement of shape and texture for local editing. 
To tackle the shortage, we propose a novel Regional GAN Inversion (RGI) method for such a disentanglement, which incorporates facial masks into the style embedding and the generation process, thus filling in the gap of GAN inversion based local editing. 
The overview of our RGI is illustrated in \cref{fig:GANInversionPipeline}.

\noindent{\textbf{Mask-guided Style Extraction.}}
\label{sec:encodingStyle}
Given an image $I$ and its corresponding segmentation mask $M$, we first feed the image $I$ into a multi-scale encoder $F_{\phi}$ to produce feature maps $[F_1,F_2,...,F_N]$ at different levels, 
where N is the number of scales and $F_{\phi}$ is a convolution network with multiple layers. Then, we can obtain the multi-scale features for each individual facial region based on the feature maps $[F_1,F_2,...,F_N]$ and the mask $M$. 
Specifically, for each feature map $F_i$, we downsize the mask $M$ to the same {spatial size} and apply the average pooling operation on $F_i$ to aggregate features for each facial region as:
\vspace{-1mm}
\begin{equation}\label{eq:avgpooling}
v_{ij} = \text{AVG}(F_i \odot (\lfloor M \rfloor_i == j)), \forall j \in \{1,2,...,C\},
\end{equation}
where $C$ is the number of segmentation categories, $\odot$ is the Hadamard product, and $\lfloor M \rfloor_{i}$ denotes the downsized mask with the same height and width as $F_i$. 
Further, the multi-scale feature vectors $\{v_{ij}\}_{i=1}^{N}$ of region $j$ are concatenated and fed into an MLP to obtain the style codes:
\vspace{-1mm}
\begin{equation}
    s_{j} = \text{MLP}([v_{1j};v_{2j};...;v_{Nj}]),
\end{equation}
where $s_j$ denotes the style codes of the $j$-th facial region. 
Then, the style codes and the mask $M$ are fed into the {StyleGAN} generator to synthesize the swapped face. 
{Here, we denote $s \in \mathbb{R}^{C \times 18 \times 512}$ as the $\mathcal{W}^{r+}$ space.}

\begin{figure}[t]
  \centering
  \vspace{-0.4cm}
  \includegraphics[width=\linewidth]{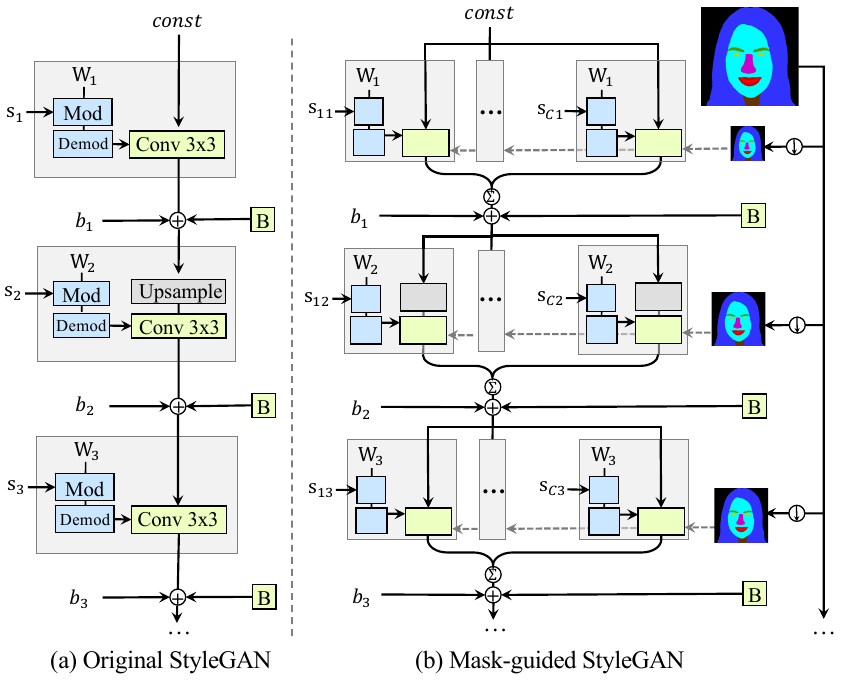}
   \captionsetup{belowskip=-13pt}
   \caption{{The comparison of the original StyleGAN and the proposed mask-guided StyleGAN which regionally extends the style block. We sum up the intermediate feature maps of each region according to its segmentation mask that is downsized in advance.}}
  \label{fig:detailed_archi}
\end{figure}

\noindent{\textbf{Mask-guided Style Injection.}}
\label{sec:localstylegan}
As shown in Fig.~\ref{fig:detailed_archi}(a), the original StyleGAN generator starts from a constant feature map with the spatial size of 4$\times$4 and consists of a serials of style blocks. 
Each style block contains a modulation, a demodulation, and a 3$\times$3 convolution layer. Besides, a noise layer \fbox{B} is introduced to increase the diversity. 
The learnable kernel weights and bias in each block are denoted as $W$ and $b$, respectively.  $W$ will be scaled by its corresponding style code with the shape of $\mathbb{R}^{512}$ before the convolution layer. An additional upsampling layer by the factor of two is employed between every two style blocks to increase the resolution of feature maps.

Different from the style code in the original StyleGAN, which globally controls the appearance of the output image, we propose to extract regional style code that controls only the appearance of the corresponding face component precisely along with its mask, as described above.
To this end, we extend the style block of the original StyleGAN to a mask-guided style block conditioned on a given mask. 
Specifically, we sum up the intermediate feature maps with the guidance of per-region mask, which can be formed as:
\vspace{-2mm}
\begin{gather}
\scalebox{0.85}{$
F_{l} =  \sum_{j=1}^{C}(F_{l-1} * W'_{jl}) \odot (\lfloor M \rfloor_{l} == j),  \forall\  l \in \{1,2,...,K\}
$}, \\
W'_{jl} = Demod(Mod(W_l, s_{jl})) \label{eqn:modulation},
\vspace{-2mm}
\end{gather}
where $F_{l-1}$ and $F_{l}$ denote the input and output feature maps of $l$-th layer, respectively.
$W'_{jl}$ represents the scaled kernel weights for the $j$-th component in the $l$-th layer, and $*$ means the convolution operation. Similar to~\cref{eq:avgpooling}, $\lfloor M \rfloor_{l}$ is the downsized mask corresponds to the $l$-th layer.
We follow the same modulation and demodulation as the original StyleGAN and extend the style modulation regionally. In ~\cref{eqn:modulation}, $W_{l}$ denotes the original kernel weights for the $l$-th layer, and the $s_{jl}$ indicates the style code of $j$-th component for the $l$-th layer. The schematic operations of our proposed mask-guided style injection are illustrated in~\cref{fig:detailed_archi}(b).

Note that the mask is only injected into the first $K$ layers of the StyleGAN. That is, we do not use the mask-guided style block for the last ($18-K$) layers. 
There are two reasons for this occurrence: (1) we conduct experiments with $K=11, 13, 15, 18$ and empirically find the reconstructed images show few visual differences when $K$ is greater than 13; (2) the training overload can be decreased without the mask-guided style block in the last ($18-K$) layers since the resolution of these layers are large (\ie, $512^2-1024^2$). Considering these two factors, we set $K=13$ as the default in all the experiments. 

\vspace{-0.2cm}
\subsection{Training Objective}\label{sec:lossfuc}
During training, we only utilize the reconstruction as the proxy task and \textit{do not} need to swap paired faces like the most existing face swapping methods, which makes our method more efficient and easier to train. Once the training is finished, the texture encoder $F_{\phi}$ can be used to produce per-region texture codes of any input face. One can easily achieve face swapping in the $\mathcal{W}^{r+}$ latent space as described in \cref{sec:swappingFace}. 
We adopt the commonly used loss functions in the GAN inversion literature, which are described in our Supp. in detail.

\vspace{-0.25cm}
\section{Experimental setup}

\noindent{\textbf{Datasets.}}
{\textbf{CelebAMask-HQ}}~\cite{lee2020maskgan} contains 30K high-quality face images, which are split into 28k and 2K for training and testing, respectively. This dataset also provides facial segmentation masks, with 19 semantic categories included. 
{\textbf{FFHQ}}~\cite{karras2019styleGAN} contains 70K high-quality images with a large diversity, but the facial segmentation masks are not officially given. 
We use a pre-trained face parser~\cite{zllrunning2013faceParser} to extract the facial segmentation masks.

\noindent{\textbf{Implementation Details.}}
We use PyTorch~\cite{paszke2019pytorch} to implement our framework, and train our model on 8 NVIDIA A100 GPUs. 
During training, we set the batch size to 2 for each GPU and initialize the learning rate as $10^{-4}$ with the Adam~\cite{kingma2014adam} optimizer ($\beta_1=0.9$, $\beta_2=0.999$). 
For CelebAMask-HQ and FFHQ datasets, we train the model for 200K and 300K iterations, respectively. 
The initial learning rate decays by the factor of $0.1$ at 100K and 150K iterations, respectively. 
Besides, we randomly flip images with a ratio of $0.5$. 
\vspace{-0.25cm}
\section{Ablation study}

\begin{table}[t]
\vspace{-0.2cm}
\caption{Quantitative comparison of our RGI under different ablative configurations. The reconstruction performance is measured.}
\small
\vspace{-0.2cm}
\centering
\begin{tabular}{|l|cccc|}
\hline
\multicolumn{1}{|c|}{\textbf{Configurations}} & \multicolumn{1}{c}{\textbf{SSIM$\uparrow$}} & \multicolumn{1}{c}{\textbf{PSNR$\uparrow$}} & \multicolumn{1}{c}{\textbf{RMSE$\downarrow$}} & \multicolumn{1}{c|}{\textbf{FID$\downarrow$}} \\ \hline
our RGI full model                                  & 0.818                             & 19.851                            & 0.105                             & 15.032                            \\
(A) w/o identity loss                           & 0.819                             & 19.888                            & 0.105                             & 15.141                            \\
(B) w/o finetuning                          & 0.827                             & 19.984                            & 0.104                             & 22.239                            \\ \hline
\end{tabular}
\label{tbl:ablation}
\end{table}
    
\begin{figure}[t]
\centering
\setkeys{Gin}{width=\linewidth}
\captionsetup{belowskip=-15pt}
\begin{subfigure}{\linewidth}
\includegraphics[width=1.0\textwidth]{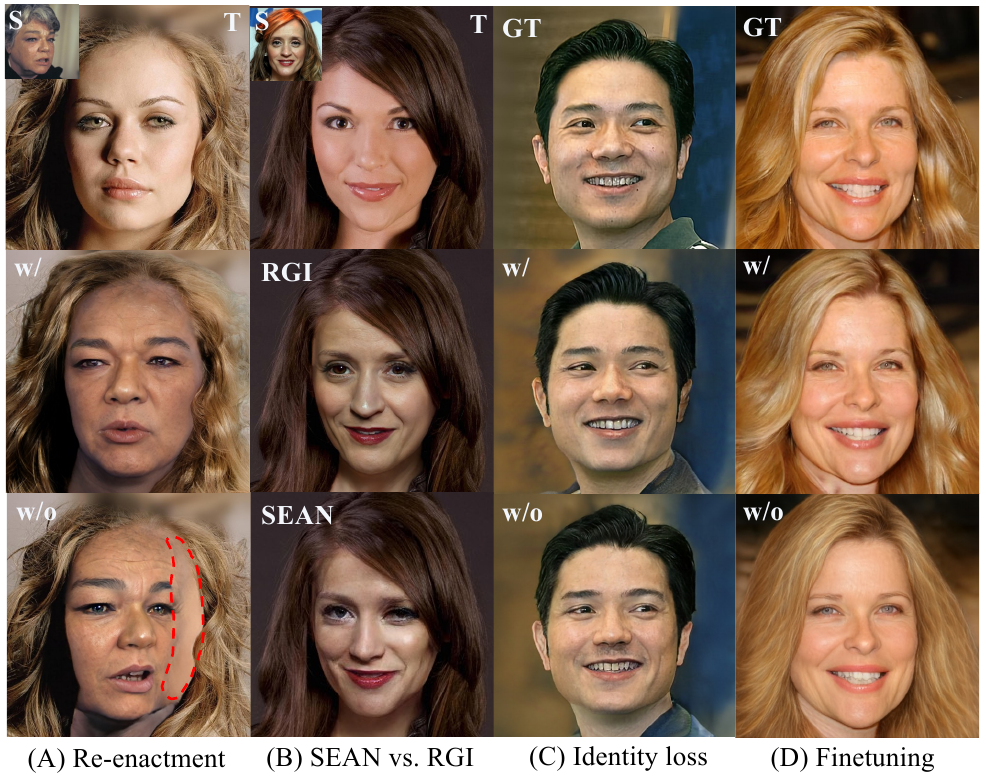}
\end{subfigure}
     \caption{Qualitative comparisons of different ablative settings. }
    \label{fig:ablation}
\end{figure}

In this section, we perform an ablation study to validate the design choices of our proposed \textit{E4S} framework and RGI method.
We show the qualitative comparison in~\cref{fig:ablation} and the quantitative comparison in~\cref{tbl:ablation}, where the reconstruction performance is considered.

\noindent{\textbf{Re-enactment.}} To drive the source to show a similar pose and expression as the target, we employ a pre-trained face reenactment model~\cite{wang2021faceVid2Vid} before the shape and texture swapping procedure.
To verify the necessity of the Re-enactment step in \textit{E4S}, we compare a standard \textit{E4S} pipeline and the one without Re-enactment.
{As shown in the 1st column of \cref{fig:ablation}, the swapped result is not aligned with the target face when the reenactment is disabled (see the circled red region), revealing the pose information is also embedded in the per-region texture represented by the style codes.} 

\noindent{\textbf{SEAN \vs RGI.}} Our \textit{E4S} framework is generic.
Specifically, those methods which contain an encoder extracting the per-region style codes and a generator controlling the per-region style codes along with the segmentation mask, can be adapted to our \textit{E4S} framework. To valid this, we replace our RGI with SEAN~\cite{zhu2020sean} to play the roles of $F_{\phi}$ and $G_{\theta}$ in \cref{fig:pipelineSwappingFace}. From the 2nd column of \cref{fig:ablation}, we can observe that SEAN can produce an overall visually pleasant result while our result preserves more details (the eyes and face skin). Moreover, SEAN only shows the capability of generating faces at $256^2$ while ours are at $1024^2$. This also shows the superiority of our proposed RGI.

\noindent{\textbf{Identity loss.}}
We add an ID loss when training our RGI under the reconstruction setting. From the configuration (A) in~\cref{tbl:ablation}, the performance is comparable to the baseline when we do not apply
the ID loss. However, without the ID loss would lead to some identity information missing, which is confirmed by the 3rd column
in~\cref{fig:ablation}.

\noindent{\textbf{Pre-trained \vs fine-tuned StyleGAN.}} 
Though the pre-trained StyleGAN can be used for face swapping, the hair texture details cannot be always well preserved. 
To achieve a more robust performance on hair, we fine-tune the first $K=13$ layers of the StyleGAN. 
The configuration (B) in~\cref{tbl:ablation} means we freeze the parameters of the StyleGAN generator and only train the texture encoder $F_{\phi}$ and the subsequent MLPs in our RGI.
Although better SSIM, PSNR, and RMSE can be achieved by (B), 
the FID is poor. 
The last column in \cref{fig:ablation} illustrates an example. 
As shown, fine-tuning can improve hair quality while maintaining the texture of other inner facial components. 

\vspace{-0.3cm}
\section{Face swapping results}

We compare our method with the previous face swapping works: FSGAN~\cite{nirkin2019fsgan}, SimSwap~\cite{chen2020simswap}, FaceShifter~\cite{li2019faceshifter}, and HifiFace~\cite{wang2021hififace}.
We also compare with state-of-the-art StyleGAN-based face swapping methods, including MegaFS~\cite{zhu2021MegaFS}, StyleFusion\cite{kafri2021stylefusion}, and HiRes\cite{xu2022high}.
Specifically, we train our model on the FFHQ dataset. Then, we randomly sample 500 source-target pairs from the CelebAMask-HQ and obtain the swapped results of each method. %

\begin{figure*}[t]
\centering
\setkeys{Gin}{width=\linewidth}
\vspace{-0.6cm}
\begin{subfigure}{0.131\linewidth}
    \caption*{Source}
\includegraphics{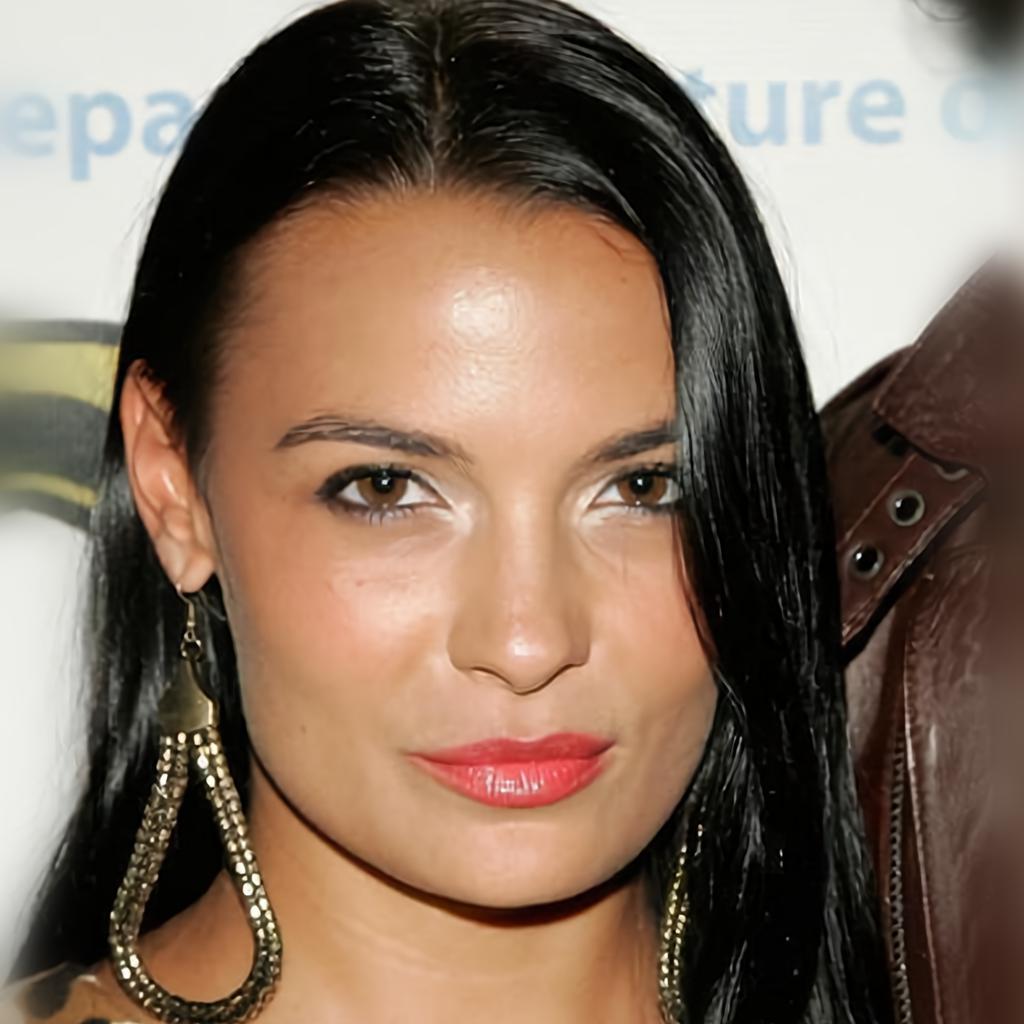}\\
\includegraphics{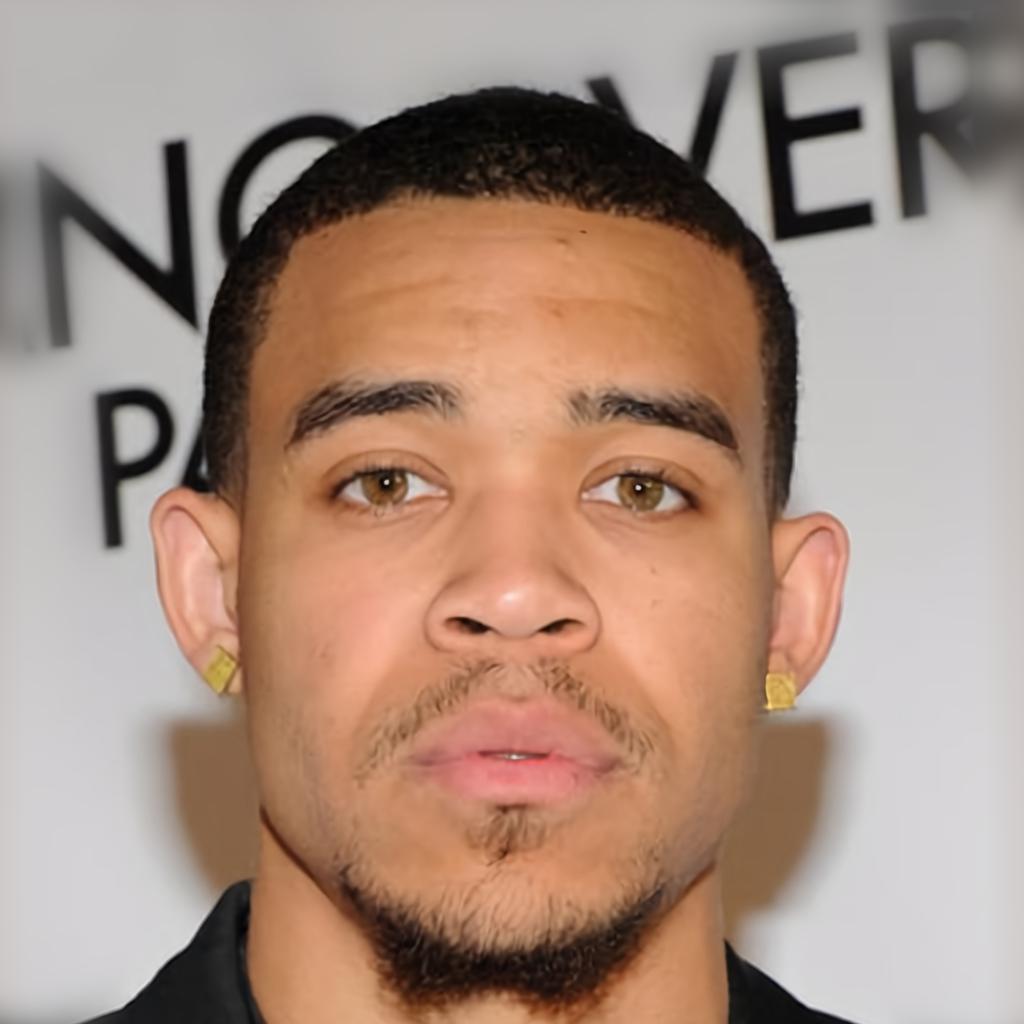}\\
\includegraphics{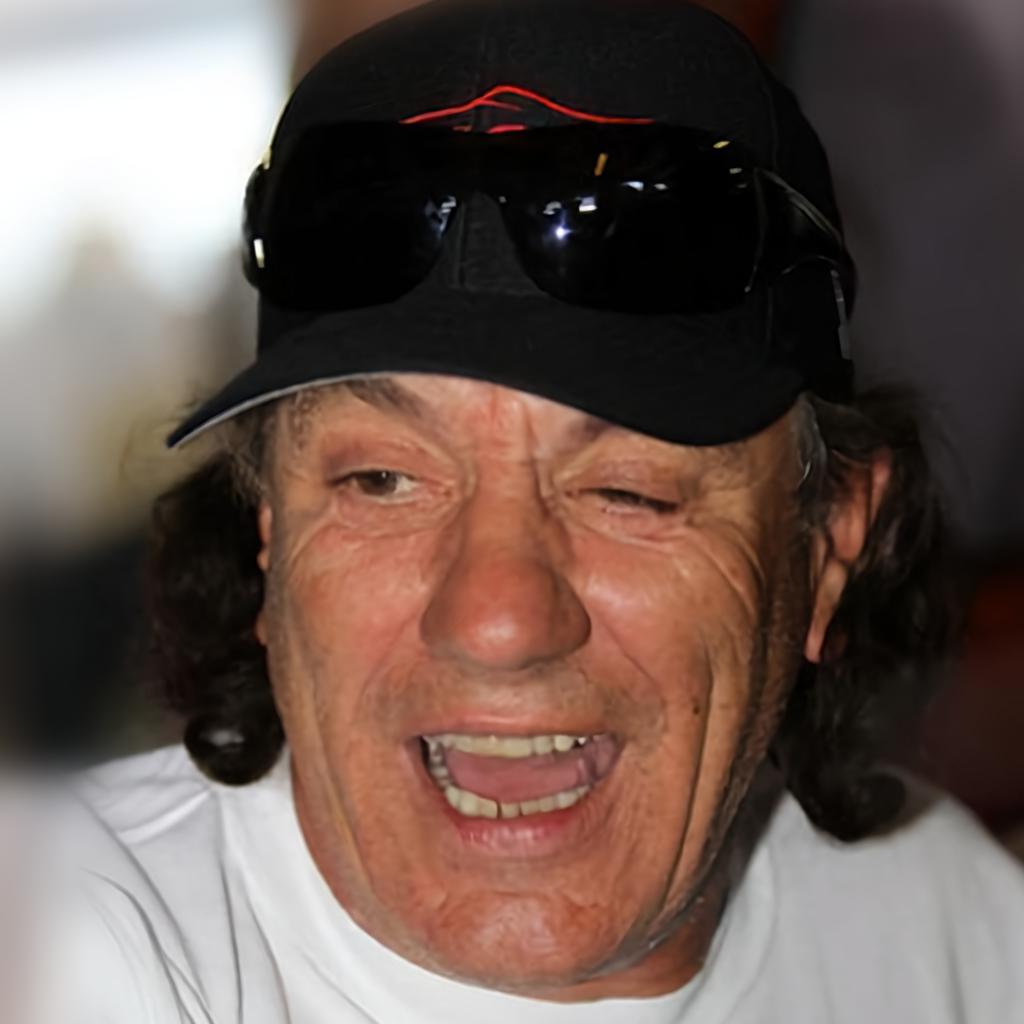}\\
\includegraphics{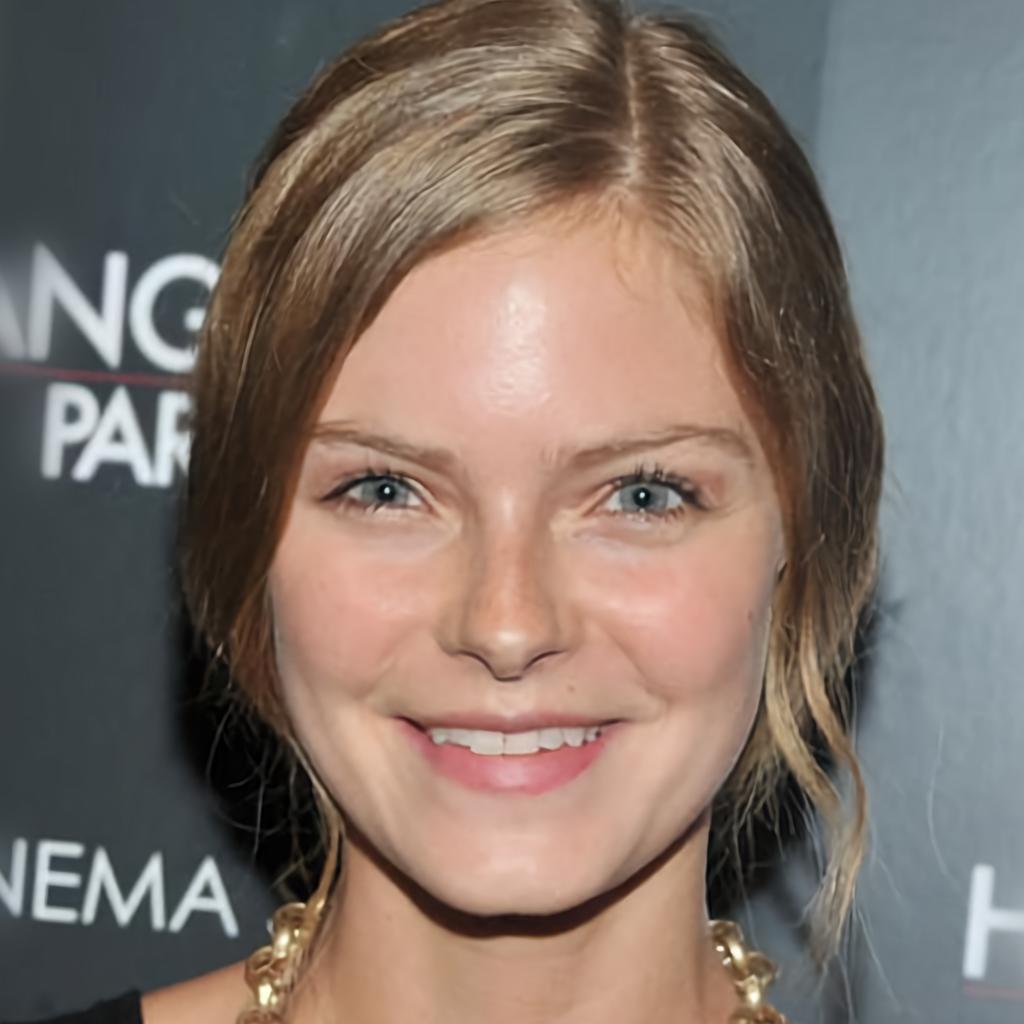}

\end{subfigure}
\begin{subfigure}{0.131\linewidth}
    \caption*{Target}

\includegraphics{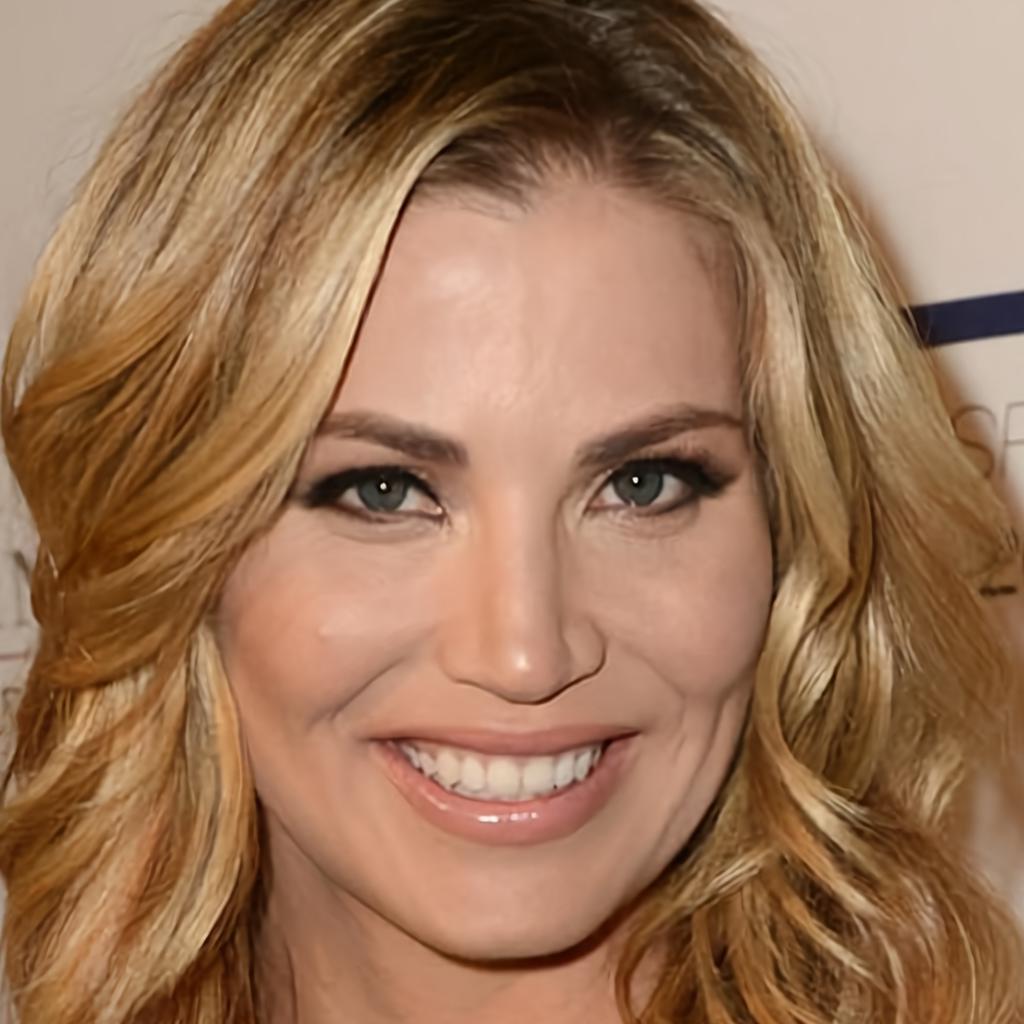}\\
\includegraphics{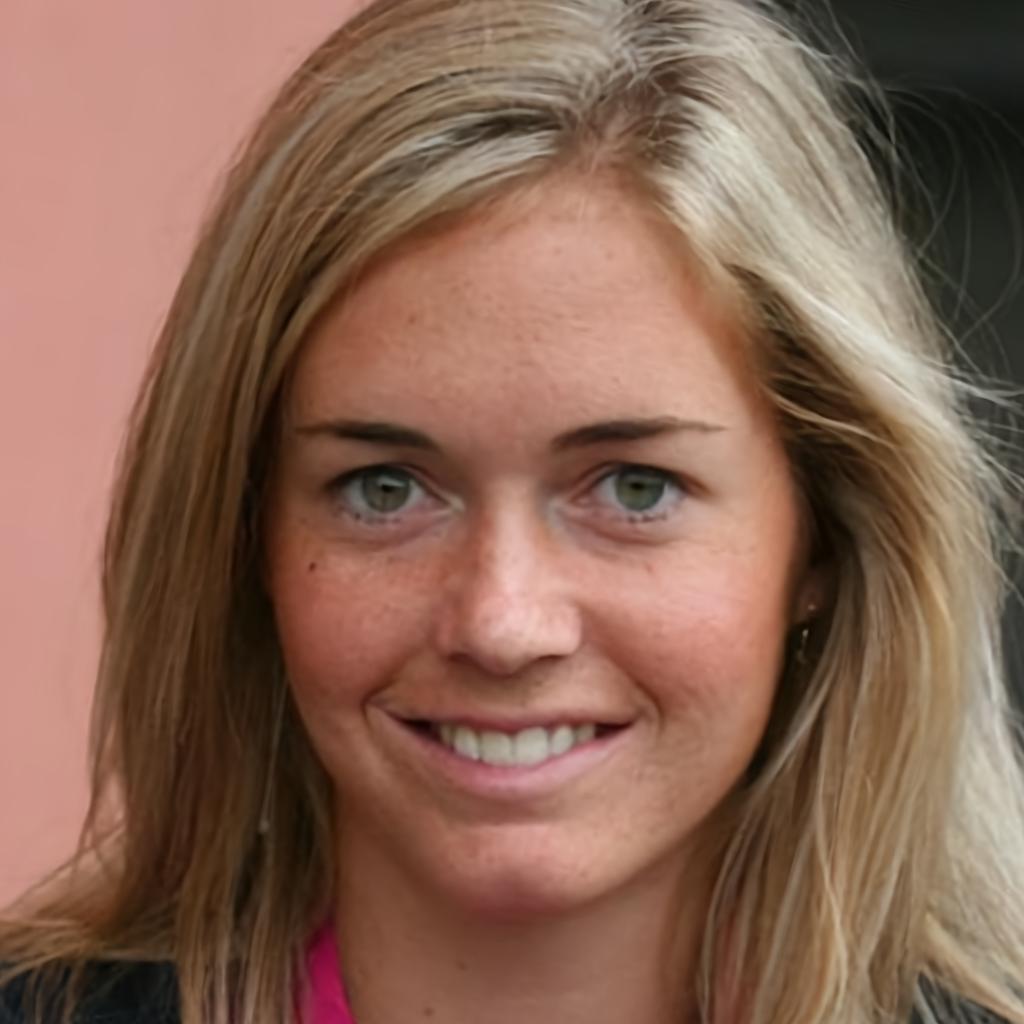}\\
\includegraphics{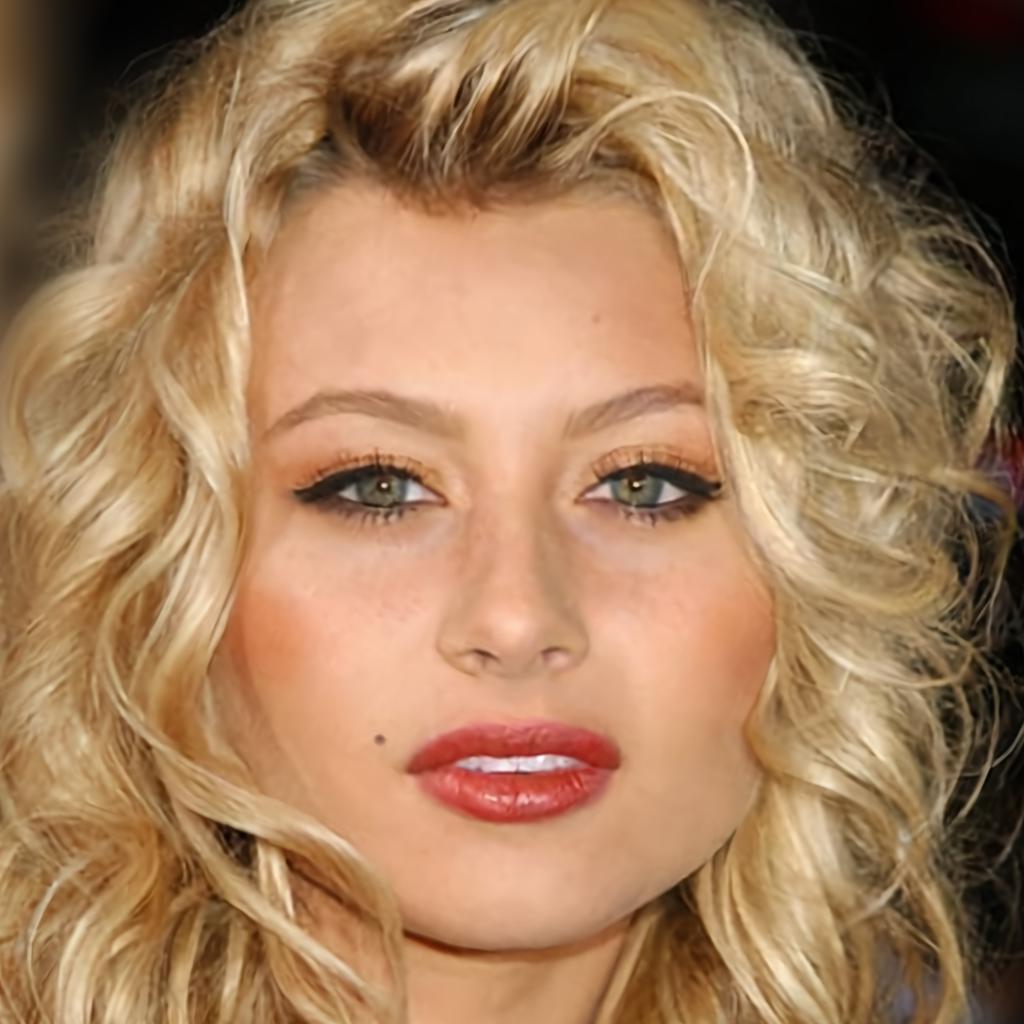}\\ 
\includegraphics{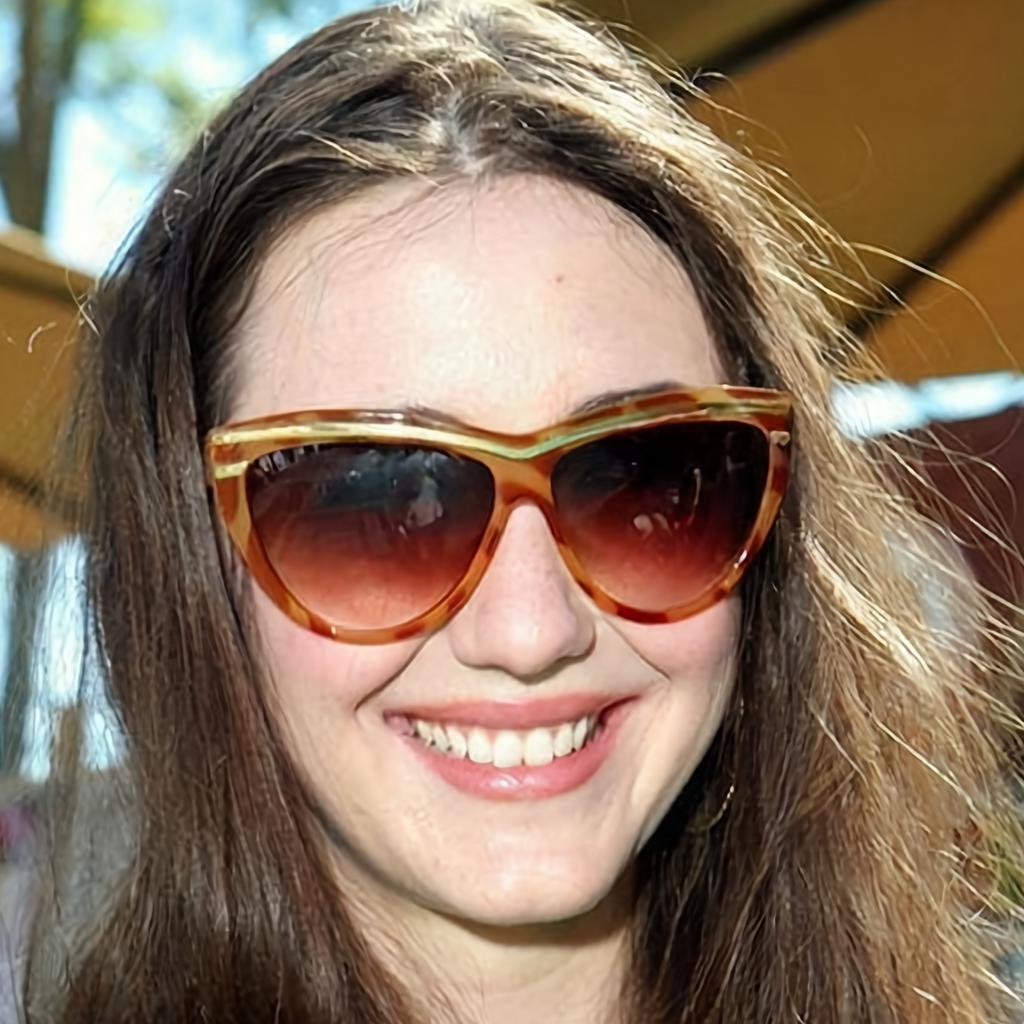}

\end{subfigure}
\begin{subfigure}{0.131\linewidth}
    \caption*{FSGAN\cite{nirkin2019fsgan}}

\includegraphics{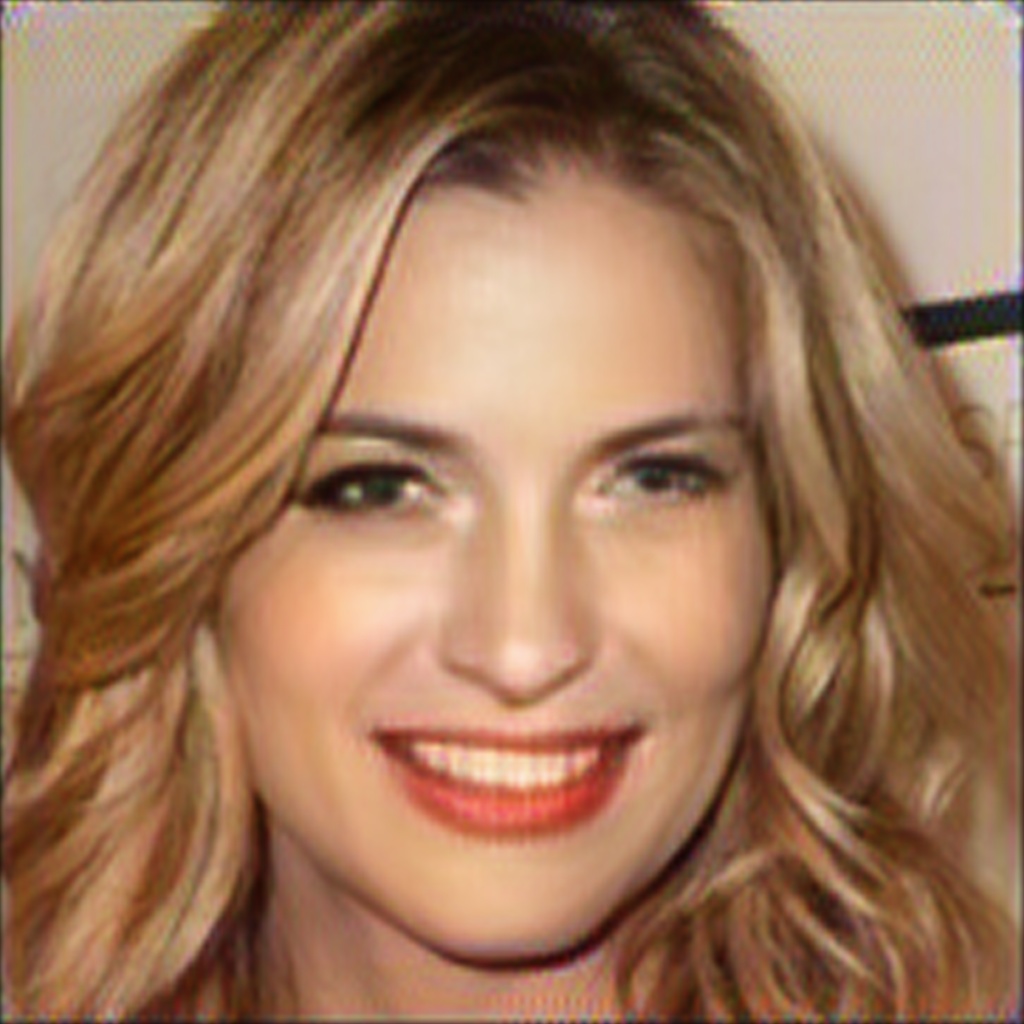}\\
\includegraphics{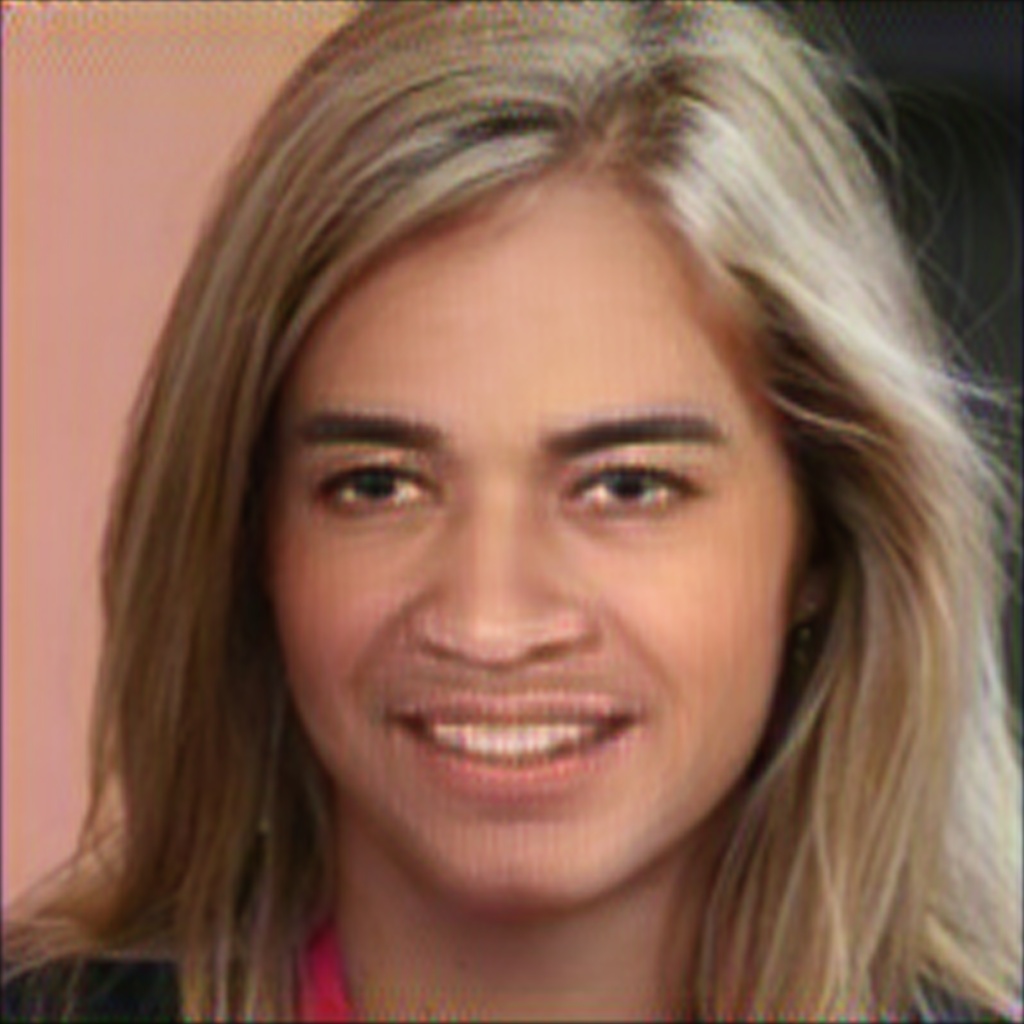}\\
\includegraphics{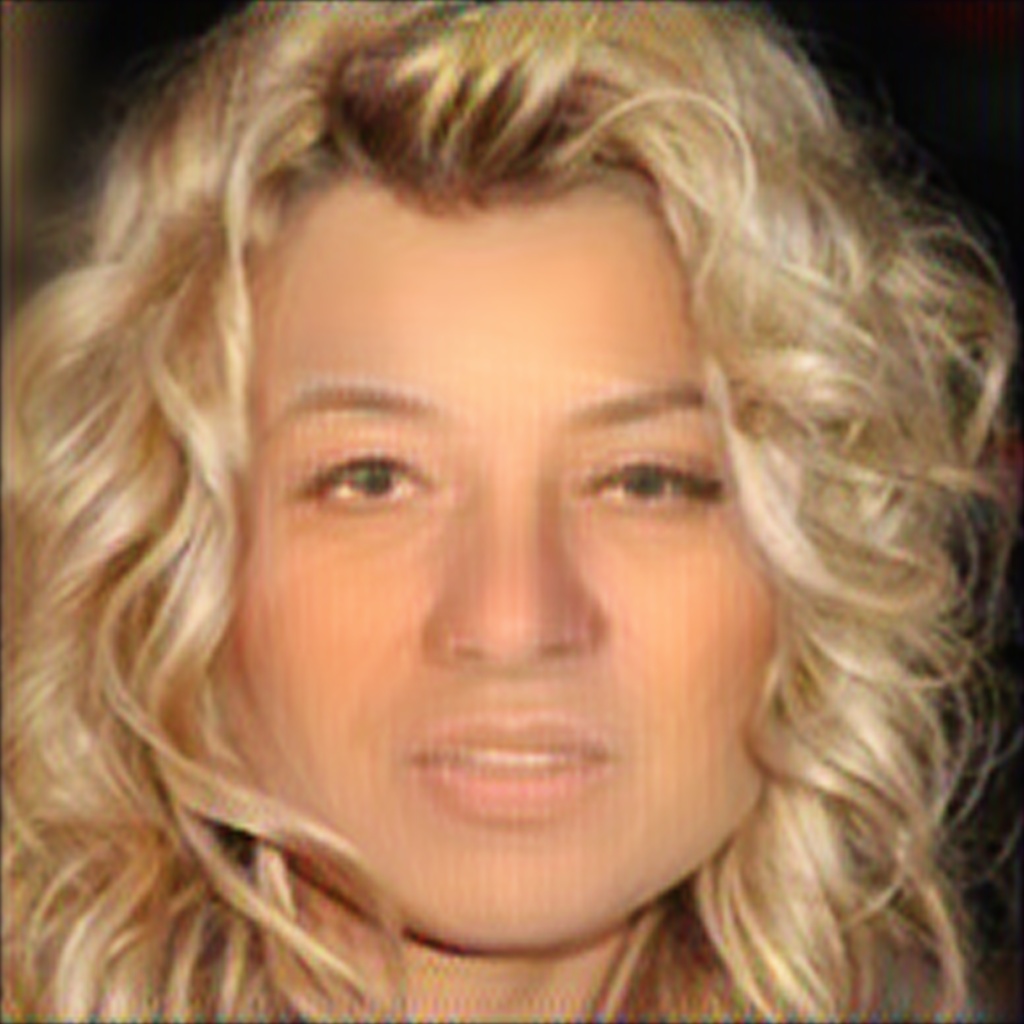}\\
\includegraphics{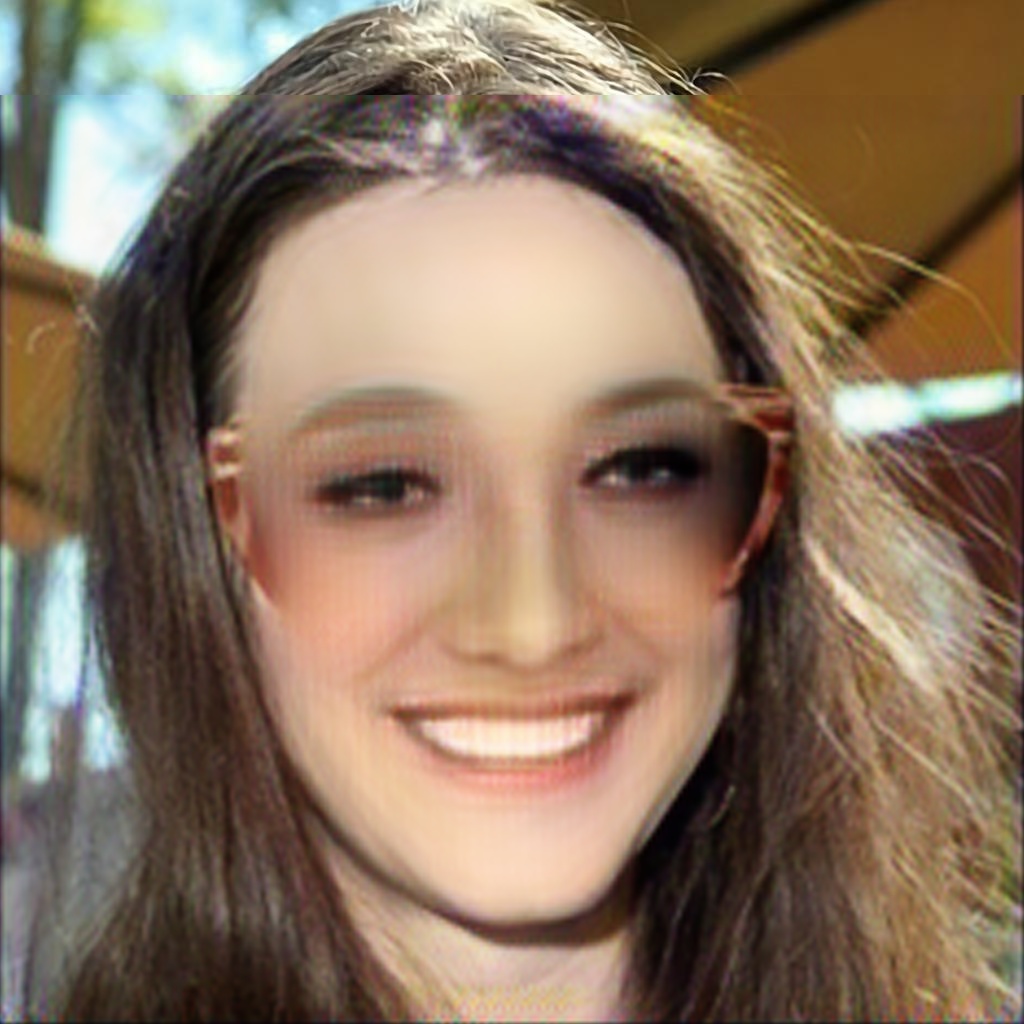}

\end{subfigure}
\begin{subfigure}{0.131\linewidth}
    \caption*{SimSwap\cite{chen2020simswap}}

\includegraphics{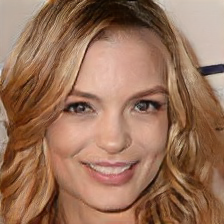}\\
\includegraphics{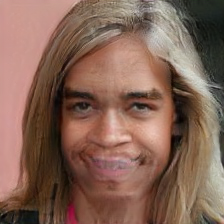}\\
\includegraphics{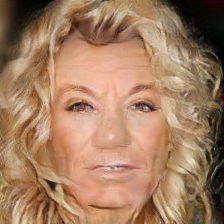}\\
\includegraphics{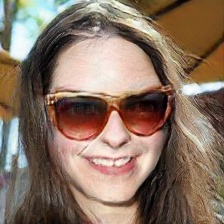}
\end{subfigure}
\begin{subfigure}{0.131\linewidth}
    \caption*{FaceShifter\cite{li2019faceshifter}}

\includegraphics{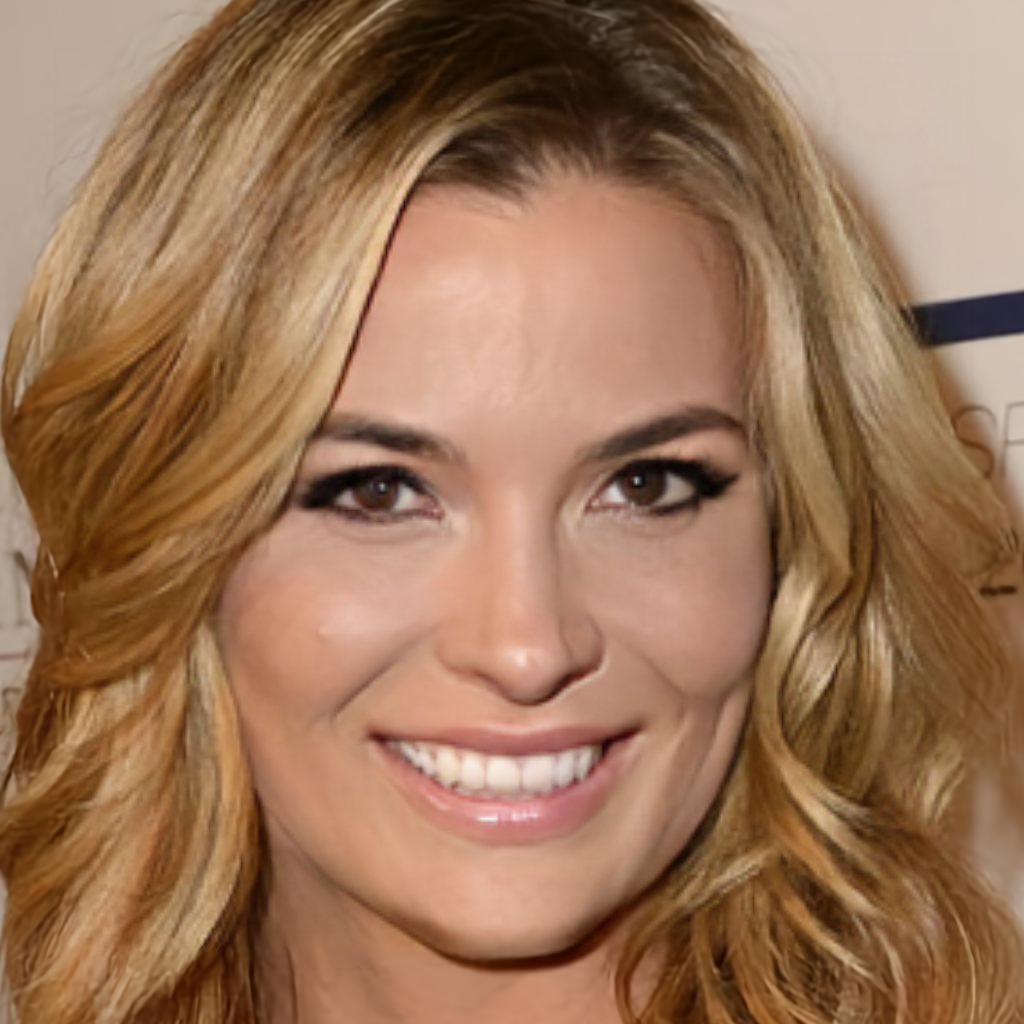}\\
\includegraphics{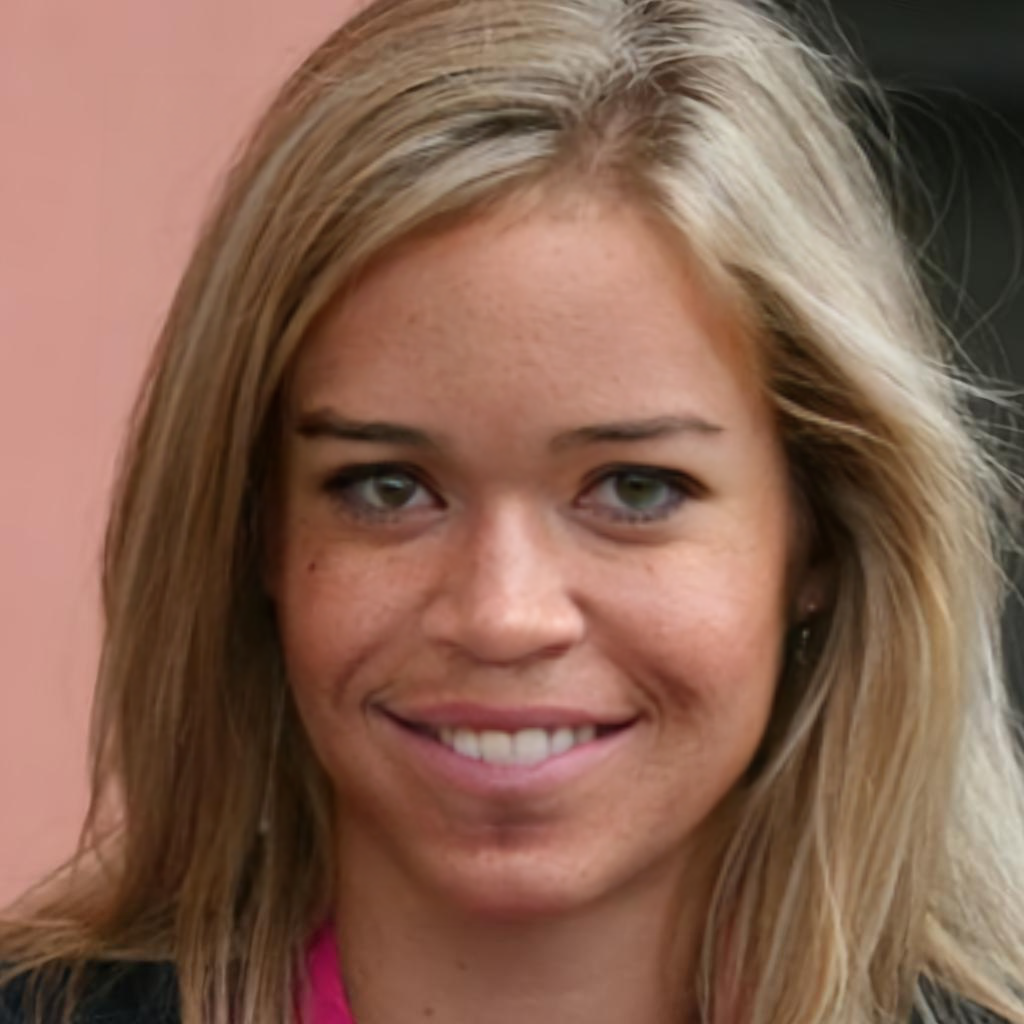}\\
\includegraphics{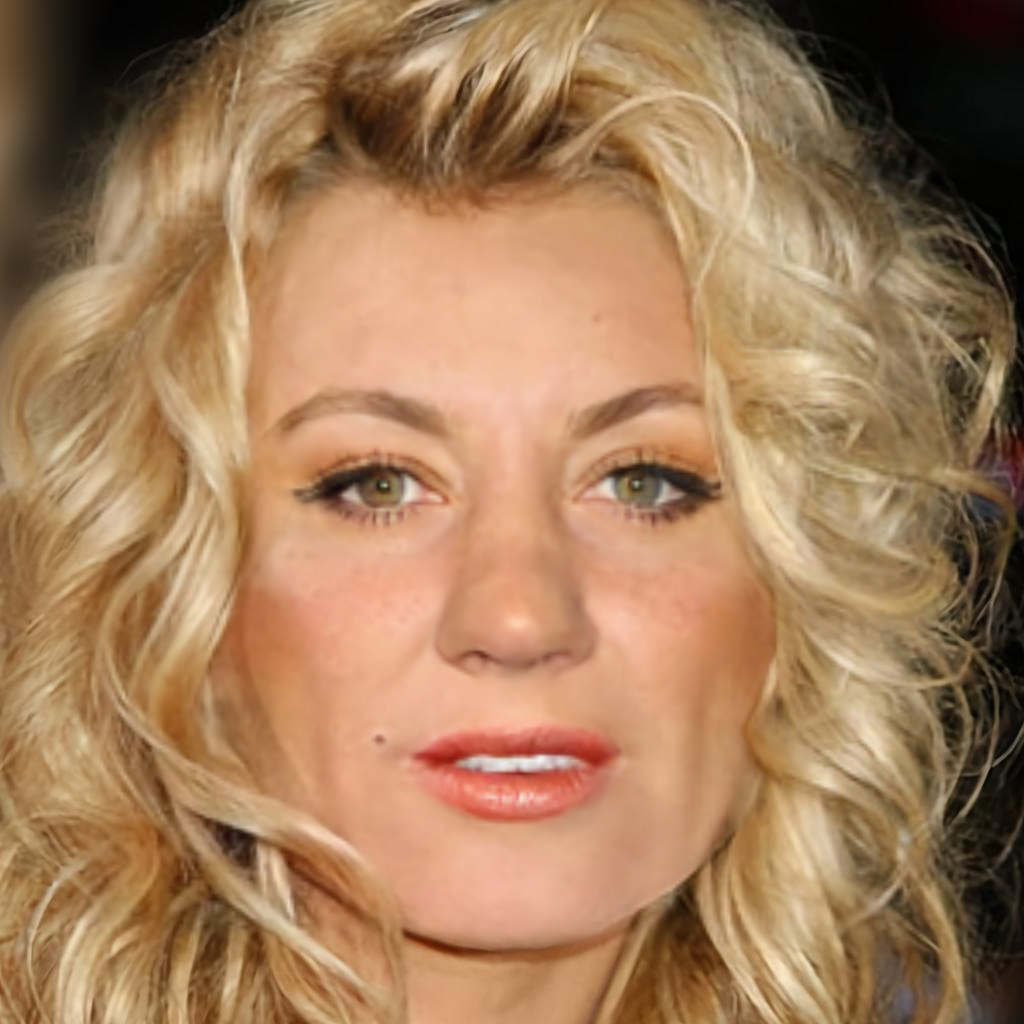}\\
\includegraphics{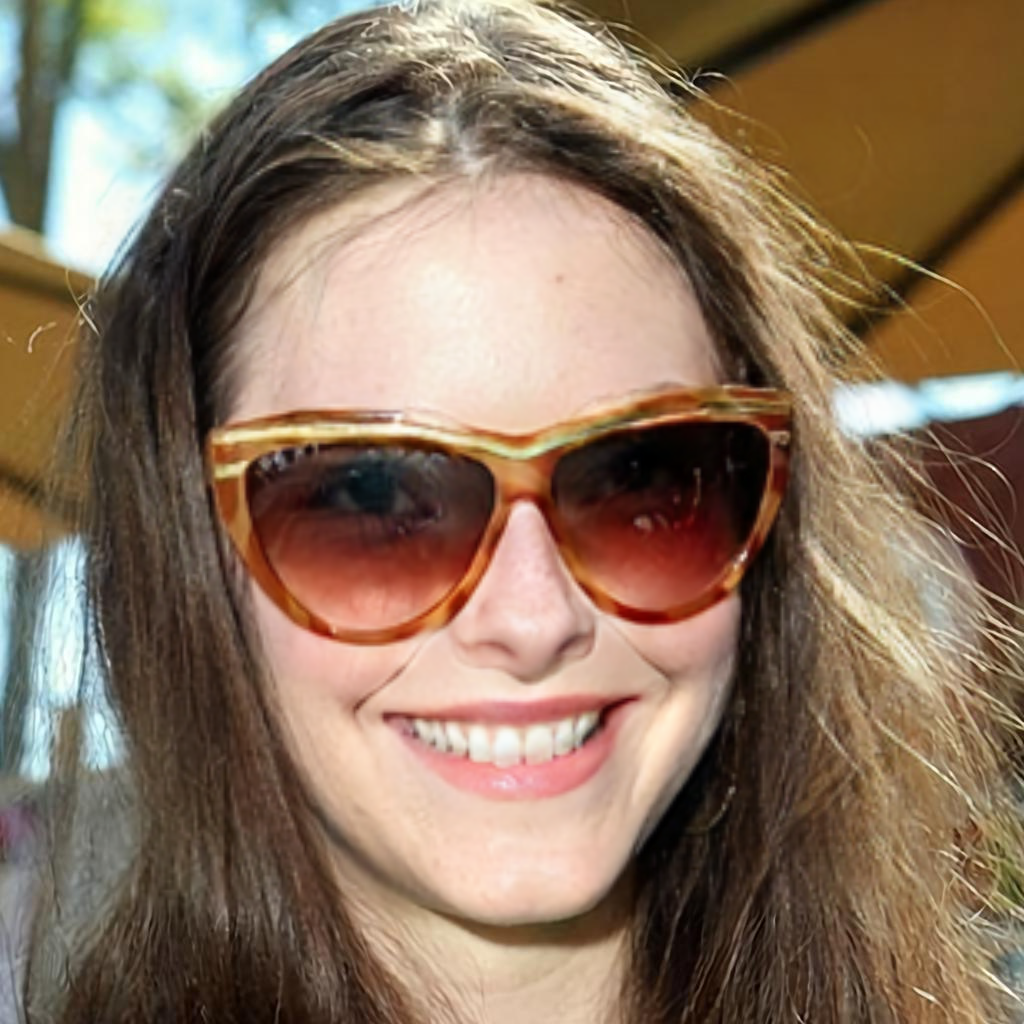}

\end{subfigure}
\begin{subfigure}{0.131\linewidth}
    \caption*{HifiFace\cite{wang2021hififace}}

\includegraphics{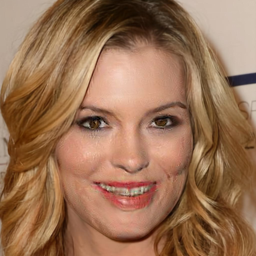}\\
\includegraphics{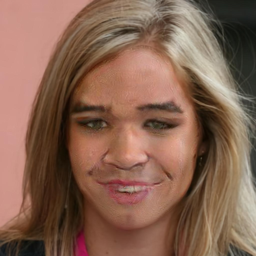}\\
\includegraphics{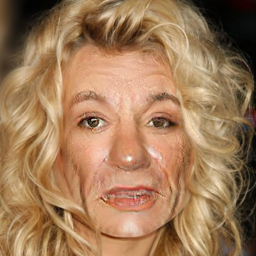}\\ 
\includegraphics{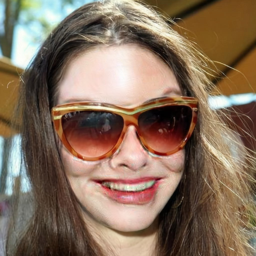}
\end{subfigure}
\begin{subfigure}{0.131\linewidth}
    \caption*{Ours}
    
\includegraphics{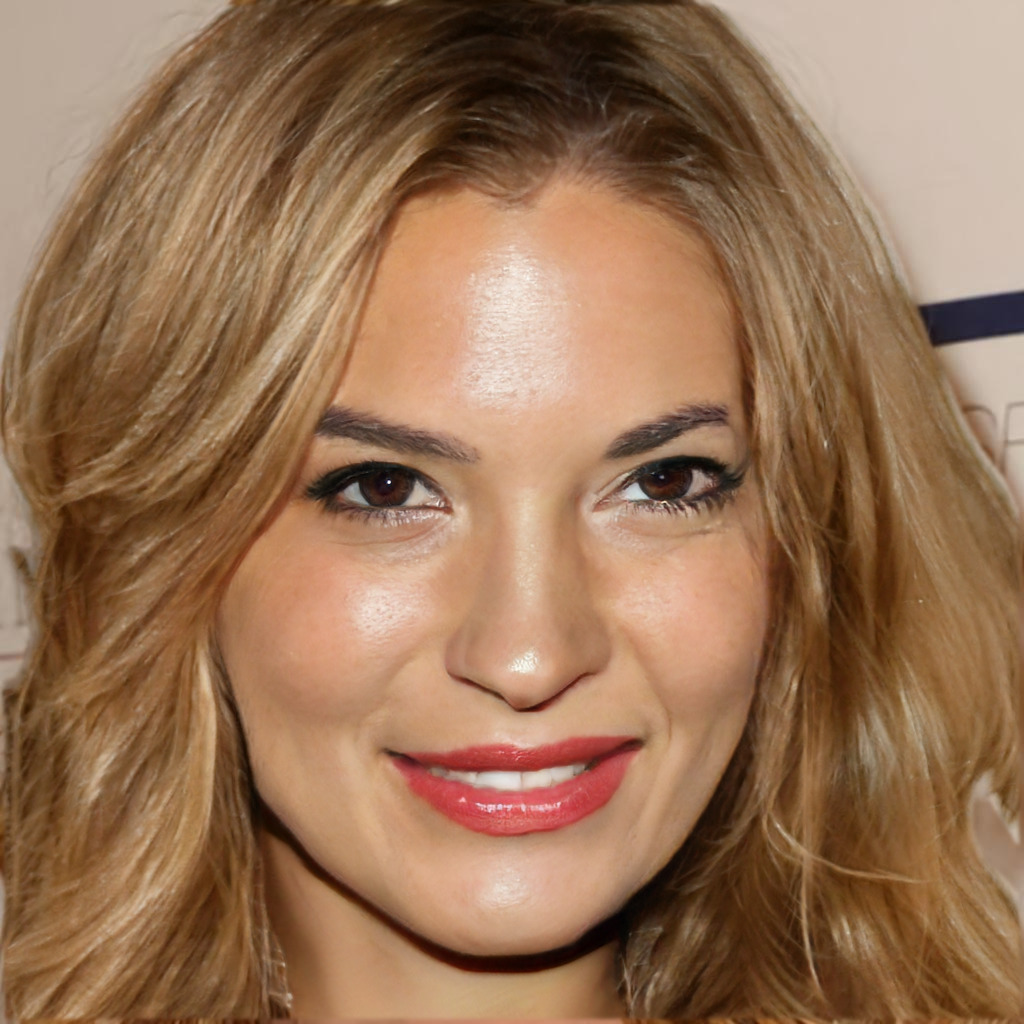}\\
\includegraphics{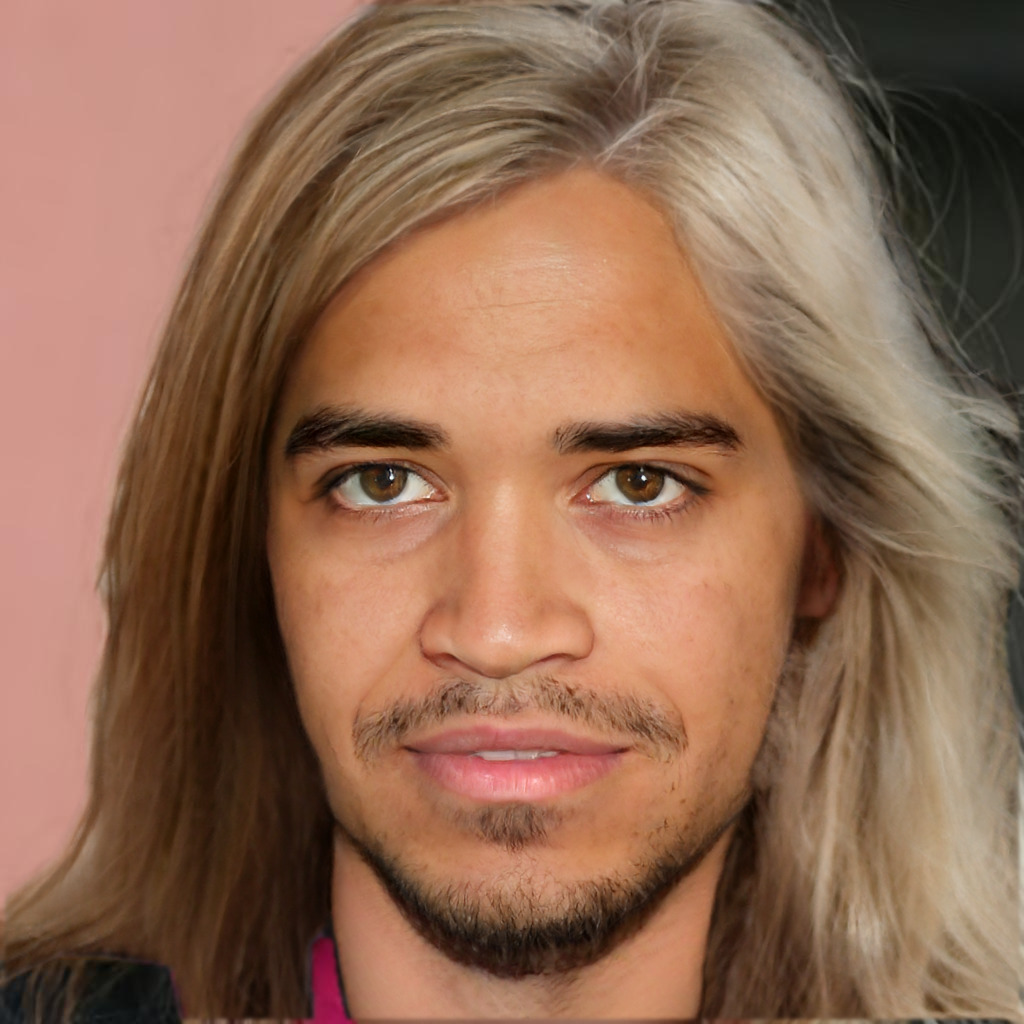}\\
\includegraphics{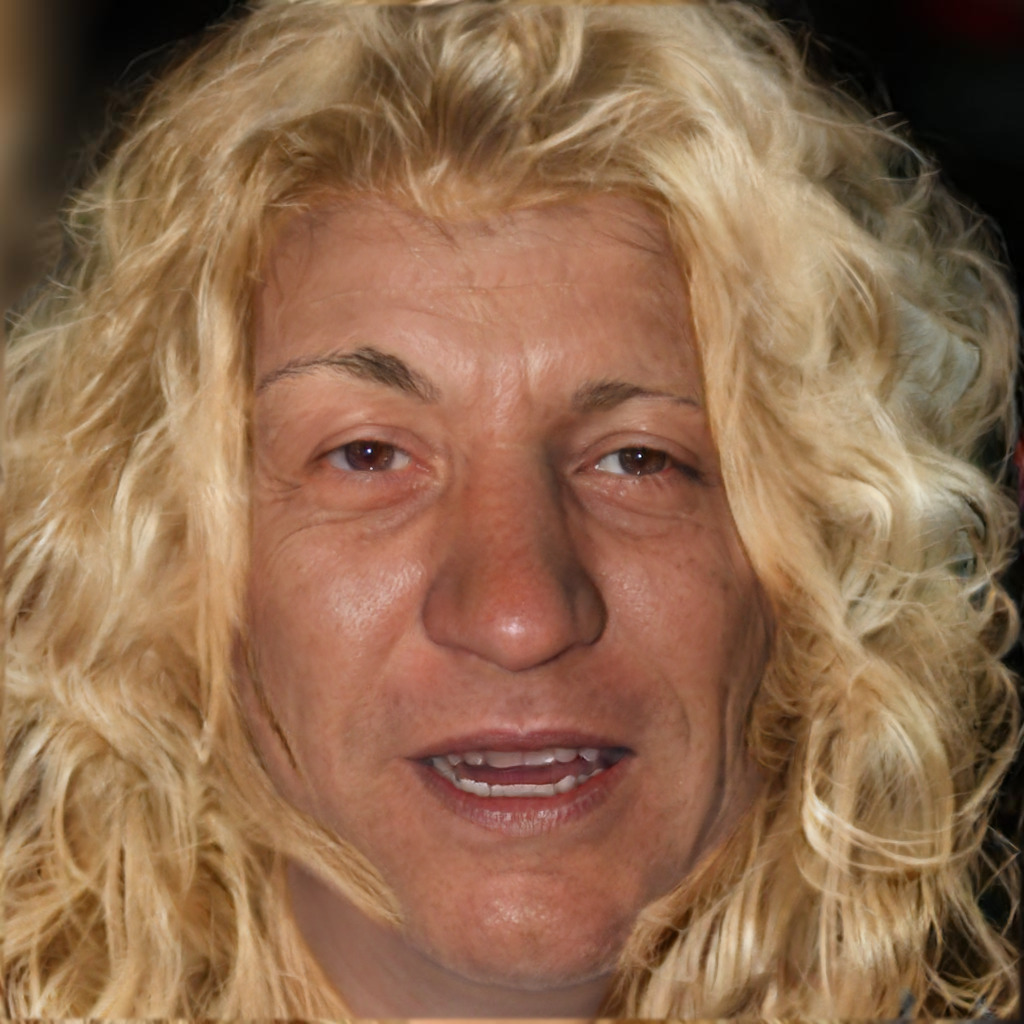}\\
\includegraphics{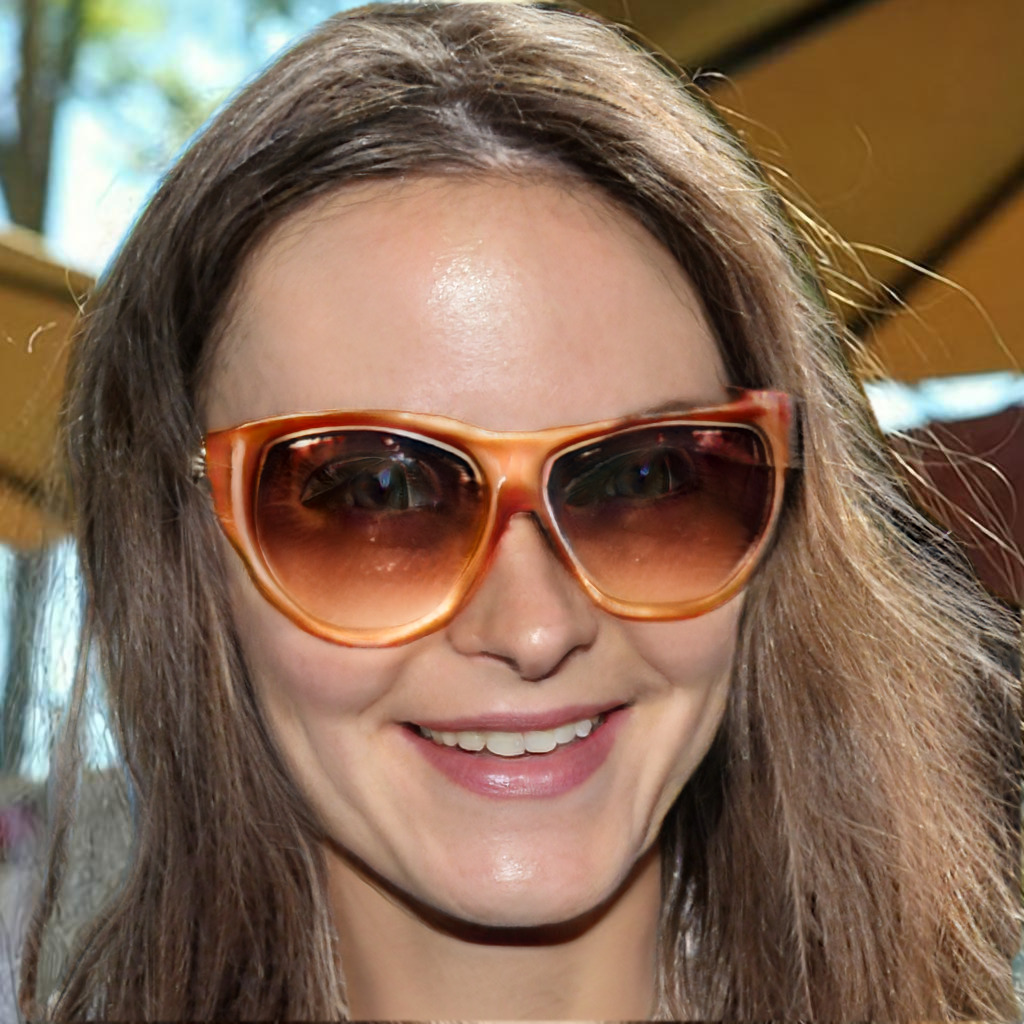}
\end{subfigure}
    \captionsetup{belowskip=-12pt}
    \caption{Qualitative comparisons of our results with state-of-the-art face swapping  methods. Best viewed in color and zoom-in.}
    \label{fig:swapComp}
    \end{figure*}

\begin{figure}[t]
\centering
\setkeys{Gin}{width=\linewidth}
\begin{subfigure}{0.15\linewidth}
    \caption*{Source}
\includegraphics{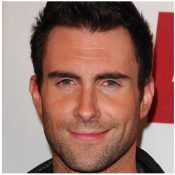}\\
\includegraphics{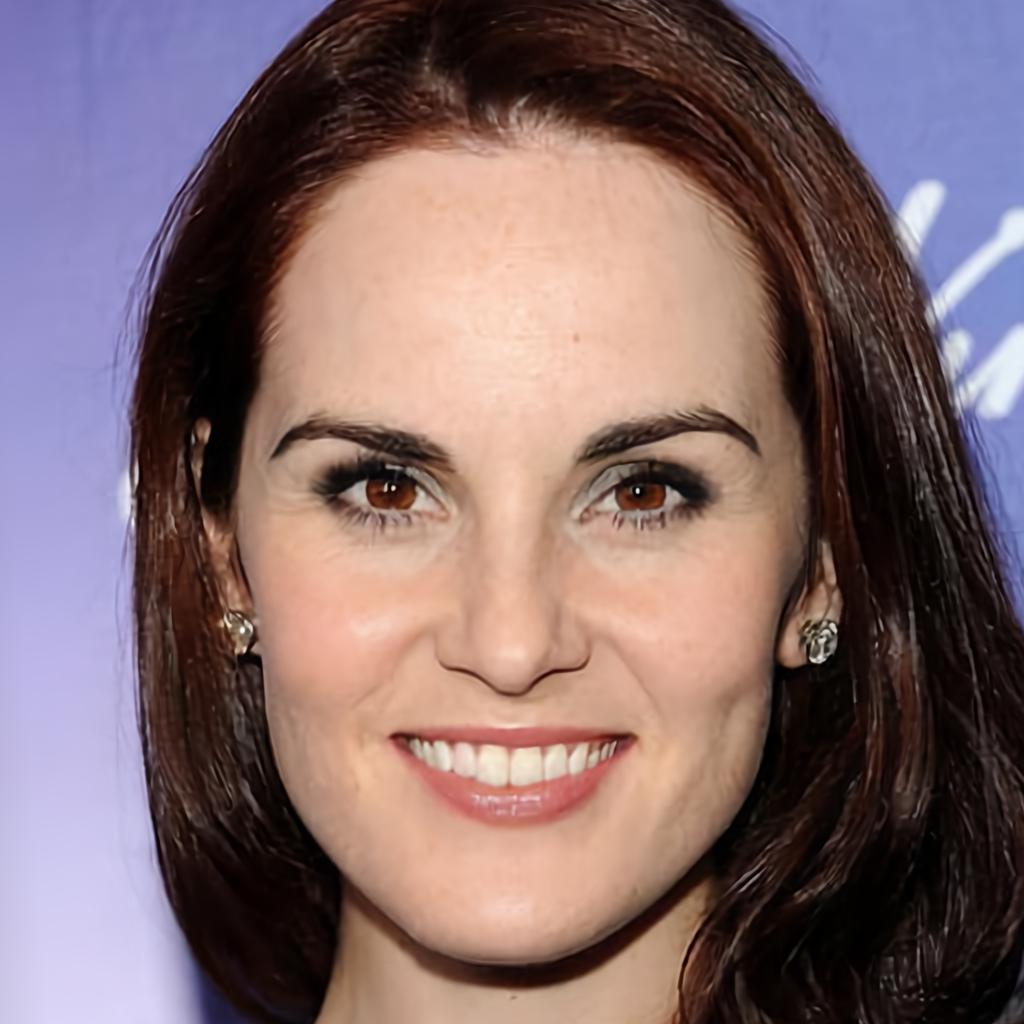}\\
\includegraphics{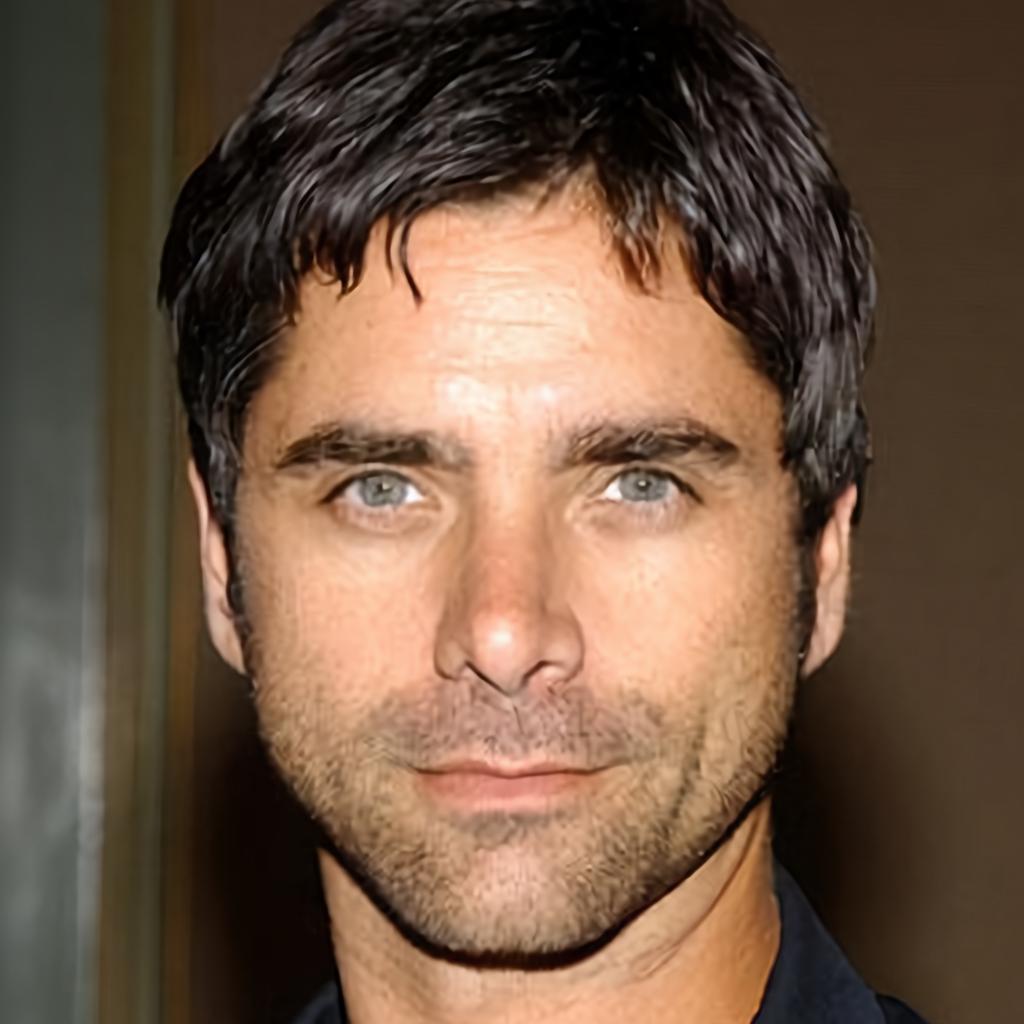}
\end{subfigure}
\begin{subfigure}{0.15\linewidth}
    \caption*{Target}
\includegraphics{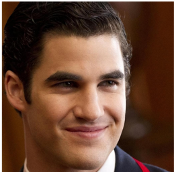}\\
\includegraphics{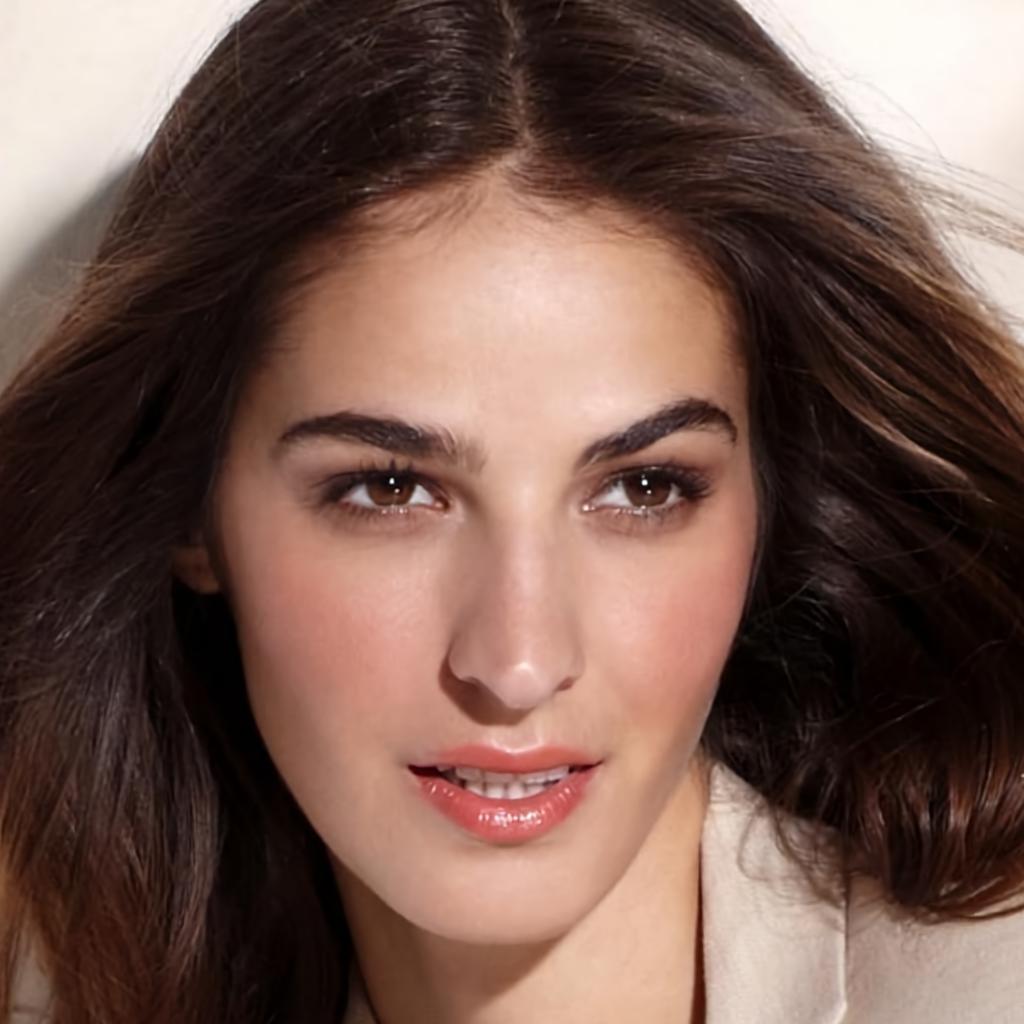}\\
\includegraphics{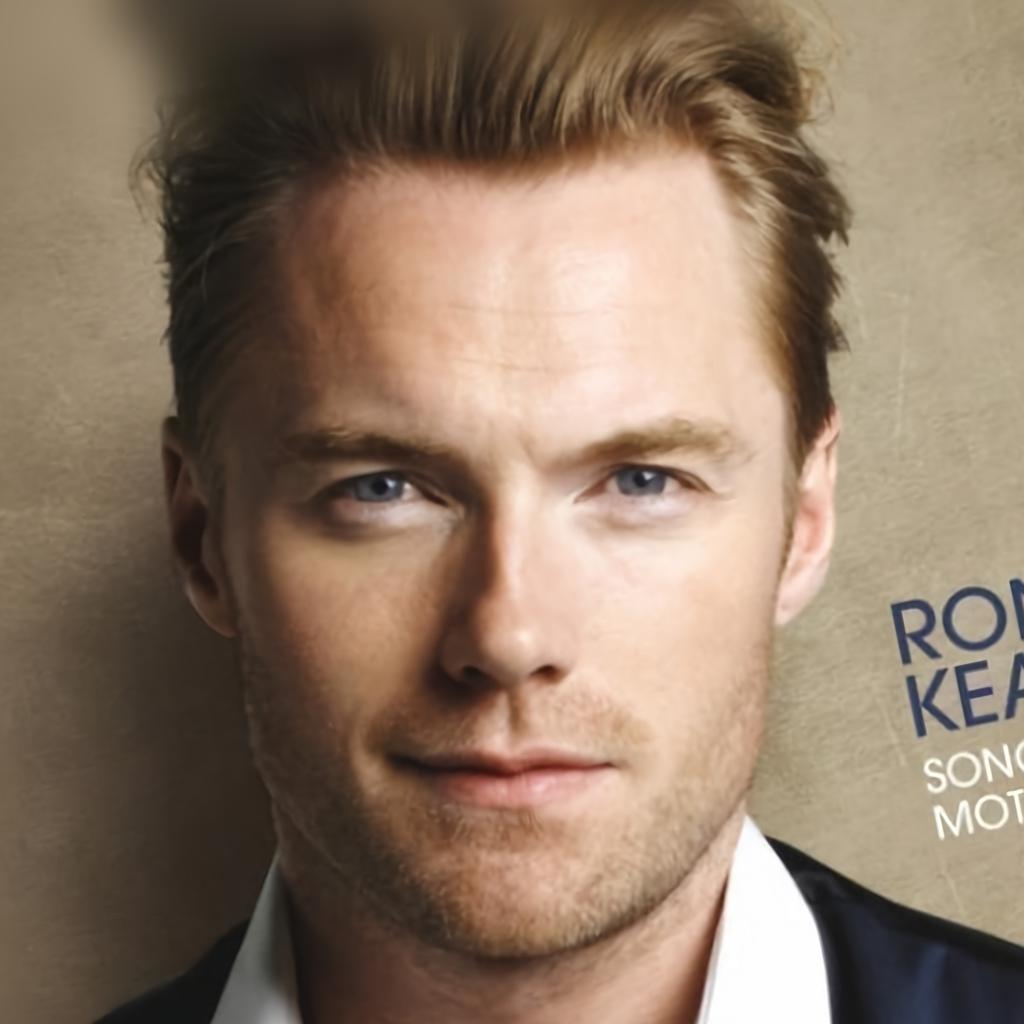}
\end{subfigure}
\begin{subfigure}{0.15\linewidth}
    \caption*{ MegaFS}
\includegraphics{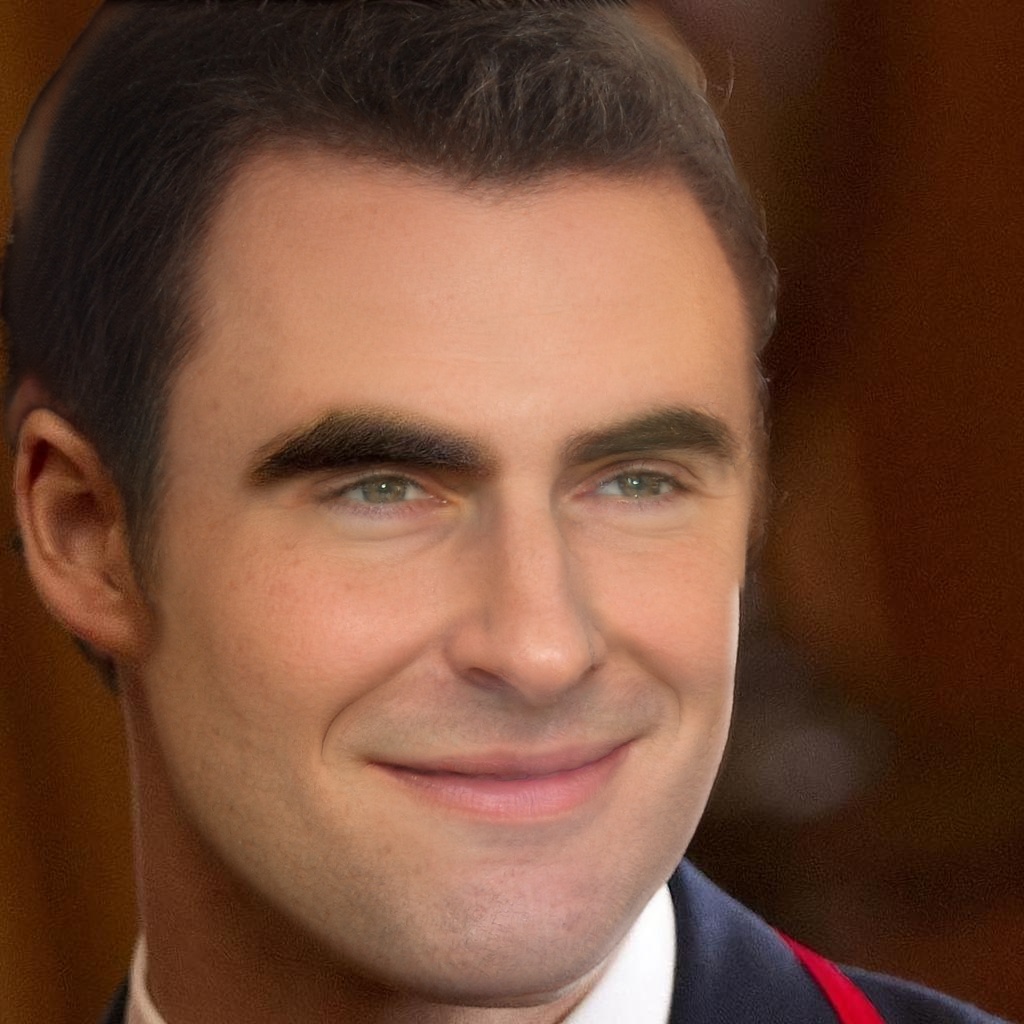}\\
\includegraphics{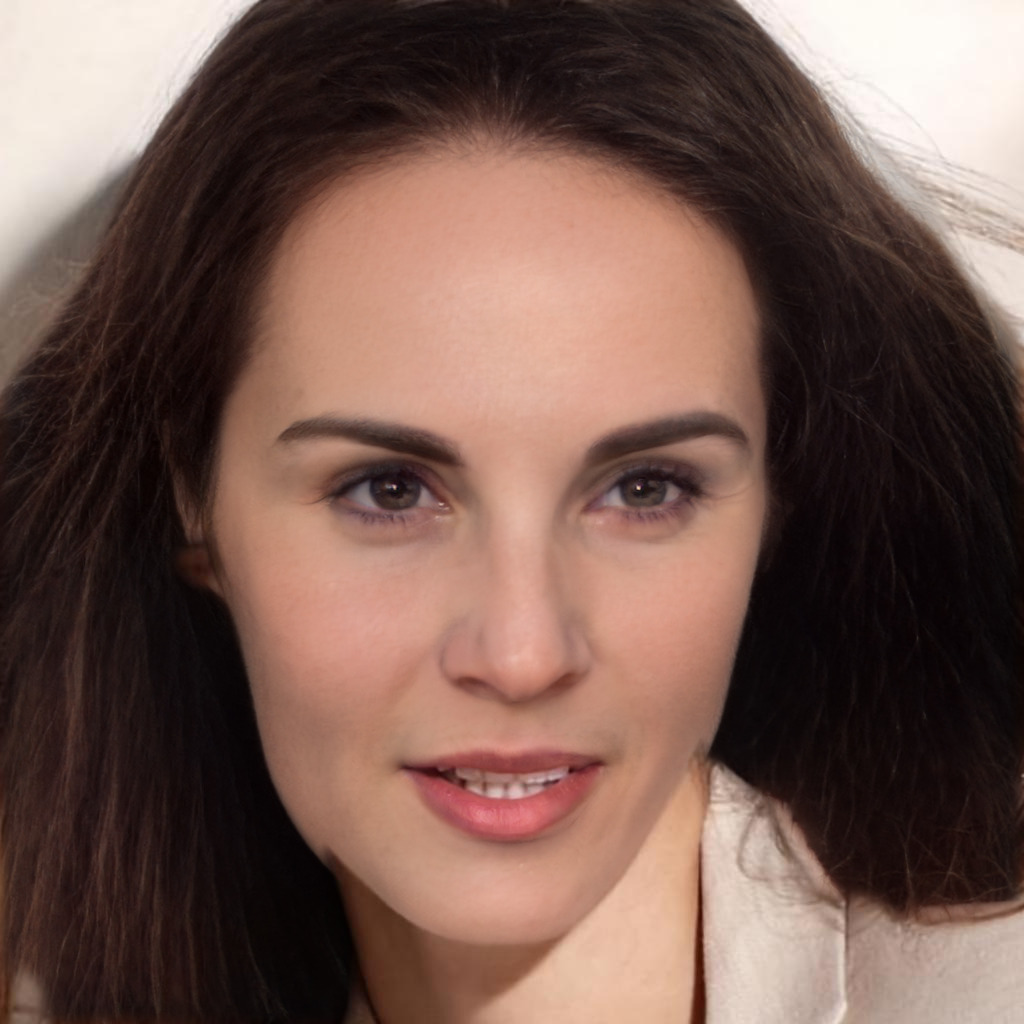}\\
\includegraphics{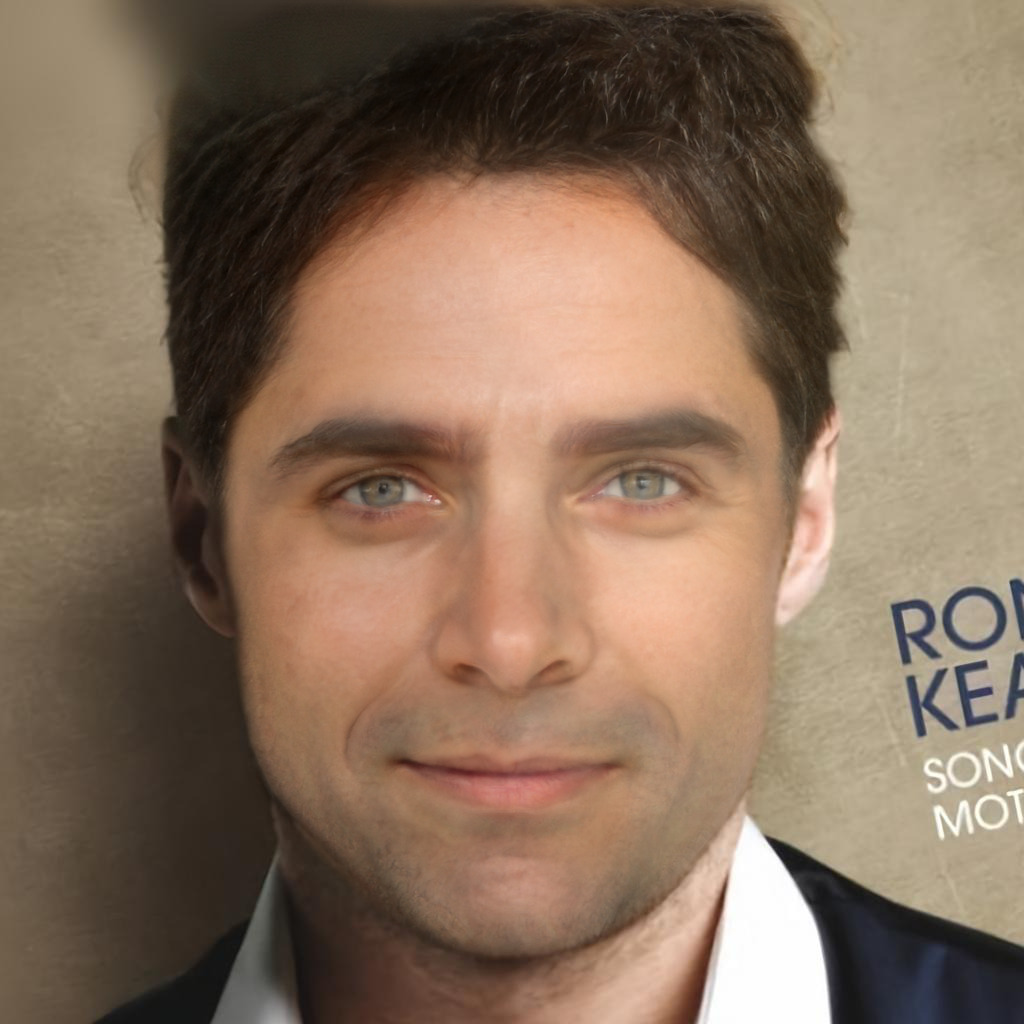}
\end{subfigure}
\captionsetup[sub]{font=scriptsize,labelfont={bf,sf}}
\begin{subfigure}{0.15\linewidth}
    \caption*{StyleFusion}
\includegraphics{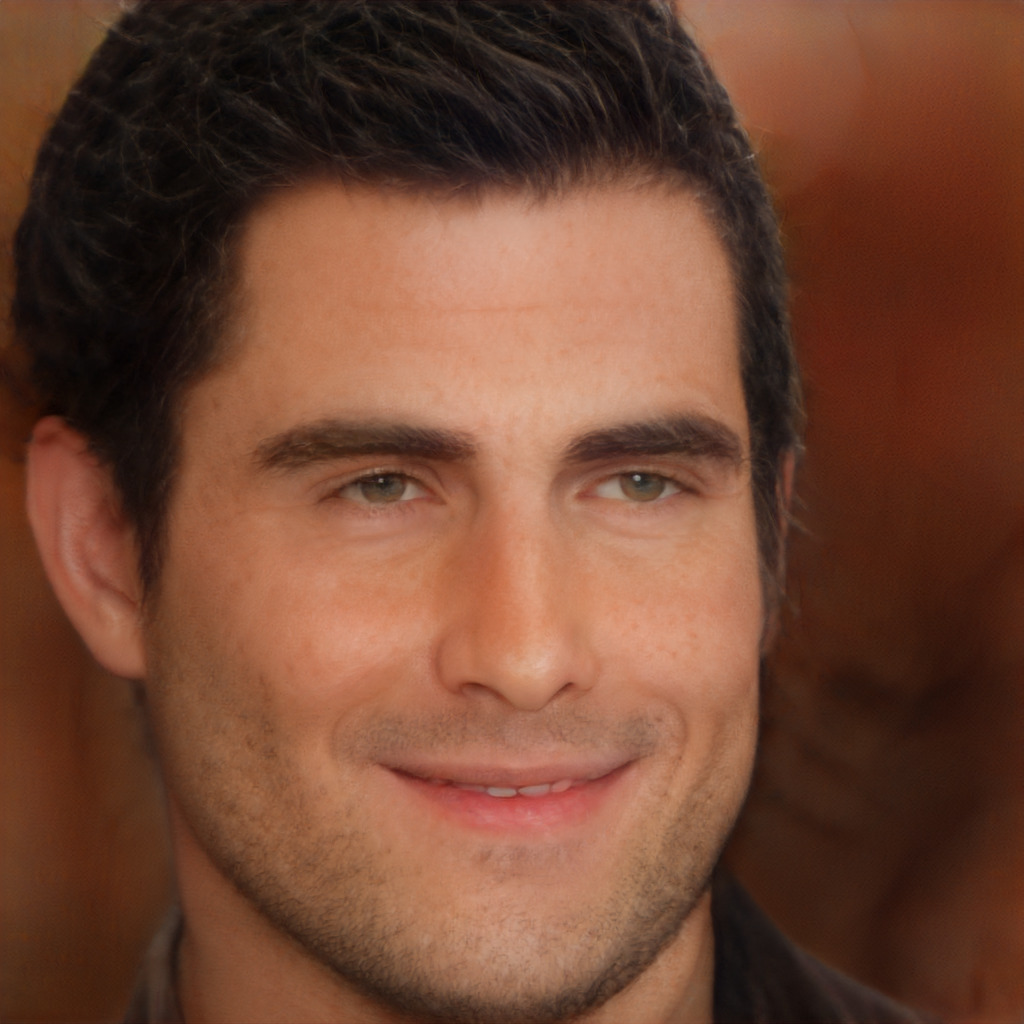}\\
\includegraphics{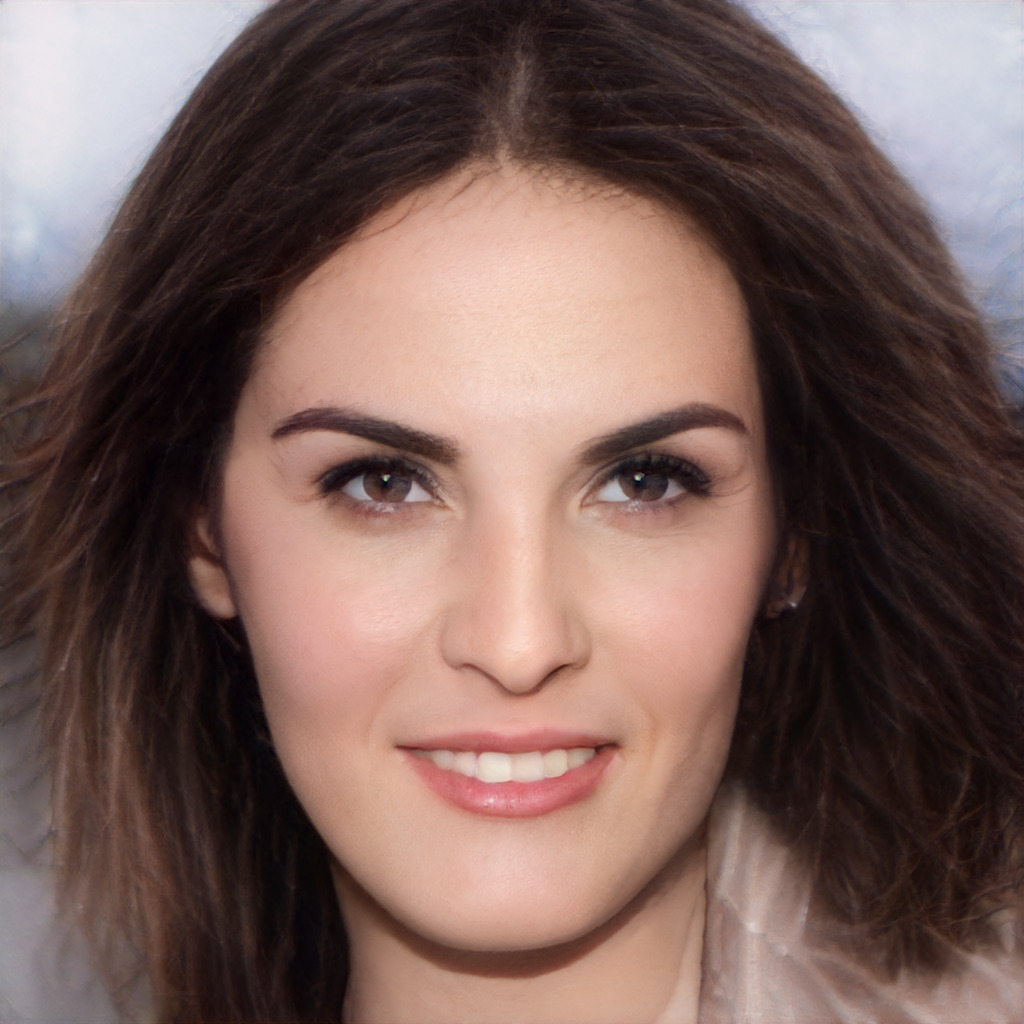}\\
\includegraphics{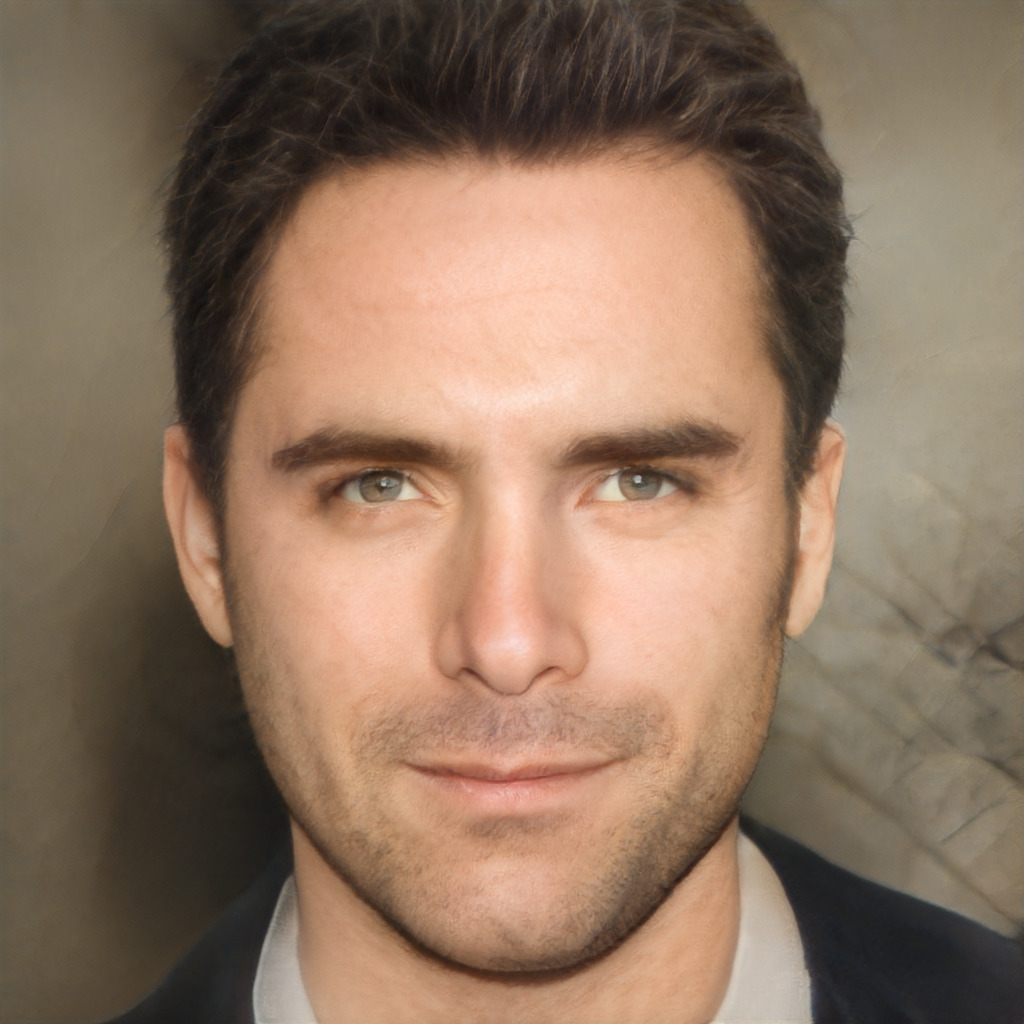}
\end{subfigure}
\captionsetup[sub]{font=footnotesize}
\begin{subfigure}{0.15\linewidth}
    \caption*{HiRes}
\includegraphics{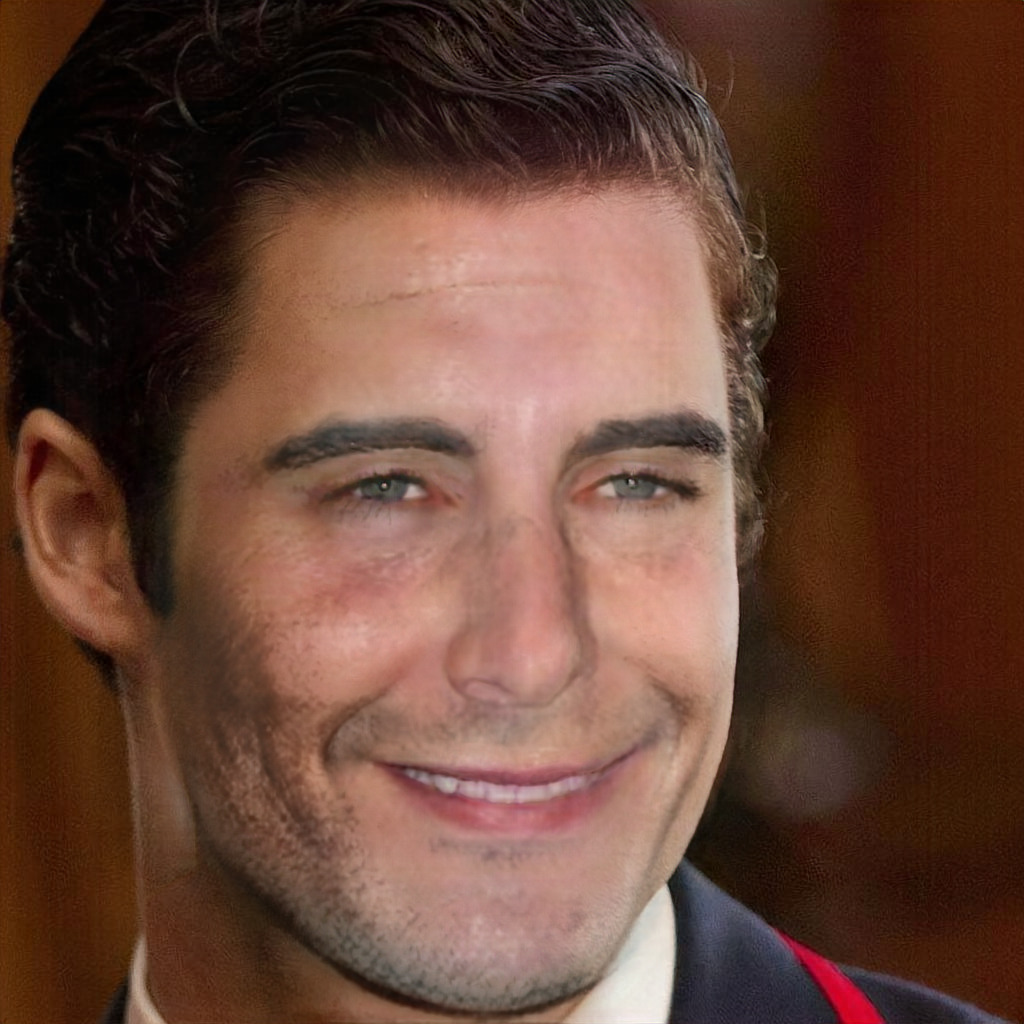}\\
\includegraphics{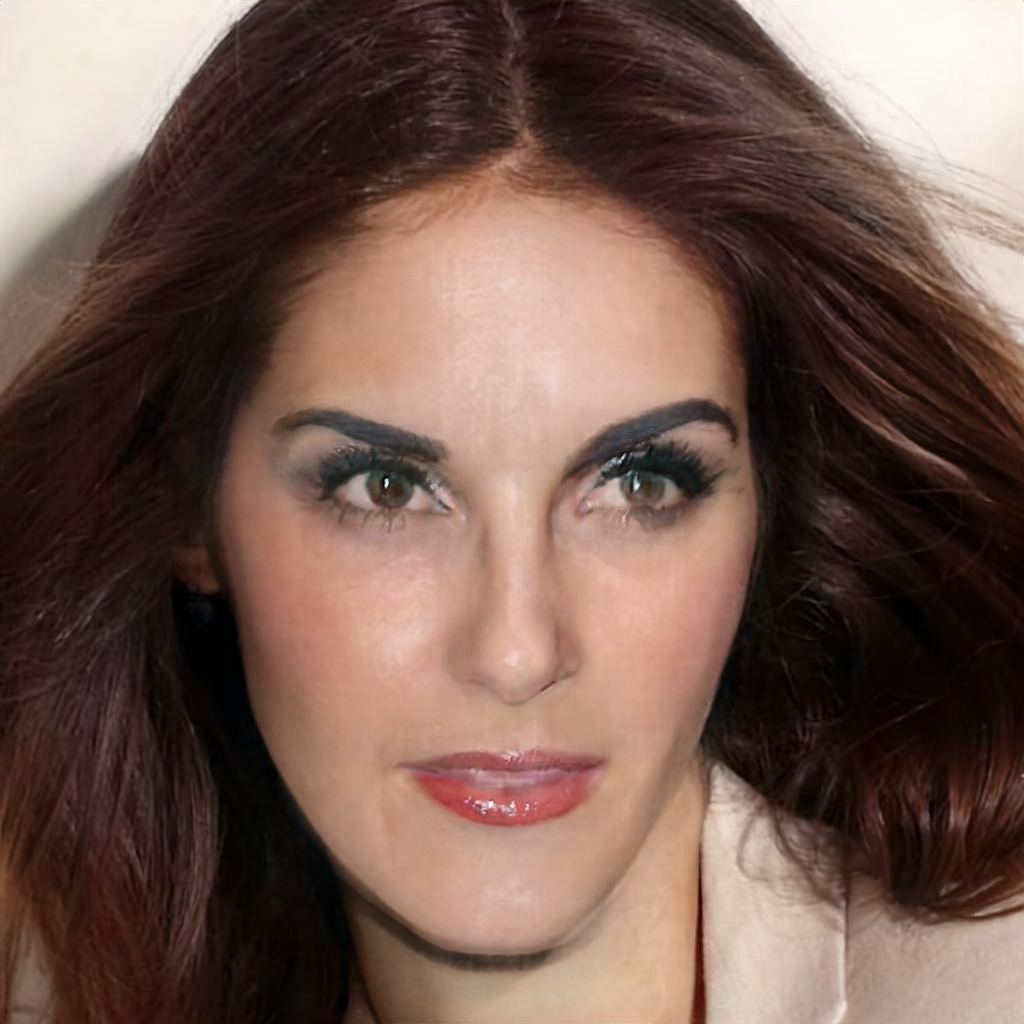}\\
\includegraphics{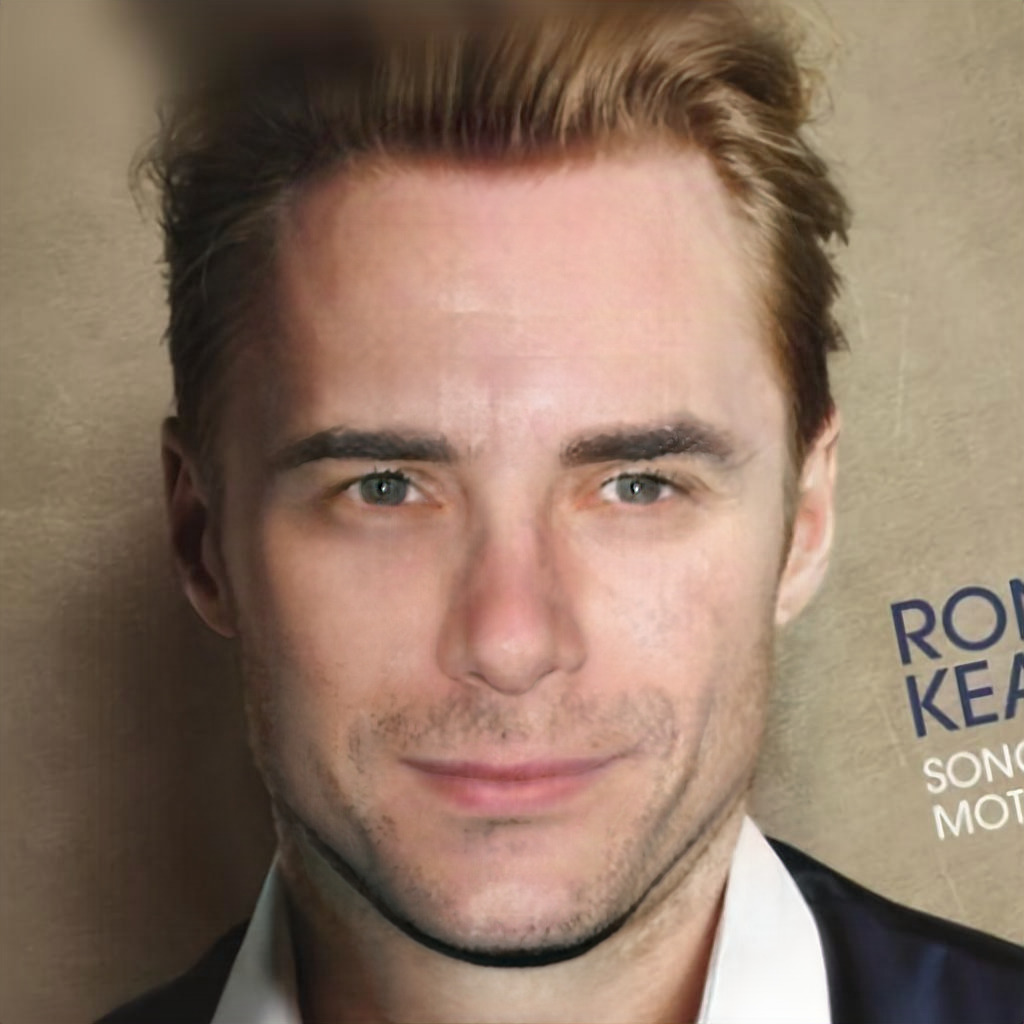}
\end{subfigure}
\begin{subfigure}{0.15\linewidth}
    \caption*{Ours}
\includegraphics{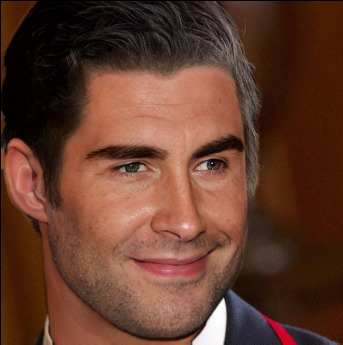}\\
\includegraphics{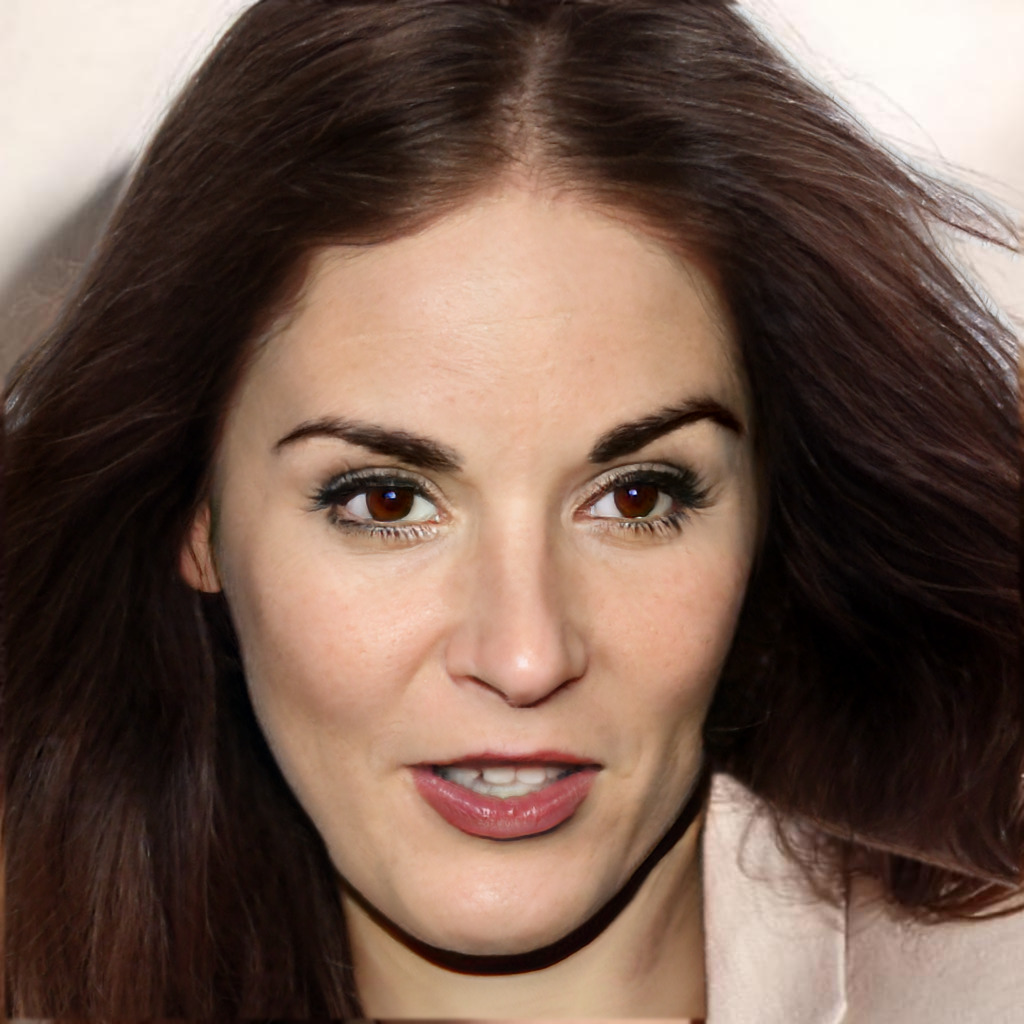}\\
\includegraphics{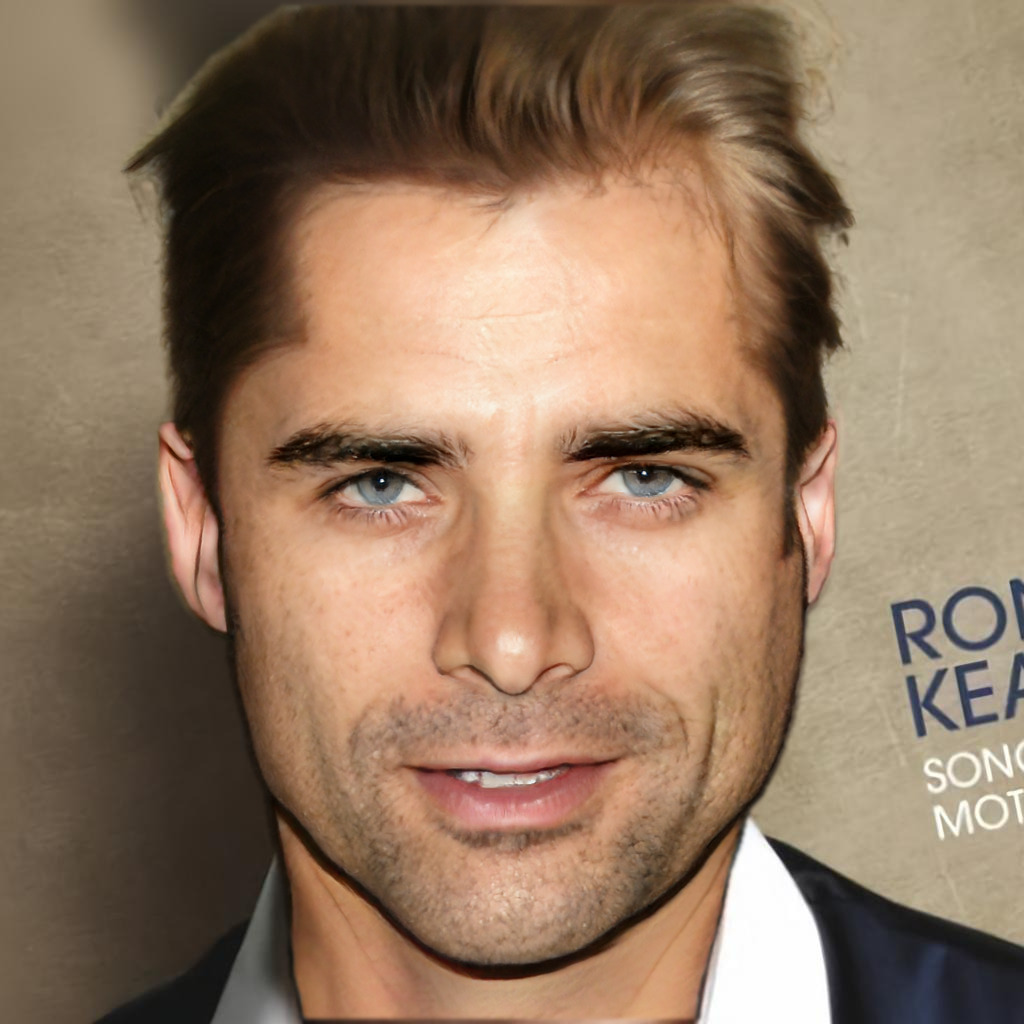}
\end{subfigure}
\captionsetup{belowskip=-15pt}
\caption{Comparisons with StyleGAN-based face swapping methods. Our method can achieve high-fidelity results while preserving the identity from the source better (\eg, skin, beard, eyes).}
    \label{fig:StyleGANBasedSwappingComp}
\end{figure}

\noindent{\textbf{Qualitative results.}} The qualitative comparisons are shown in \cref{fig:swapComp} and \cref{fig:StyleGANBasedSwappingComp}. 
It can be observed that our method achieves more realistic and high-fidelity swapped results. 
Compared with FSGAN~\cite{nirkin2019fsgan}, our results are much sharper. 
For SimSwap~\cite{chen2020simswap} and Hififace~\cite{wang2021hififace}, their swapped faces suffer from some artifacts and distortions (the 2nd row). 
Our E4S and FaceShifter~\cite{li2019faceshifter} can generate visually satisfying results; however, our approach retains detailed textures better.
We further compare the performance in more challenging cases where the occlusion exists in the source and target faces (the last two rows in \cref{fig:swapComp}). 
We can clearly see that our method can fill out the missing skin for the source face (the 3rd row), and maintain the glasses in the target face (the last row). 
Though a dedicated inpainting sub-network is designed in FSGAN, their inpainted results are quite blurry. 
FaceShifter proposes a refinement network to maintain the occlusion in the target image, but this may bring back some identity information of the target, making their swapped results similar to the target.
Note that ours is the only method that can well preserve the skin color of the source, which is also an identity-related attribute. In case the source skin is not fully adapted to the target due to the different lighting conditions, one can add a lighting transfer step for the source in advance. We leave disentangling the light from the texture in future work.

For a fair comparison, we also show the results of some StyleGAN-based approaches against ours in \cref{fig:StyleGANBasedSwappingComp}.
Although all these methods utilize the pre-trained StyleGAN, we find the results of MegaFS~\cite{zhu2021MegaFS} look to be a mixture of the source and target, which are blurry and lack of textures.
The results of StyleFuison~\cite{kafri2021stylefusion} show a bit of over-smoothing, \eg the Adam Levine case. Though $1024^2$ resolution can be achieved by HiRes~\cite{xu2022high}, their results still suffer from some artifacts. In contrast, our method can generate more realistic and high-quality faces. Please refer to our supplement for more results.

\begin{table}[]
\caption{Quantitative comparison for face swapping. The numbers in \textbf{bold} denote the best results. $\dagger$: source-oriented method, $\ddagger$: target-oriented method, *: StyleGAN-based method}
\vspace{-0.17cm}
\begin{tabular}{|l|cccc|}
\hline
\multicolumn{1}{|c|}{\multirow{2}{*}{\textbf{Method}}} & \multicolumn{2}{c}{\textbf{ID retrieval$\uparrow$}} & \multirow{2}{*}{\textbf{Pose$\downarrow$}} & \multirow{2}{*}{\textbf{Expr.$\downarrow$}}  \\
\multicolumn{1}{|c|}{}  & Top-1  & Top-5   &  &  \\ \hline
FSGAN $\dagger$~\cite{nirkin2019fsgan}   & 0.17   & 0.32  & 2.33  & 2.45    \\
SimSwap $\ddagger$~\cite{chen2020simswap}      & 0.12   & 0.32  & 2.89  & 2.84 \\
FaceShifter $\ddagger$~\cite{li2019faceshifter} & 0.06  & 0.26  & \textbf{1.73} & \textbf{2.35}  \\
HifiFace $\ddagger$~\cite{wang2021hififace}      & 0.15   & 0.37  & 2.77  & 2.82 \\
MegaFS *~\cite{zhu2021MegaFS}        & 0.29   & 0.45  & 3.03  & 3.05   \\
StyleFusion *~\cite{kafri2021stylefusion}    &  0.35   &  0.18  & 5.37  & 2.94  \\
HiRes $\ddagger$ *~\cite{xu2022high}    & 0.05   & 0.51 & 2.71  & 2.83  \\
Ours $\dagger$ *           & \textbf{0.38}  & \textbf{0.54}  & 3.29  & 3.05  \\ \hline
\end{tabular}
\vspace{-0.35cm}
\label{tbl:swapComp}
\end{table}
\raggedbottom
\noindent{\textbf{Quantitative results.}} We also conduct a quantitative comparison with the leading methods with respect to the identity preservation from the source and the attribute preservation from the target. The results are reported in~\cref{tbl:swapComp}.
For source identity preservation, we first extract the ID feature vectors of all the source faces and the swapped results using CosFace~\cite{wang2018cosface}. 
For each swapped face, we perform face retrieval by searching for the most similar face from all the source faces. 
The similarity is measured by the cosine distance.
Top-1 and Top-5 accuracy are the evaluation metrics. 
As for the target attribute preservation, we use HopeNet~\cite{Ruiz2018HopeNet} and a 3D face reconstruction model~\cite{deng2019accurate} to estimate the pose and expression, respectively. 
We calculate the $\ell_2$ distance of the pose and expression between each swapped face and its ground-truth target face.

It can be observed that our method achieves the best retrieval accuracy, which demonstrates that our swapped faces keep the identity from the source mostly. 
The visual comparisons shown in Fig.~\ref{fig:swapComp} also support this observation.
As for the target attribute preservation, our performance in pose and expression is still comparable with SOTA methods. Generally speaking, the target-oriented methods perform better in maintaining pose and expression, as they start from the target while our source-oriented method needs to generate these information starting from the source. However, one side effect of the target-oriented methods is that they only modify the shape and texture of the target face slightly and cannot fully preserve the identity (see FaceShifter in~\cref{fig:swapComp} and HiRes in~\cref{fig:StyleGANBasedSwappingComp}). That is, there is a trade-off between identity and attribute preservation in these methods. 
Note that the accuracy of the face reenactment method in our \textit{E4S} framework is the key factor that affects the pose and expression preservation. 
Our \textit{E4S} framework is generic, and the performance could be further improved with a more advanced reenactment model.

\begin{figure}[t]
\centering
\setkeys{Gin}{width=\linewidth}
\vspace{-0.4cm}
\begin{subfigure}{0.131\linewidth}
    \caption*{orig. msk}
\includegraphics{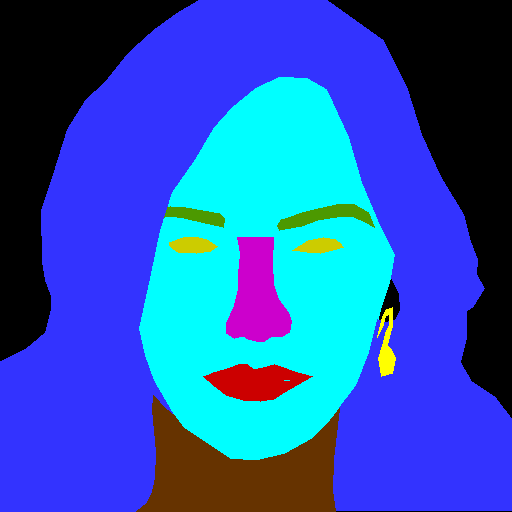}\\
\includegraphics{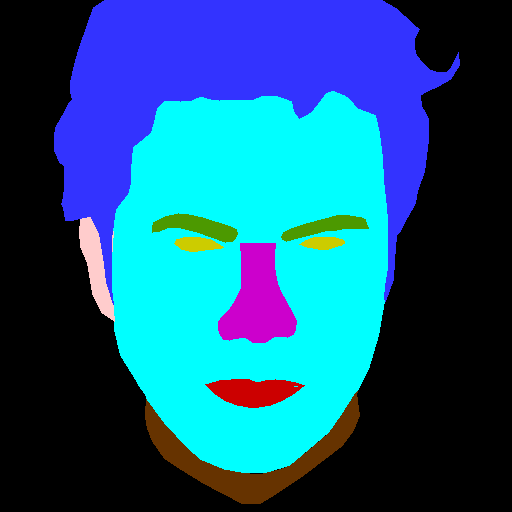}\\
\includegraphics{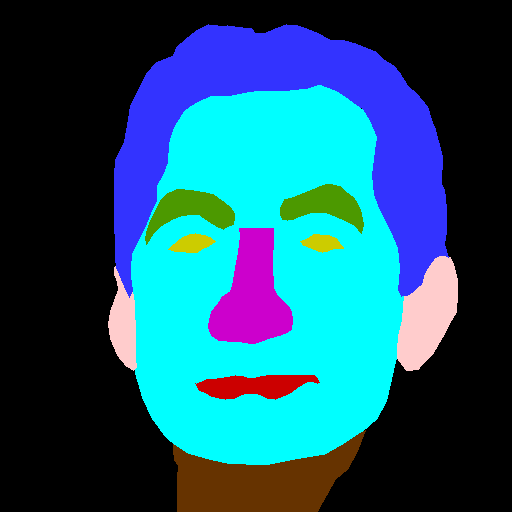}
\end{subfigure}
\begin{subfigure}{0.131\linewidth}
    \caption*{orig. img}
\includegraphics{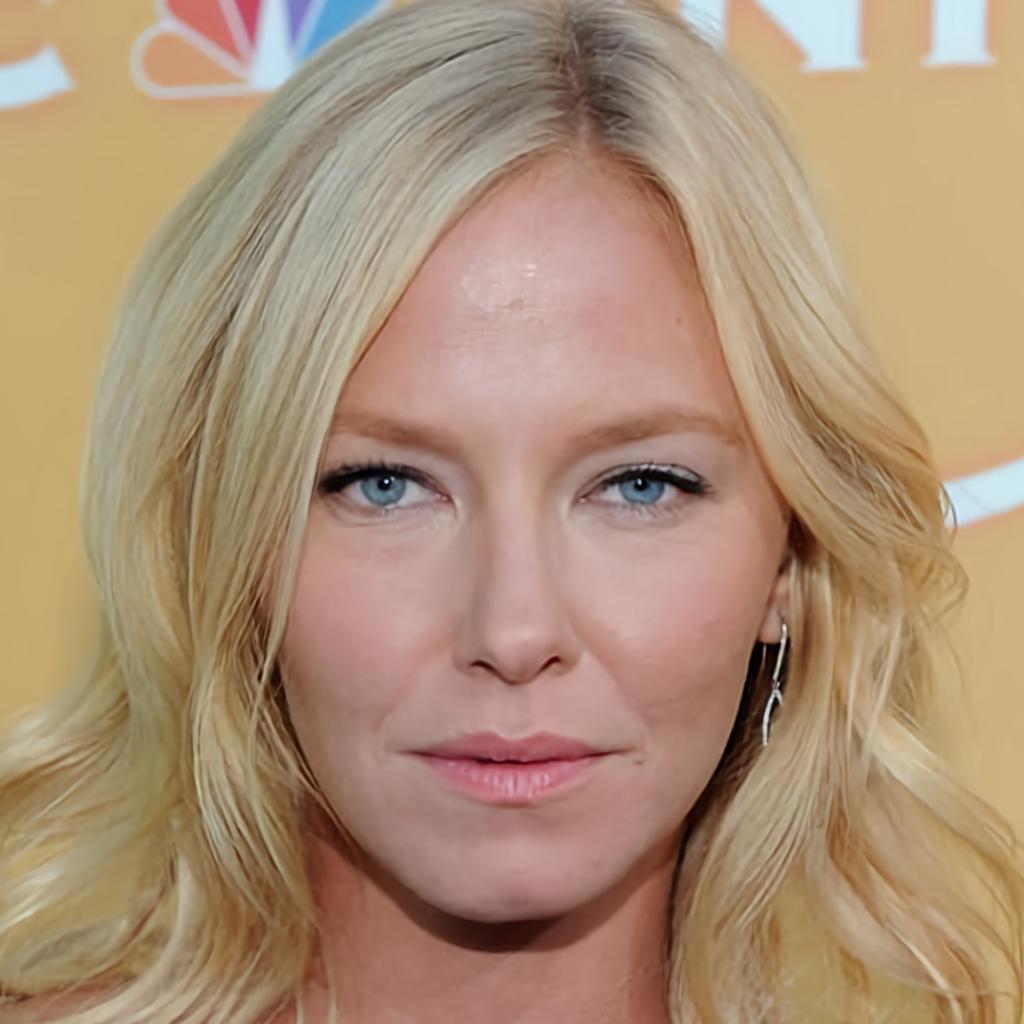}\\
\includegraphics{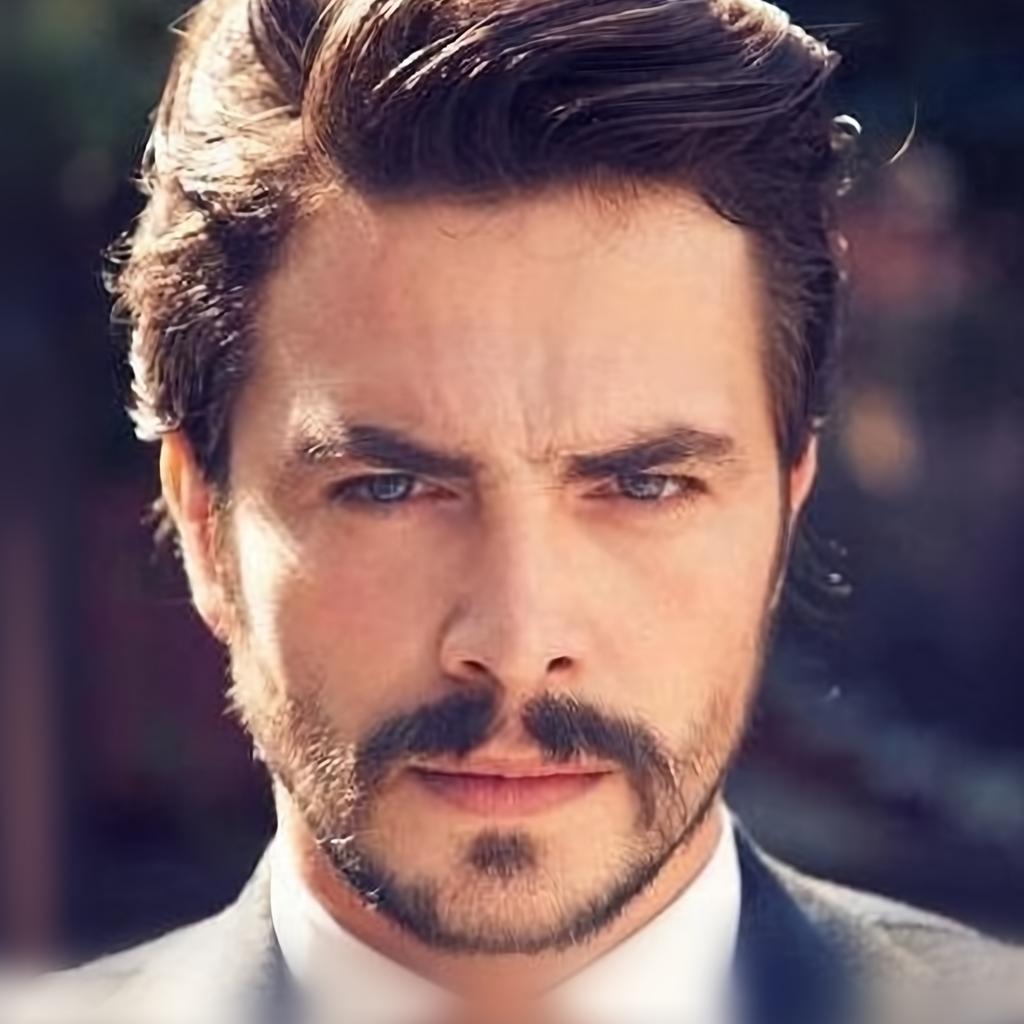}\\
\includegraphics{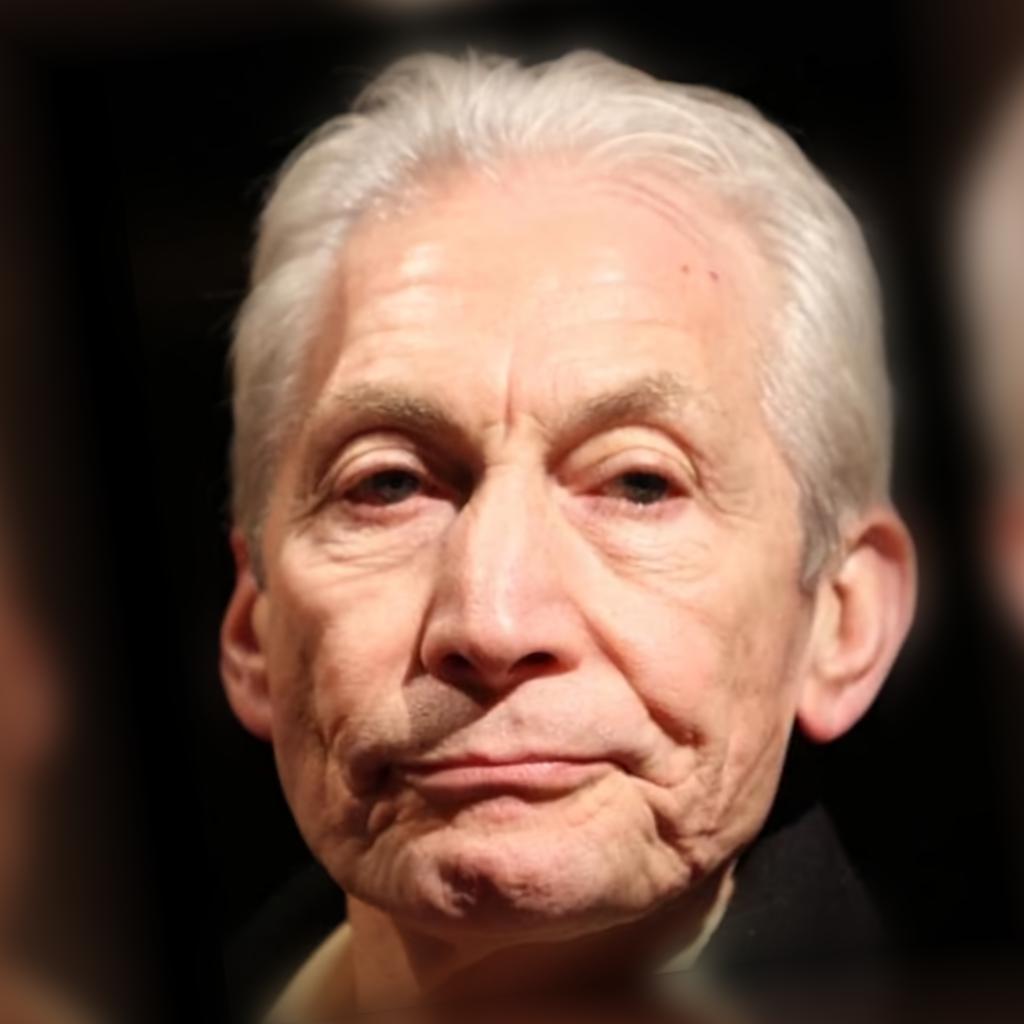}
\end{subfigure}
\begin{subfigure}{0.131\linewidth}
    \caption*{edited}
\includegraphics{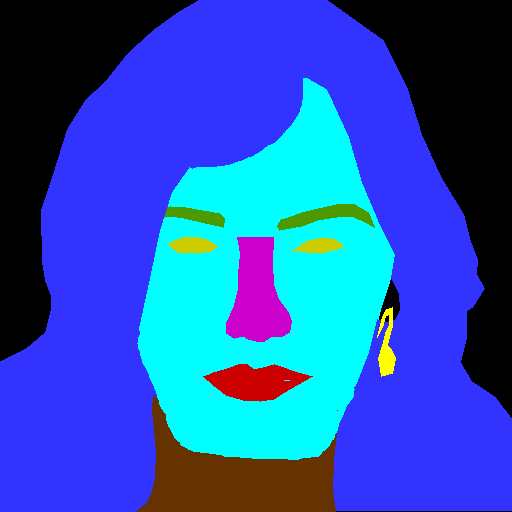}\\
\includegraphics{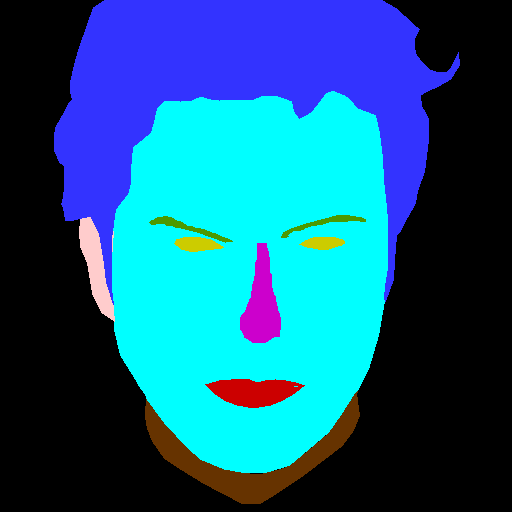}\\
\includegraphics{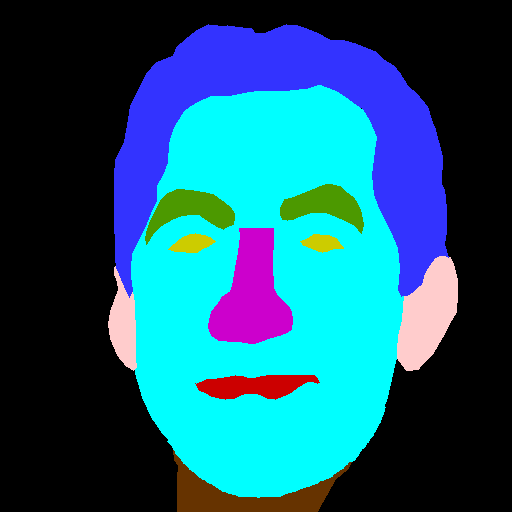}
\end{subfigure}
\begin{subfigure}{0.131\linewidth}
    \caption*{SPADE}
\includegraphics{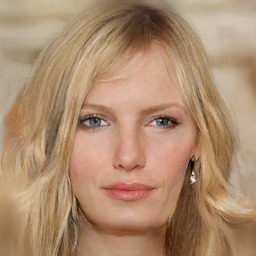}\\
\includegraphics{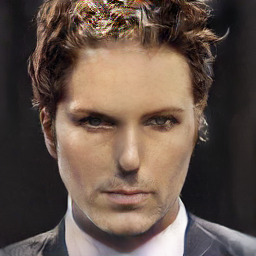}\\
\includegraphics{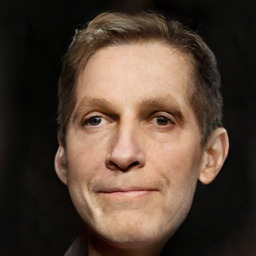}
\end{subfigure}
\begin{subfigure}{0.131\linewidth}
    \caption*{SEAN}
\includegraphics{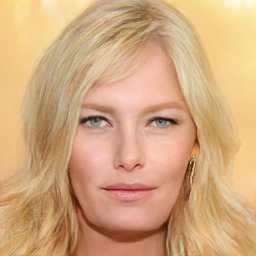}\\
\includegraphics{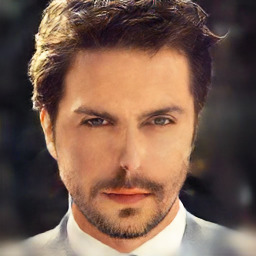}\\
\includegraphics{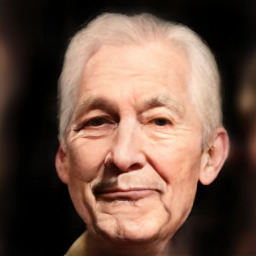}
\end{subfigure}
\begin{subfigure}{0.131\linewidth}
    \caption*{MaskGAN}
\includegraphics{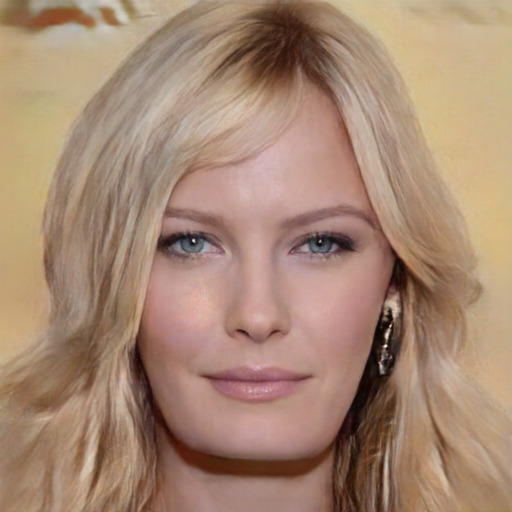}\\
\includegraphics{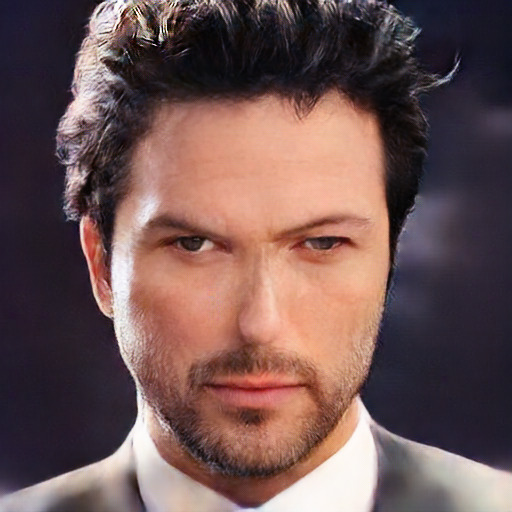}\\
\includegraphics{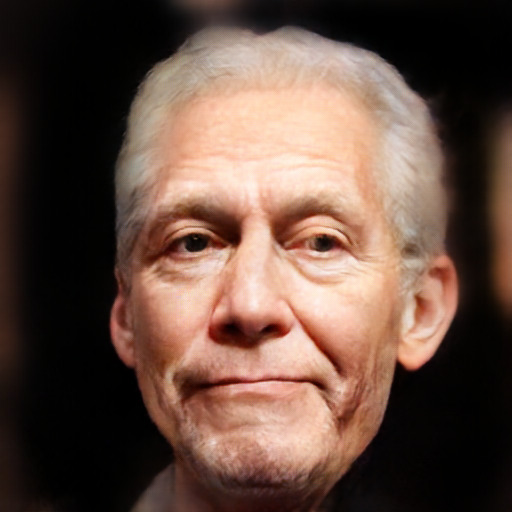}
\end{subfigure}
\begin{subfigure}{0.131\linewidth}
    \caption*{Ours}
\includegraphics{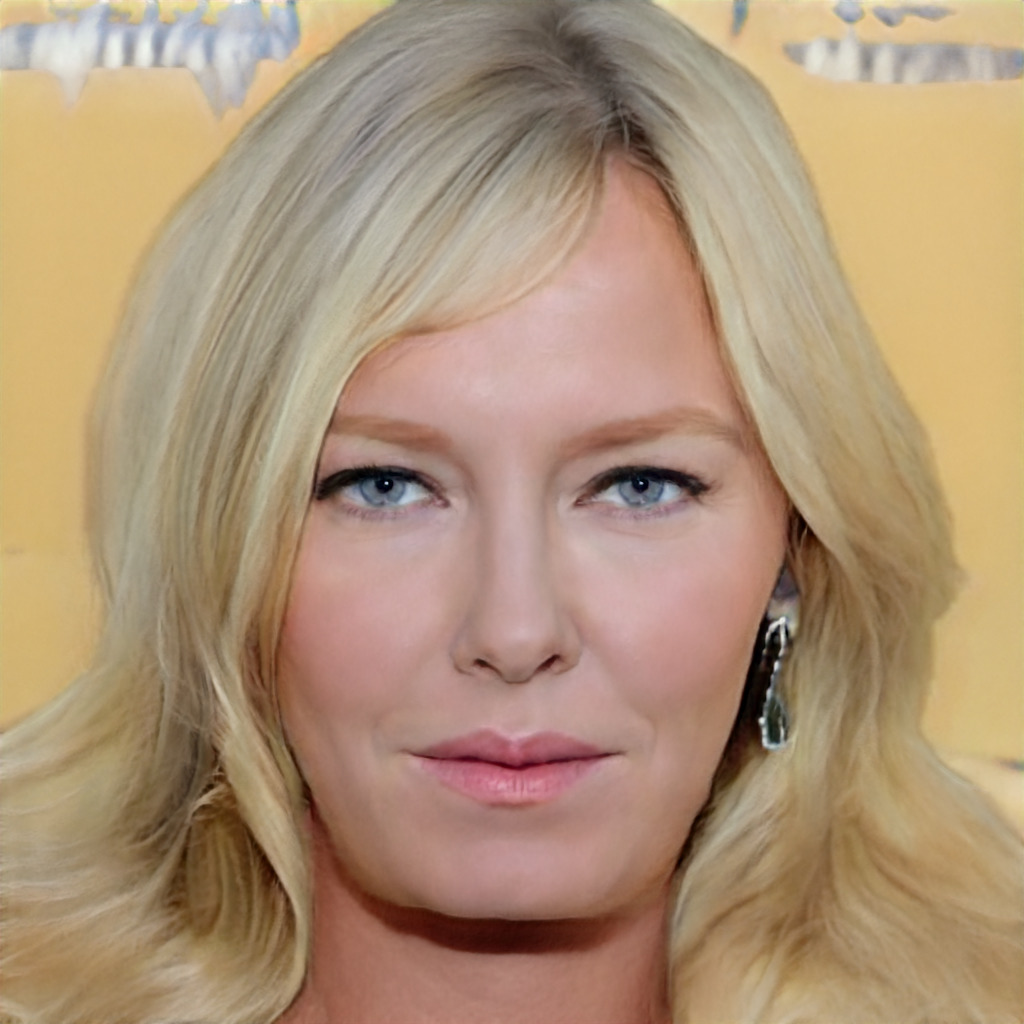}\\
\includegraphics{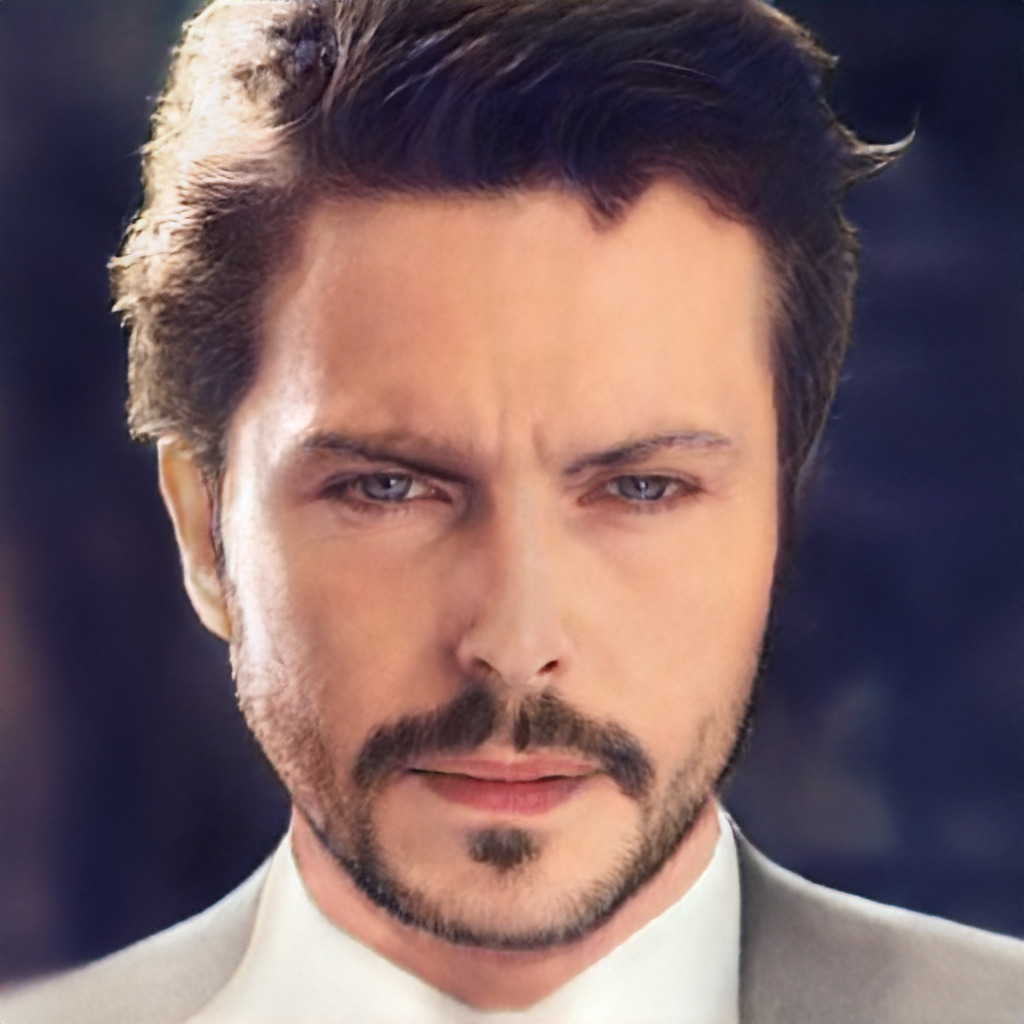}\\
\includegraphics{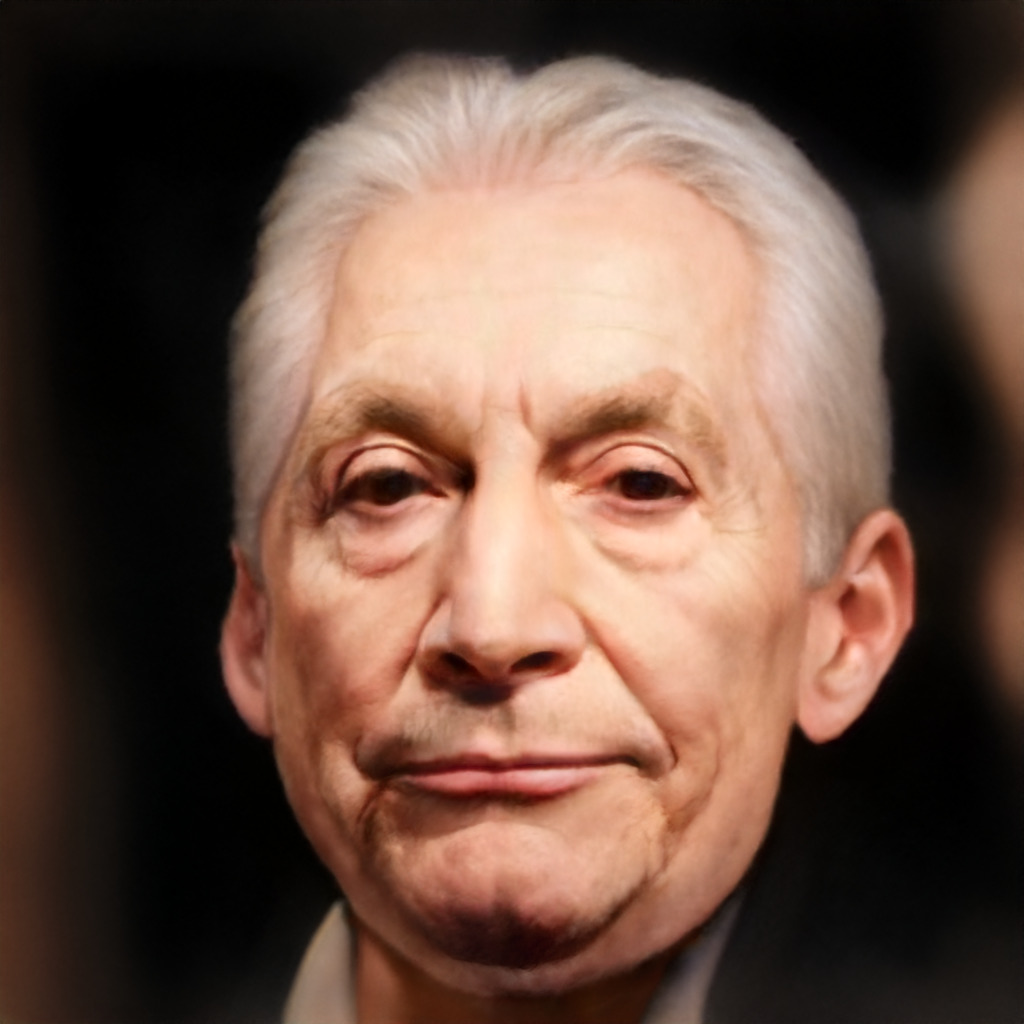}
\end{subfigure}
    \captionsetup{belowskip=-15pt}
    \caption{Qualitative comparisons of our results with state-of-the-art face editing methods. Some modifications are made on the face contour, hair, nose, and eyebrows. Our method can produce more high-fidelity editing results while maintaining the details of other components and the overall identity information well.}
    \label{fig:edit}
    \end{figure}

\vspace{-0.2cm}
\section{Face editing results}
Other than face swapping, our RGI {resided in the $\mathcal{W}^{r+}$} space can also be used for fine-grained face editing conveniently. One can edit the mask of a specific region or swap the style of a specific region (\eg, eyes, lips) with a reference image to obtain the desired editing results. This enables our RGI to support various applications such as face beautification, hairstyle transfer, and controlling the swapping extent of face swapping.
Please consult our supplement for the details.
In this section, we compare our method with the leading fine-grained face editing works: SPADE~\cite{park2019semantic}, 
SEAN~\cite{zhu2020sean}, and MaskGAN~\cite{lee2020maskgan}. For a fair comparison, we train our RGI network on the training set of CelebAMask-HQ and evaluate it on the test set.
We use the officially released pre-trained models of the competing methods to obtain their inference results. 

\begin{table}[t]
\vspace{-0.3cm}
\caption{Quantitative comparison for image reconstruction on CelebAMask-HQ~\cite{lee2020maskgan} test set. The rows in \colorbox{mygray}{gray} indicate the reconstruction images are obtained via style code optimization.}
\vspace{-0.2cm}
\small
\label{tab:recon}
\begin{tabular}{|l|cccc|}
\hline
\multicolumn{1}{|c|}{\textbf{method}} & \textbf{SSIM$\uparrow$} & \textbf{PSNR$\uparrow$} & \textbf{RMSE$\downarrow$} & \textbf{FID$\downarrow$} \\ \hline
SPADE~\cite{park2019semantic}                           & 0.64       & 15.67    & 0.17        & 20.45        \\
SEAN~\cite{zhu2020sean}          & 0.71                    & 18.57  & 0.12       & 17.74        \\
MaskGAN~\cite{lee2020maskgan}    & 0.75                    & 19.42  & 0.11       & 19.03        \\ 
Our RGI & \textbf{0.82}         & \textbf{19.85}  & \textbf{0.10}       & \textbf{15.03}        \\
\hline
\rowcolor{mygray} SofGAN~\cite{chen2021sofgan}  & 0.76            & 14.86   & 0.19  & 26.73          \\ 
\rowcolor{mygray} RGI-Optim.  & \textbf{0.86}  & \textbf{23.02} & \textbf{0.07} &\textbf{14.73}        \\ \hline
\end{tabular}
\vspace{-0.35cm}
\end{table}
\raggedbottom
\noindent{\textbf{Qualitative results.}}
We show the visual comparison with the competing methods in~\cref{fig:edit}. We make some modifications to the original facial mask, such as hair, eyebrows, and chin. It can be observed that our approach produces more high-fidelity and natural editing results, where the details of other components and the overall identity information are well maintained.

\noindent{\textbf{Quantitative results.}} We measure the image reconstruction quality of the competing methods and our RGI. The results are reported in~\cref{tab:recon}, where the SSIM~\cite{wang2004SSIM}, PSNR, RMSE, and FID~\cite{heusel2017FID} are used as the metrics. 
We also compare with  SofGAN~\cite{chen2021sofgan}, which is a StyleGAN-like generative model that relies on style code optimization for the reconstruction. For a fair comparison, an optimization stage is applied to our RGI (\ie, RGI-Optim.). 
As shown in~\cref{tab:recon}, our method always beats others in terms of all metrics, which indicates the visual inspection superiority of our method. 
We find SEAN~\cite{zhu2020sean} sometimes produces artifacts on hair regions. 
In contrast, our RGI can achieve high-fidelity reconstructions, keeping identity, texture, and illumination better. 
Besides, our RGI-Optim. can preserve the facial details better (\eg, the curly degree of hair, the thickness of the beard, dimples, and background). 
For more visual comparisons, please check our supplement.

\vspace{-0.1cm}
\section{Conclusion} 
In this paper, we present a novel framework \textit{E4S} for face swapping, which explicitly disentangles the shape and texture of each facial component and reformulates face swapping as a simplified problem of texture and shape swapping. To seek such disentanglement as well as high resolution and high fidelity, we propose a novel Regional GAN Inversion (RGI) method. Concretely, a multi-scale mask-guided encoder projects input faces into the per-region style codes. Besides, a mask-guided injection module uses the style codes to manipulate the feature maps in the generator according to the given masks. Extensive experiments on face swapping, face editing and other extended applications demonstrate the superiority of our method.

\section*{Acknowledgement} 
This research was sponsored by Natural Science Foundation of China (62072191). We thank Hanbang Liang for sharing the reproduced code of FaceShifter, and thank reviewers for their thoughtful comments and suggestions.

{\small
\bibliographystyle{ieee_fullname}
\bibliography{ref}
}

\end{document}